\PassOptionsToPackage{pagebackref}{hyperref}
\documentclass[]{OpenEnvision}

\usepackage[toc,page,header]{appendix}

\usepackage{natbib}

\usepackage{CJKutf8}

\usepackage{xargs}  

\usepackage{todonotes}  

\usepackage{multirow}

\usepackage{cleveref}

\usepackage{amsmath}
\usepackage{dsfont}

\usepackage{subcaption}

\usepackage{svg}

\usepackage{mathrsfs}
\usepackage{adjustbox}
\usepackage{multirow}
\usepackage{multicol}
\usepackage{xcolor}
\usepackage{caption}
\usepackage[most]{tcolorbox}
\usepackage{float}
\usepackage{changepage}
\usepackage{graphicx}
\usepackage{amssymb}
\usepackage{array}
\usepackage{bm}
\usepackage{minitoc}
\usepackage{booktabs}
\usepackage{tabularx}
\usepackage{longtable}
\usepackage{booktabs}
\usepackage{tabularx}
\usepackage{array}
\usepackage{makecell}
\usepackage{hyperref}
\hypersetup{breaklinks=true,colorlinks=true,allcolors=oeredark}
\ifPDFTeX
  \usepackage{fontawesome5}
\else
  \newcommand{\faIcon}[1]{%
    \ifstrequal{#1}{eye}{\ensuremath{\odot}}{%
    \ifstrequal{#1}{cubes}{\ensuremath{\Box}}{%
    \ifstrequal{#1}{magic}{\ensuremath{\star}}{%
    \ifstrequal{#1}{wave-square}{\ensuremath{\sim}}{%
    \ifstrequal{#1}{gamepad}{\ensuremath{\leftrightarrow}}{\ensuremath{\bullet}}}}}}%
  }
\fi
\usepackage[most]{tcolorbox}

\newcolumntype{Y}{>{\raggedright\arraybackslash}X}
\definecolor{tabhead}{HTML}{FCEAE8}
\definecolor{tabalt}{HTML}{F7F7F4}
\definecolor{seedred}{HTML}{E53A32}
\newcolumntype{Y}{>{\raggedright\arraybackslash}X}

\title{\textcolor{oered}{WorldFoundry}: Towards Unified Infrastructure for World Intelligence}

\renewcommand{\authorfont}{\fontsize{10}{12}\selectfont}
\author[]{OpenEnvision}
\author[\dagger,*]{Juanxi~Tian}
\author[*]{Jiadong~Hou}
\author[]{Baiang~Li}
\author[]{Xin~Lin}
\author[]{Zhengbo~Jiao}
\author[]{Weijia~Zhang}
\author[]{Yeying~Jin}
\author[]{Truong~Nguyen}
\author[]{Haitian~Zheng}
\author[]{Zhifei~Zhang}
\author[]{Jianfei~Yang}
\author[]{Ming-Hsuan~Yang}
\author[\ddagger]{Felix~Heide}
\author[\ddagger]{Wei~Xiong}
\author[\ddagger]{Enze~Xie}
\author[\ddagger]{Wanhua~Li}

\contribution[\dagger]{Project Leader}
\contribution[*]{Core Contributor}
\contribution[\ddagger]{Project Supervisors}
\appto\contributionlist{%
  \par
  {\contributionfont\itshape See \hyperref[sec:appendix]{Appendix~\ref*{sec:appendix}} for author contributions.\par}%
}

\renewcommand{\affiliationlist}{%
  \raisebox{-0.5\height}{\includegraphics[height=11.5mm,trim=0 0 511.44bp 0,clip]{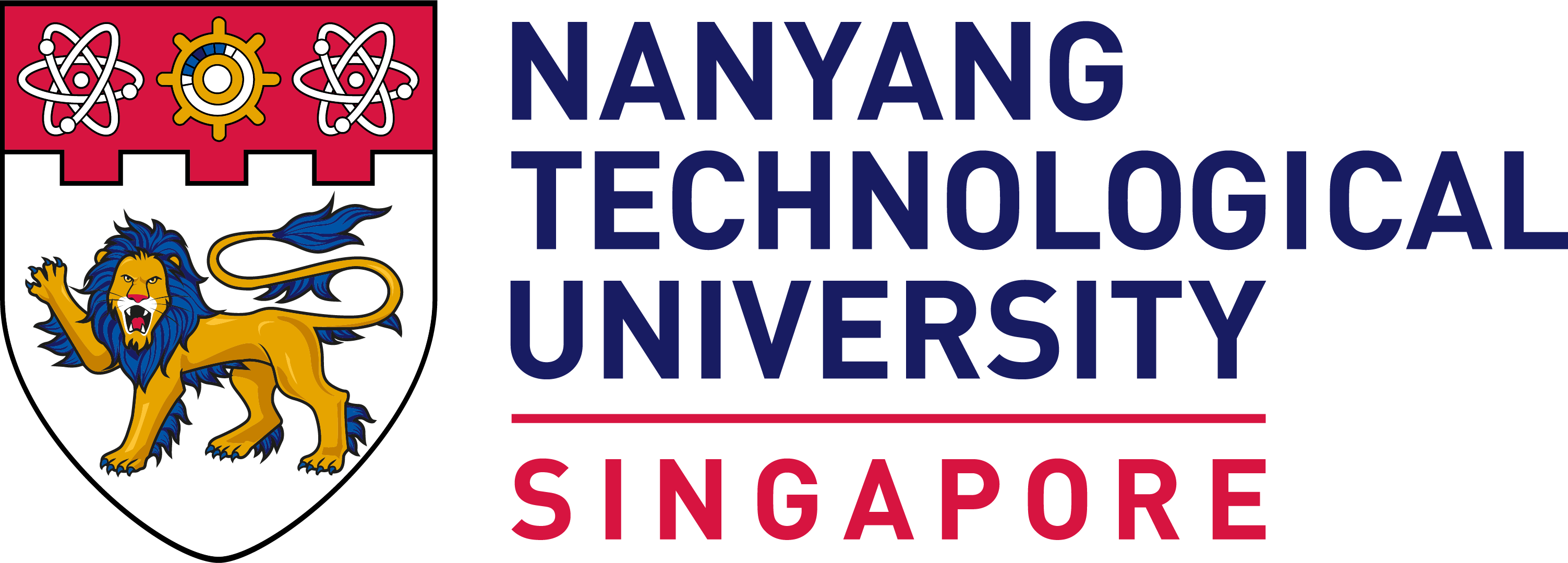}}\hspace{11mm}%
  \raisebox{-0.5\height}{\includegraphics[height=11.5mm,trim=0 0 153.25bp 0,clip]{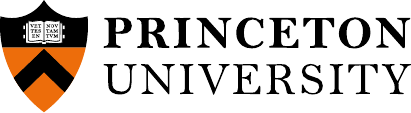}}\hspace{11mm}%
  \raisebox{-0.5\height}{\includegraphics[height=11.5mm]{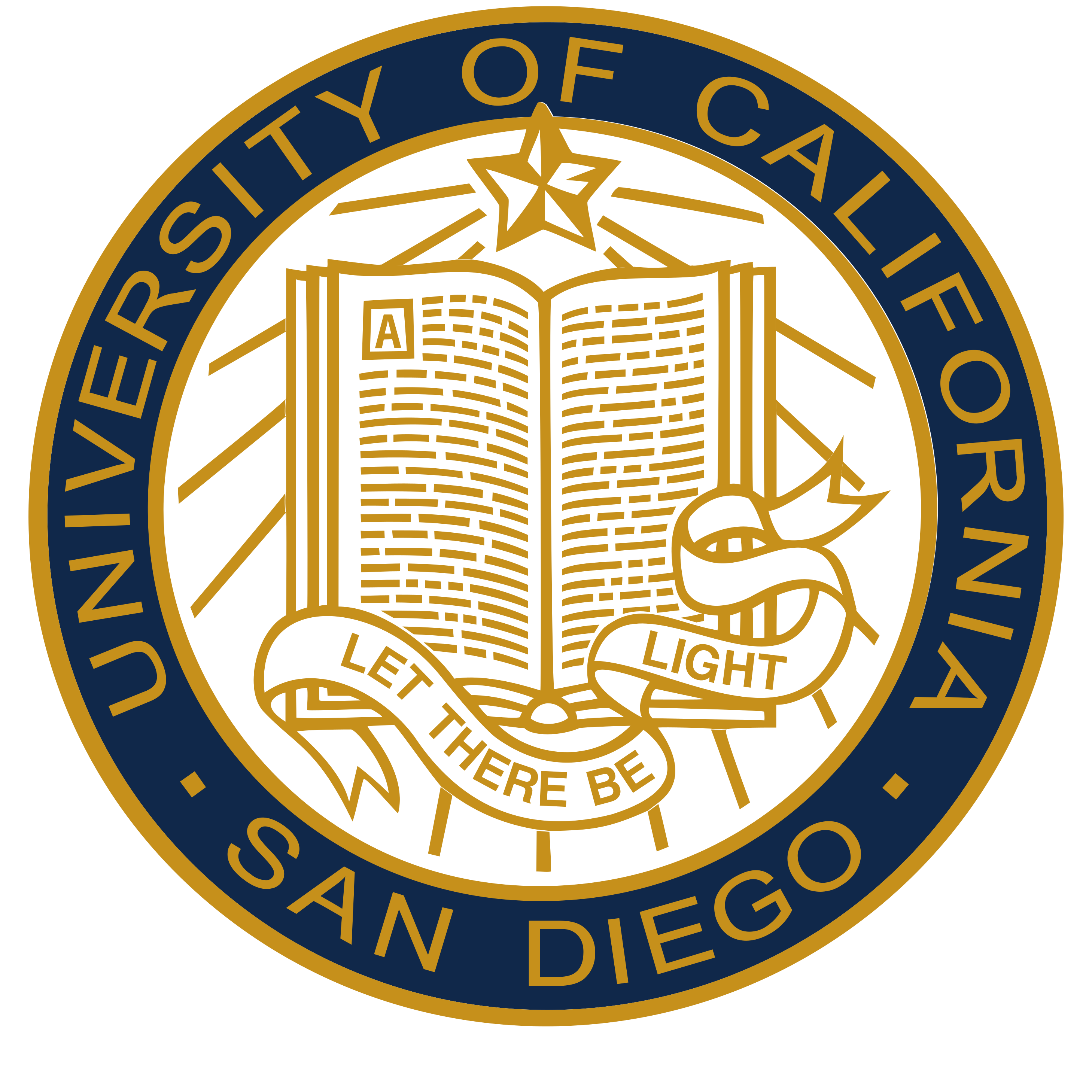}}\hspace{11mm}%
  \raisebox{-0.5\height}{\includegraphics[height=10.5mm,trim=0 0 1096.84bp 0,clip]{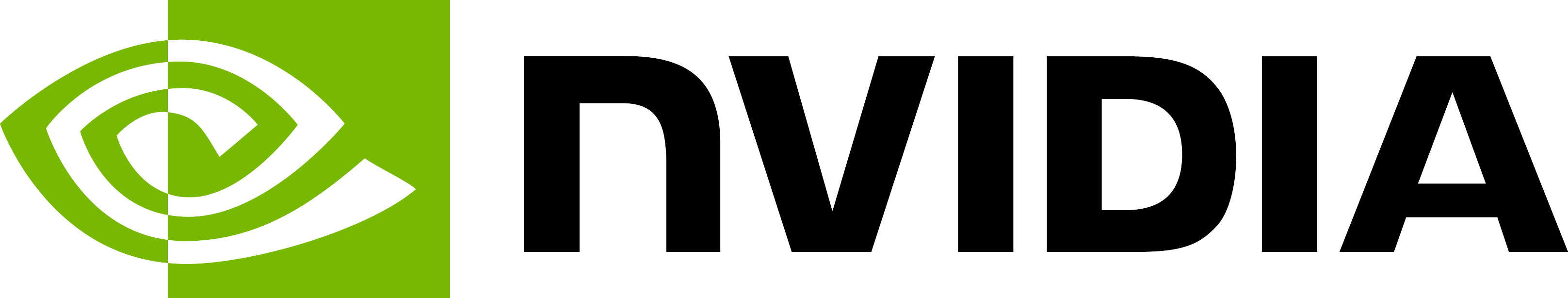}}\hspace{11mm}%
  \raisebox{-0.5\height}{\includegraphics[height=11.5mm]{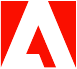}}%
  \par\vspace{3mm}%
  \raisebox{-0.5\height}{\includegraphics[height=11.5mm]{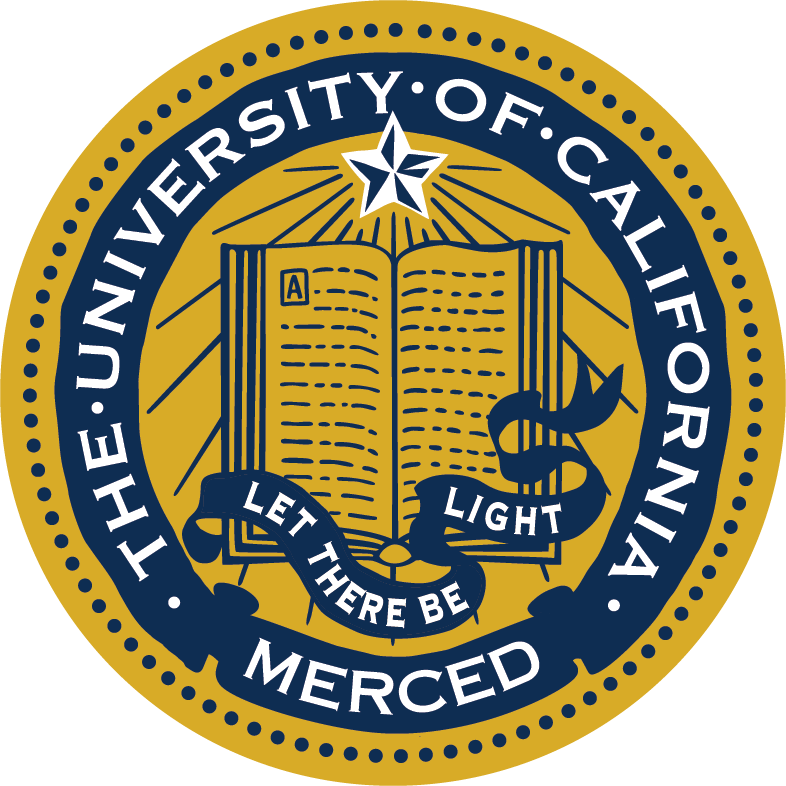}}\hspace{17mm}%
  \raisebox{-0.5\height}{\includegraphics[height=11.5mm,trim=0 0 330bp 0,clip]{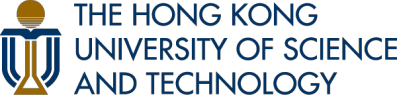}}\hspace{17mm}%
  \raisebox{-0.5\height}{\includegraphics[height=11.5mm]{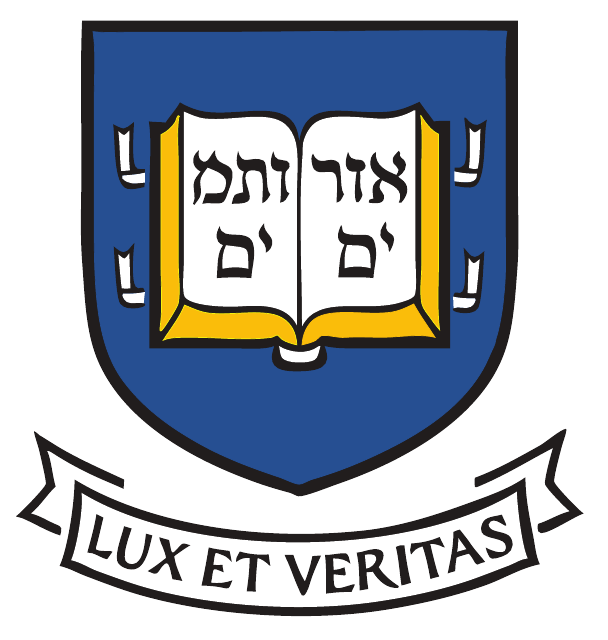}}\hspace{17mm}%
  \raisebox{-0.5\height}{\includegraphics[height=11.5mm,trim=0 0 100bp 0,clip]{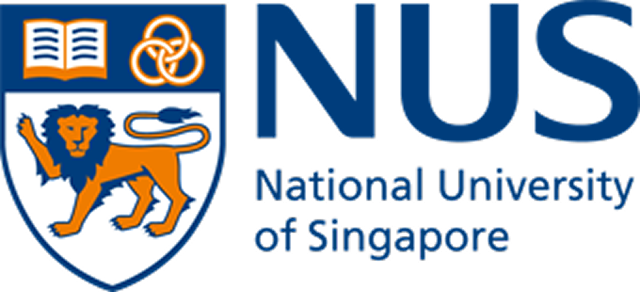}}%
}

\abstract{
Visual generative models increasingly blur the line between synthesizing observations and simulating environments. Yet visual plausibility alone does not constitute successful world modeling: compelling outputs may lack persistent state, physical dynamics, or causally consistent responses to interactions. The absence of a shared operational basis, together with perception-centric benchmarks and fragmented workflows, obscures the distinction between generative foundation models and generative world models and hinders reproducible assessment of dynamics, controllability, and interactive utility. We define a generative world model as a predictive dynamical system that maintains a persistent, history-conditioned representation of an environment, updates that representation in response to actions or interventions, and generates temporally coherent trajectories that reflect outcomes under observed and counterfactual interventions. From this definition, we derive an intelligence-oriented framework spanning perceptual grounding, persistent state modeling, controllable transformation, dynamics-aware prediction, and closed-loop interaction. We further present \textbf{\textsc{WorldFoundry}}, open-source infrastructure for reproducible inference and evaluation across video, 3D and 4D, embodied action, and interactive world systems. WorldFoundry unifies model and benchmark discovery, runtime and asset preparation, execution orchestration, artifact management, and interactive visualization, while modality-specific adapters preserve native inputs, execution semantics, output representations, and evaluation protocols. Explicit manifests, durable artifact contracts, and evidence-aware scorecards separate execution readiness from evaluative claims, making each claim traceable to its supporting artifacts. The current catalog contains 270 model entries, 74 benchmarks, and 62 registered metric entries, with integration readiness and execution evidence recorded separately for each entry. By coupling world modeling with evidence-grounded executable infrastructure, WorldFoundry enables reproducible comparison of persistent, controllable dynamics across heterogeneous systems, advancing the field from perceptual synthesis toward measurable world intelligence.
}

\newcommand{\covericon}[1]{%
  \makebox[4mm][l]{\raisebox{-0.15ex}{#1}}%
}
\DeclareRobustCommand{\covercalendaricon}{%
  \covericon{\includegraphics[height=2.5mm]{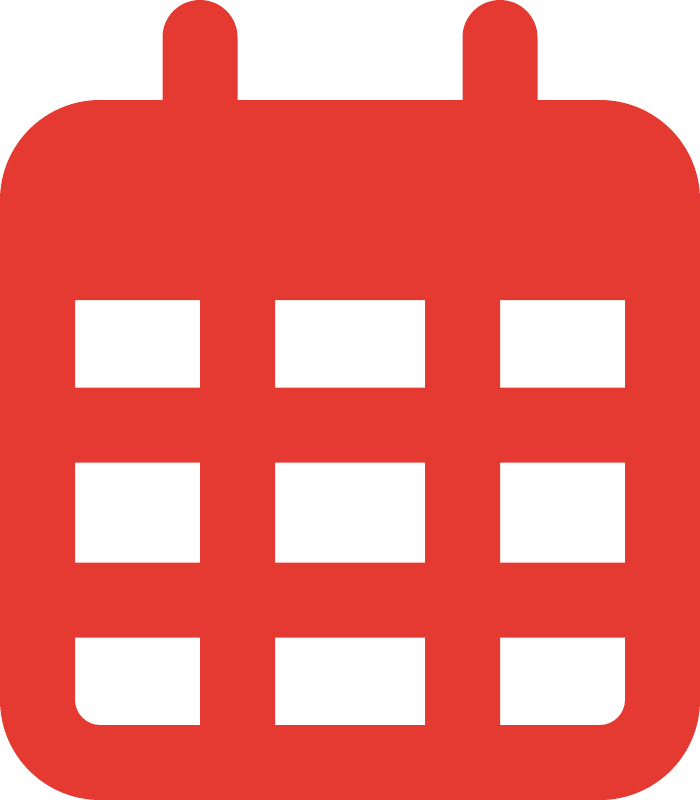}}%
}
\DeclareRobustCommand{\coverglobeicon}{%
  \covericon{\includegraphics[height=2.5mm]{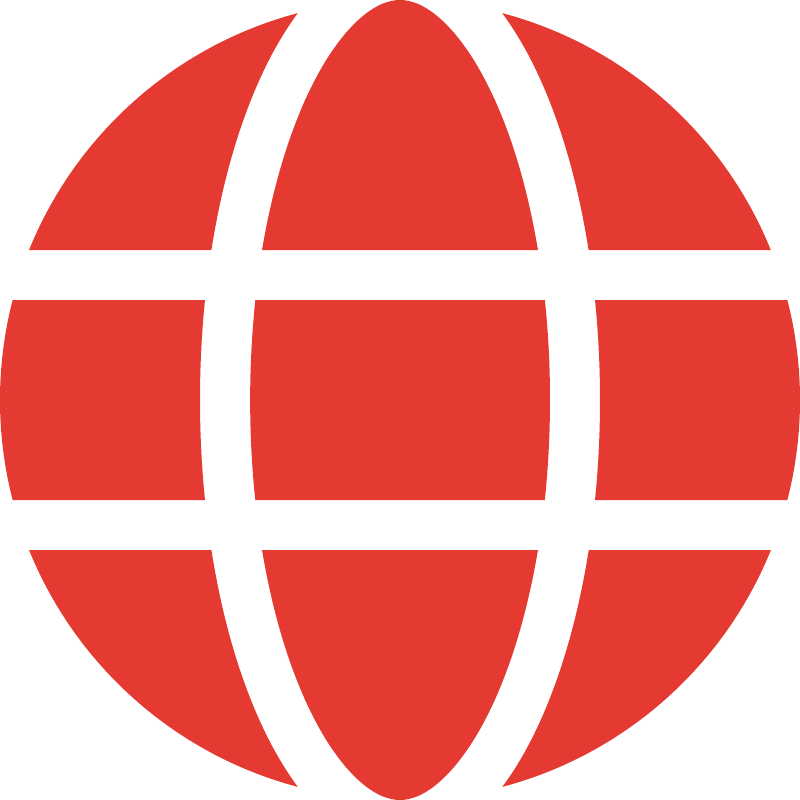}}%
}
\DeclareRobustCommand{\covergithubicon}{%
  \covericon{\includegraphics[height=2.5mm]{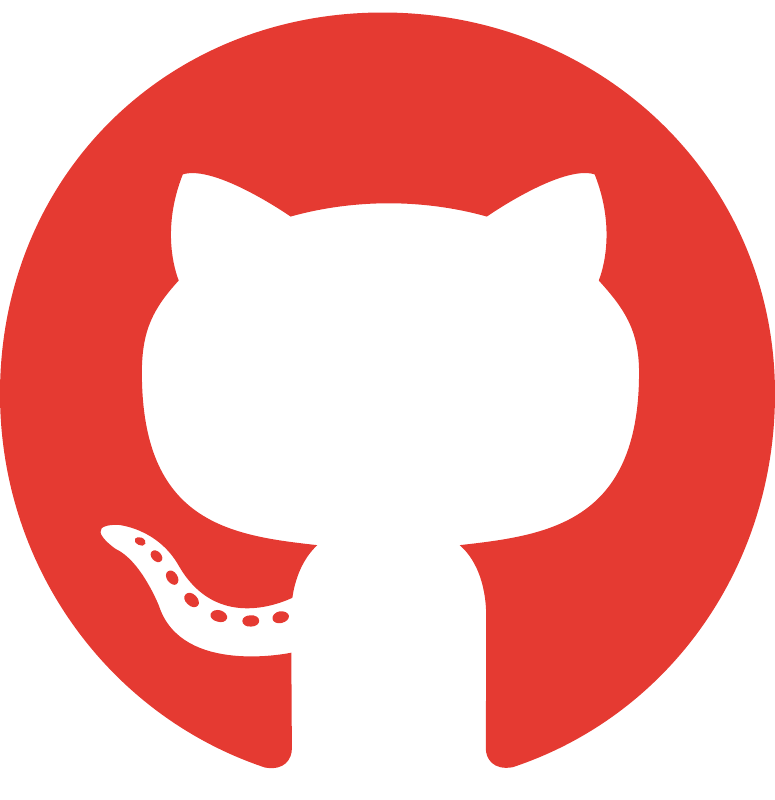}}%
}
\DeclareRobustCommand{\covermailicon}{%
  \makebox[4mm][l]{\raisebox{-0.15ex}{%
    \tikz[x=1mm,y=1mm]{%
      \fill[oered,rounded corners=0.3mm] (0,0) rectangle (3,2.25);
      \draw[white,line width=0.18mm,line cap=round,line join=round]
        (0.3,1.85) -- (1.5,0.95) -- (2.7,1.85);
    }%
  }}%
}
\checkdata[\covercalendaricon Date]{November 7, 2026}
\checkdata[\covermailicon Email]{\email{openenvision.lab@gmail.com}}
\checkdata[\coverglobeicon Project Page]{\url{https://openenvision.org/WorldFoundry/}}
\checkdata[\covergithubicon Code]{\url{https://github.com/OpenEnvision/WorldFoundry}}

\begin{document}
\begin{CJK*}{UTF8}{gbsn}

% Fit the author list, affiliations, and abstract on the cover.
\newgeometry{top=14mm,bottom=15mm,left=25mm,right=25mm}
\maketitle
\restoregeometry 
\tableofcontents
\newpage
\pagestyle{reportstyle}

\section{Introduction}
\label{sec:introduction}

\begin{figure}[H]
    \centering
    \includegraphics[width=\textwidth]{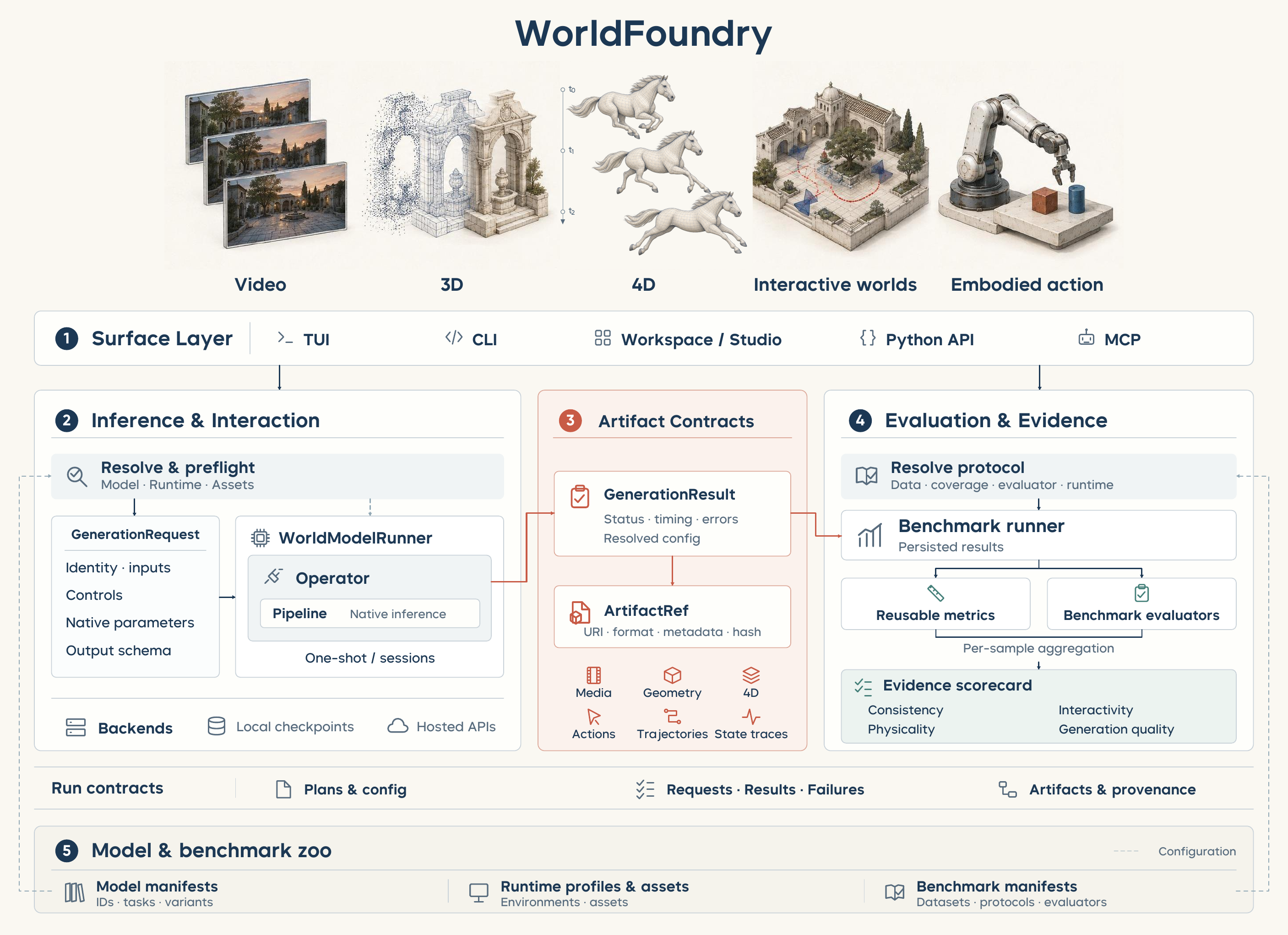}
    \caption{\textbf{Overview of \textsc{WorldFoundry}.} Open-source infrastructure for generative world models. (1) one run is reachable from every interactive and programmatic surface; (2) preflight requests execute through a unified runner, one-shot or session-resident; (3) outputs are persisted behind durable artifact contracts with typed references; (4) benchmarks score persist in artifacts reported in an evidence scorecard; and (5) declarative manifests bind discovery, runtime and evaluation, keeping claims traceable to evidence.}
    \label{fig:overview}
\vspace{-1em}
\end{figure}

Recent advances in generative modeling are moving beyond synthesizing individual objects or scenes toward modeling worlds that evolve over time and respond to interventions. This shift calls for a broader notion of \textit{world intelligence}, in which systems must perceive observations, infer structured states, predict their evolution, express resulting consequences through task-specific outputs, and exploit these predictions for interaction~\citep{worldmodels,planet,dreamerv3,vjepa2}. Video generators, neural scene representations, and embodied policies address complementary parts of this problem. However, incompatible representations, runtimes, interfaces, and evaluation protocols make capabilities difficult to compose, results difficult to reproduce, and progress toward genuine world models difficult to measure. The bottleneck is therefore not only algorithmic, but also infrastructural. Recent efforts such as OpenWorldLib~\citep{openworldlib} address complementary aspects of this integration problem. Progress therefore requires shared infrastructure grounded in a common functional abstraction that makes heterogeneous models composable, reproducible, and comparable across modalities while preserving their native semantics.

To establish common ground without constraining how different systems represent the world, we formulate the abstraction at the level of function. Building on insights from earlier definitions in reinforcement learning~\citep{worldmodels,planet,dreamerv3} and recent efforts to delineate their scope~\citep{openworldlib}, we identify three tightly coupled components. \textit{State inference} estimates a latent environmental state from the observation history; \textit{forward simulation} predicts its evolution under endogenous dynamics and external interventions; and \textit{interactive response} produces distinct, action-conditioned futures as visual, geometric, or task-specific outputs. This decomposition is functional rather than architectural and defines a shared interface, not a requirement that every system implement all three operations. Static 3D methods emphasize persistent-state inference and geometry decoding, video models primarily learn implicit transitions and observation generation, 4D methods explicitly represent time-varying structure, and embodied models connect state estimation and action prediction to decision-making. Characterizing systems by the operations they implement and the capability profiles they support provides a common vocabulary for comparing them.

These operations alone, however, describe what a system does without determining whether it qualifies as a world model. Qualification is decided not by any single operation but by the state that persists across all of them: we define a generative world model as a learned, history- and control-conditioned dynamical system that infers and maintains a persistent environmental state, evolves that state under endogenous dynamics and external interventions, and assimilates observational feedback into coherent future trajectories. Persistence in this sense is operational rather than nominal: an intervention must leave a trace in the state that constrains later predictions, and each incoming observation must in turn revise that state. Only when both hold is the world-modeling loop closed. Visual plausibility, by contrast, reflects only the observation decoder; it is neither necessary nor sufficient evidence that the underlying state persists, a gap that recent physics- and world-oriented benchmarks repeatedly expose~\citep{vbench2,videophy2,worldmodelbench,worldarena}. It is closure, not plausibility, that draws a principled boundary between \textit{foundational generative models}, which synthesize or transform observations without closing the loop, and \textit{generative world models}, which couple generation to an evolving environmental state through closed-loop updates. Measuring progress from foundational generation toward world intelligence therefore calls for evidence of closure, which in turn depends on shared infrastructure for reproducible execution and comprehensive, fair evaluation.

We realize this infrastructure in \textbf{\textsc{WorldFoundry}}, an open-source system for integrating, executing, and evaluating video, 3D, 4D, interactive world, and embodied-action models. WorldFoundry spans the research lifecycle, from model and benchmark discovery and runtime and asset preparation to execution orchestration, artifact management, evaluation, and interactive visualization. Modality-specific adapters preserve native execution and evaluation semantics; declarative manifests specify dependencies, assets, and interfaces; durable artifact contracts decouple execution from evaluation; and evidence-aware scorecards separate declared support, operational readiness, and validated performance. The current release integrates 270 models and over 100 benchmark and metric implementations, with every model runner validated through end-to-end execution of a documented reference workflow. By combining our theory of world modeling with the proposed infrastructure, WorldFoundry transforms isolated generative demonstrations into cumulative, auditable evidence and advances the field from perceptual synthesis toward measurable world intelligence.

Our key contributions can be summarized as follows:
\begin{itemize}[leftmargin=*]
    \item \textbf{World Model Definition.} We formally define world models through a minimal abstraction comprising state inference, forward simulation, and interactive response. We then characterize the core capabilities of perception, modeling, manipulation, dynamics, and interaction within a structured functional framework, thereby organizing and differentiating existing generative and world-modeling systems.
    
    \item \textbf{Unified Inference and Evaluation Infrastructure.} We introduce WorldFoundry, an open-source infrastructure that standardizes execution and evaluation across video, 3D, 4D, embodied-action, and interactive world systems, making inference reproducible and configurable without overriding native model semantics, and placing heterogeneous systems under a common operational definition of world modeling.

    \item \textbf{Validation at Scale.} We integrate 270 generative models and over 100 benchmark and metric implementations. We validate every model runner through a documented end-to-end reference workflow, yielding an auditable record of operational readiness under the tested configurations.
\end{itemize}

% Keep the definition section heading with its opening paragraph.
\clearpage
\section{Related Work}
\label{sec:related-work}
\DeclareRobustCommand{\rwPartial}{\tikz[baseline=-0.45ex]{\fill (0,0.7ex) arc[start angle=90,end angle=270,radius=0.7ex] -- cycle;\draw[line width=0.3pt] (0,0) circle[radius=0.7ex];}}
\DeclareRobustCommand{\rwYes}{\mbox{\fontsize{10}{10}\selectfont\ensuremath{\checkmark}}}
\begin{table}[H]
    \centering
    \caption{Comparison with world-model infrastructures, modality-specific toolkits, and evaluation suites. \rwYes: supported; \rwPartial: specialized support; \textcolor{oemuted}{\ensuremath{\times}}: no documented workflow. Column definitions follow the table.}
    \label{tab:related-infrastructure}
    \begingroup
    \normalem
    \hypersetup{urlcolor=black}
    \newcommand{\rwNo}{\textcolor{oemuted}{\ensuremath{\times}}}
    \newcommand{\rwHead}[1]{{\sffamily\bfseries\fontsize{7.1}{8.3}\selectfont #1}}
    \newcommand{\rwGroup}[1]{\rowcolor{tabhead}\multicolumn{13}{@{}l@{}}{{\color{oered}\sffamily\bfseries\fontsize{8}{9.5}\selectfont\rule[-0.1ex]{1.4pt}{1.1ex}\hspace{5pt}#1}}\\}
    \fontsize{7.8}{9.5}\selectfont
    \setlength{\tabcolsep}{1.1pt}
    \renewcommand{\arraystretch}{1.12}
    \begin{tabularx}{\linewidth}{@{}
        >{\raggedright\arraybackslash}p{0.205\linewidth}
        *{11}{>{\centering\arraybackslash}p{0.054\linewidth}}
        >{\raggedright\arraybackslash}X@{}}
    \toprule
    & \multicolumn{4}{c}{\textcolor{oered}{\rwHead{Coverage}}}
    & \multicolumn{4}{c}{\textcolor{oered}{\rwHead{Execution}}}
    & \multicolumn{2}{c}{\textcolor{oered}{\rwHead{Evaluation}}}
    & \multicolumn{2}{c}{\textcolor{oered}{\rwHead{Basis}}} \\
    \cmidrule(lr){2-5}\cmidrule(lr){6-9}\cmidrule(lr){10-11}\cmidrule(lr){12-13}
    \rwHead{System} & \rwHead{Video} & \makecell{\rwHead{3D/}\\\rwHead{4D}} & \makecell{\rwHead{Inter.}\\\rwHead{WM}} & \makecell{\rwHead{Embo-}\\\rwHead{died}} & \makecell{\rwHead{Model}\\\rwHead{runner}} & \makecell{\rwHead{Hosted}\\\rwHead{API}} & \makecell{\rwHead{Closed}\\\rwHead{loop}} & \makecell{\rwHead{Train-}\\\rwHead{ing}} & \makecell{\rwHead{Proto-}\\\rwHead{cols}} & \makecell{\rwHead{Evid-}\\\rwHead{ence}} & \makecell{\rwHead{WM}\\\rwHead{def.}} & \rwHead{Interfaces} \\
    \midrule
    \rwGroup{WORLD-MODEL INFRASTRUCTURE}
    \href{https://github.com/OpenDCAI/OpenWorldLib}{OpenWorldLib}~\citep{openworldlib,openworldlib_software} & \rwYes & \rwYes & \rwYes & \rwYes & \rwYes & \rwYes & \rwYes & \rwNo & \rwYes & \rwPartial & \rwYes & Py $\cdot$ CLI \\
    \href{https://github.com/NVIDIA/cosmos-framework}{Cosmos}~\citep{cosmos,cosmos_software,cosmos_evaluator} & \rwYes & \rwNo & \rwYes & \rwYes & \rwPartial & \href{https://build.nvidia.com/nvidia/cosmos3-nano/modelcard}{\textcolor{black}{\rwYes}} & \rwYes & \rwYes & \rwYes & \rwPartial & \rwPartial & Py $\cdot$ CLI $\cdot$ Web \\
    \href{https://github.com/AgibotTech/Genie-Envisioner-V1}{Genie Envisioner}~\citep{wfmodel_genieenvisioner,ge_sim2} & \rwPartial & \rwNo & \rwYes & \rwYes & \rwPartial & \rwNo & \rwYes & \rwYes & \rwPartial & \rwPartial & \rwPartial & Py $\cdot$ CLI\newline Web (monitor) \\
    \href{https://github.com/shengshu-ai/minWM}{minWM}~\citep{wfmodel_minwmhyaction2v} & \rwPartial & \rwNo & \rwYes & \rwNo & \rwPartial & \rwNo & \rwNo & \rwYes & \rwNo & \rwPartial & \rwPartial & Py $\cdot$ CLI \\
    \rwGroup{MODALITY-SPECIFIC TOOLKITS}
    \href{https://github.com/huggingface/diffusers}{Diffusers}~\citep{diffusers} & \rwYes & \rwPartial & \rwPartial & \rwPartial & \rwYes & \rwNo & \rwNo & \rwYes & \rwPartial & \rwPartial & \rwNo & Py \\
    \href{https://github.com/huggingface/lerobot}{LeRobot}~\citep{lerobot} & \rwPartial & \rwNo & \rwYes & \rwYes & \rwYes & \rwNo & \rwYes & \rwYes & \rwPartial & \rwPartial & \rwNo & Py $\cdot$ CLI \\
    \href{https://github.com/starVLA/starVLA}{StarVLA}~\citep{starvla} & \rwNo & \rwNo & \rwNo & \rwYes & \rwPartial & \rwNo & \rwYes & \rwYes & \rwPartial & \rwPartial & \rwNo & Py $\cdot$ CLI \\
    \rwGroup{EVALUATION SUITES}
    \href{https://github.com/Vchitect/VBench}{VBench}~\citep{vbench,wfb_vbenchplusplus,vbench2} & \rwYes & \rwNo & \rwNo & \rwNo & \rwNo & \rwNo & \rwNo & \rwNo & \rwPartial & \rwPartial & \rwNo & Py $\cdot$ CLI $\cdot$ LB \\
    \href{https://github.com/haoyi-duan/WorldScore}{WorldScore}~\citep{worldscore} & \rwYes & \rwYes & \rwNo & \rwNo & \rwPartial & \rwPartial & \rwNo & \rwNo & \rwPartial & \rwPartial & \rwPartial & Py $\cdot$ CLI $\cdot$ LB \\
    \href{https://github.com/WorldArena2/WorldArena-2.0}{WorldArena}~\citep{worldarena,worldarena2} & \rwPartial & \rwNo & \rwPartial & \rwYes & \rwPartial & \rwNo & \rwYes & \rwPartial & \rwPartial & \rwPartial & \rwPartial & Py $\cdot$ CLI $\cdot$ LB \\
    \href{https://github.com/World-In-World/world-in-world}{World-in-World}~\citep{wfb_worldinworld} & \rwPartial & \rwNo & \rwPartial & \rwYes & \rwPartial & \rwPartial & \rwYes & \rwPartial & \rwPartial & \rwPartial & \rwPartial & Py $\cdot$ CLI \\
    \midrule
    \rowcolor{tabhead!45}
    \textbf{WorldFoundry} & \rwYes & \rwYes & \rwYes & \rwYes & \rwYes & \rwYes & \rwYes & \rwNo & \rwYes & \rwYes & \rwYes & TUI $\cdot$ CLI $\cdot$ Web\newline Py $\cdot$ MCP \\
    \bottomrule
    \end{tabularx}
    \par\smallskip
    \begin{minipage}{\linewidth}
    \fontsize{7.5}{9}\selectfont
    \textcolor{oered}{\textbf{Reading the table.}}
    \emph{Coverage}: \rwPartial\ restricted task or domain; Inter.\ WM requires action-conditioned dynamics, not camera-path rendering alone.
    \emph{Model runner}: \rwYes\ independently released families behind shared interfaces; \rwPartial\ first-party models or benchmark-specific adapters.
    \emph{Hosted API}: provider-hosted generation or policy services.
    \emph{Closed loop}: controls depend on prior outputs or environment feedback.
    \emph{Training}: \rwPartial\ protocol-specific recipe only.
    \emph{Protocols}: \rwYes\ independent benchmarks across task families; \rwPartial\ a single suite or domain.
    \emph{Evidence}: \rwYes\ links entry readiness to result provenance; \rwPartial\ stores configurations or scores only.
    \emph{WM def.}: \rwYes\ framework-level criterion and taxonomy; \rwPartial\ domain-specific formulation.
    VBench, WorldArena, and Genie Envisioner rows include VBench++/2.0, WorldArena~2.0, and GE-Sim~2.0.
    LB: leaderboard; MCP: Model Context Protocol.
    Marks describe documented workflows; entry-level readiness and validation scope are reported in the appendices. Validation applies to the documented route and configuration.
    \end{minipage}
    \endgroup
\end{table}
\subsection{World Models and Research Infrastructure}

Learned world models link environmental prediction to planning and control. World Models~\citep{worldmodels} combined compact visual representations with recurrent dynamics and policy learning; PlaNet~\citep{planet} planned through latent dynamics, while DreamerV3~\citep{dreamerv3} learned policies from imagined trajectories. Research on video generation, spatial reconstruction, and embodied control introduces diverse conditioning schemes, state representations, and execution requirements. Specialized toolkits address these requirements within individual domains: Diffusers~\citep{diffusers} organizes generative pipelines, Nerfstudio~\citep{nerfstudio} supports neural reconstruction and rendering, LeRobot~\citep{lerobot} and StarVLA~\citep{starvla} support policy development and deployment, and Isaac Lab~\citep{isaaclab} provides simulation and learning workflows. World-model platforms cover broader development lifecycles. Cosmos~\citep{cosmos,cosmos_software} combines data preparation, foundation models, and post-training, complemented by scoring orchestration and durable result storage in Cosmos Evaluator~\citep{cosmos_evaluator}. Genie Envisioner and GE-Sim~2.0~\citep{wfmodel_genieenvisioner,ge_sim2} connect action-conditioned video simulation with robot-policy evaluation and learning, while minWM~\citep{wfmodel_minwmhyaction2v} supports fine-tuning and distillation for controllable autoregressive generation. OpenWorldLib~\citep{openworldlib,openworldlib_software} is the most closely related infrastructure for heterogeneous integration: its pipeline, operator, representation, reasoning, and memory modules unify video, spatial, and embodied systems through common interfaces, supporting multi-turn execution and evaluation from saved results. Its formulation emphasizes perception, interaction, and long-term memory. WorldFoundry couples an operational criterion grounded in persistent state, intervention effects, and observational feedback (\S\ref{subsec:definition}--\ref{subsec:capability}) with explicit execution contracts: manifests specify runtime and asset requirements, serializable requests preserve native controls, and typed artifact references link outputs to execution provenance. These contracts connect model invocation, artifact management, and evaluation within a shared experimental lifecycle; general training remains future work.

\subsection{Evaluation Protocols and Experimental Evidence}

Evaluation suites operationalize complementary notions of generative quality, environmental consistency, and interactive utility. VBench and its extensions~\citep{vbench,wfb_vbenchplusplus,vbench2} provide a multidimensional assessment of perceptual quality, semantic alignment, and intrinsic faithfulness, while VideoPhy~\citep{wfb_videophy} and WorldModelBench~\citep{worldmodelbench} examine whether generated events and interactions satisfy physical constraints and commonsense expectations. WorldScore~\citep{worldscore} evaluates video and spatial generation under camera-conditioned protocols, connecting assessment to specified controls over the generated scene. iWorld-Bench~\citep{iworldbench} focuses on interactive control and memory, where model responses must be interpreted in relation to preceding inputs and interactions. WorldArena~\citep{worldarena,worldarena2} and World-in-World~\citep{wfb_worldinworld} extend assessment to embodied utility, policy performance, and closed-loop planning, linking world-model capabilities to their effectiveness within agent workflows. Evaluation infrastructure supports these assessments through complementary mechanisms: AIGVE-Tool~\citep{aigve_tool} provides modular scoring components, and HarnessEval-W~\citep{wfb_harnessevalw} organizes judgments around structured supporting evidence. The evidence required for evaluation consequently varies across protocols. Some assessments can operate on fixed generated artifacts, whereas interactive and embodied tasks may require control histories, environment feedback, and executed rollouts. Differences in these requirements and in score aggregation make execution conditions and protocol compliance central to meaningful comparison. WorldFoundry records resolved configurations, artifact references, sample coverage, failures, and validation scope for each evaluation (\S\ref{sec:evaluation_evidence}). Its scorecards distinguish diagnostic results, benchmark-comparable results, and complete reproductions under the corresponding protocol. Compatible evaluators can rescore saved outputs; protocols requiring new interventions or policy rollouts retain their execution requirements. These records make evaluative claims traceable to their supporting artifacts, execution conditions, and validation evidence.

\section{Unified Definition of World Modeling}
\label{sec:taxonomy}
\begin{figure}[H]
    \centering
    \includegraphics[width=\linewidth]{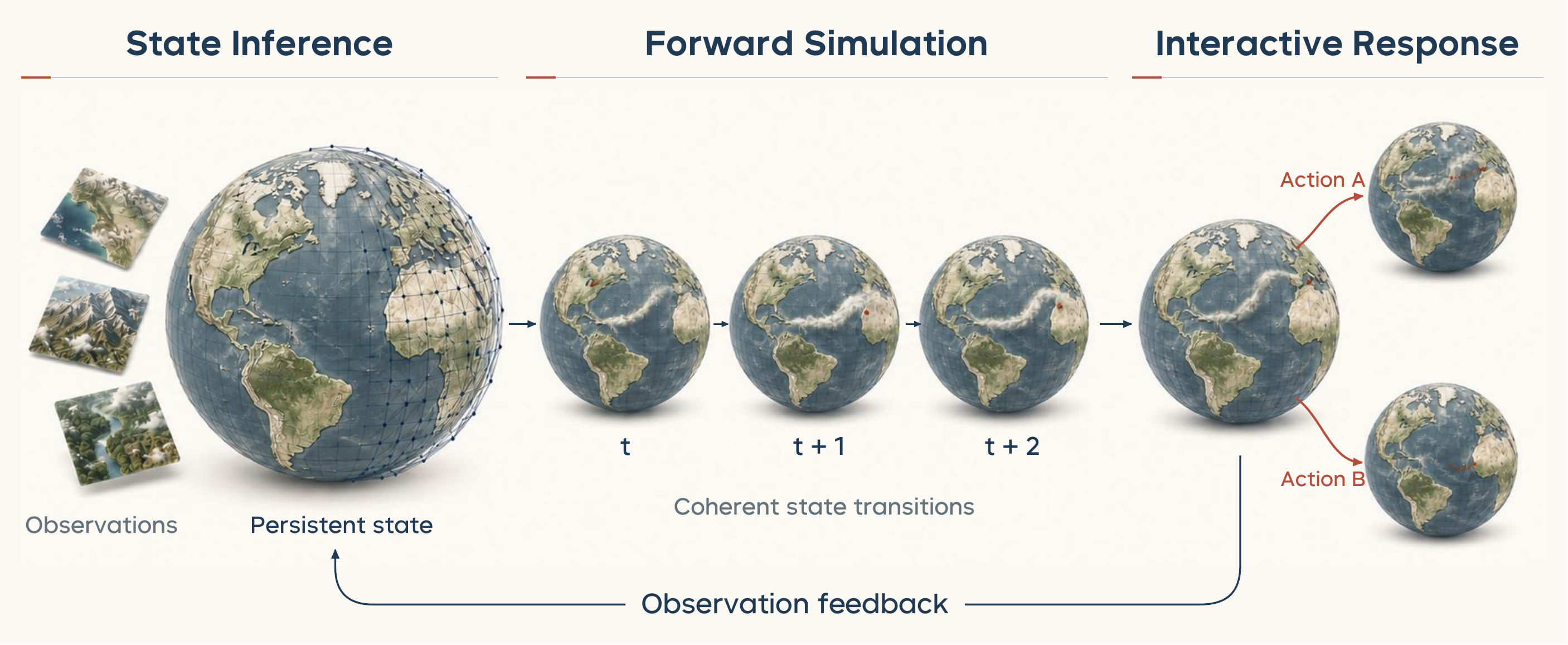}
    \caption{\textbf{World Modeling.}
    State inference maintains a persistent environment representation from observations; forward simulation predicts coherent state transitions; interactive response produces action-conditioned futures. Observation feedback updates the state for subsequent predictions.}
    \label{fig:world-model-definition}
\end{figure}
A fundamental obstacle to progress in generative world modeling research is the absence of a precise and shared conceptual foundation. Without an operational definition of what constitutes a world model and a principled criterion for distinguishing it from related systems, the field risks conflating fundamentally different modeling objectives. This section aims to address this gap. We first establish an operational definition of world modeling; we then situate the model landscape along an intelligence-oriented capability progression that traces increasingly demanding functional behaviors; finally, we formulate a principled boundary between foundational generative models that synthesize observations and generative world models that simulate world.

\subsection{What is World Model: An Operational Definition}
\label{subsec:definition}

%we adopt a functional definition tailored to the generative and interactive setting considered here.

%Motivated by previous work on learned world models in reinforcement learning~\citep{worldmodels,planet,dreamerv3} and recent efforts to delimit their scope~\citep{openworldlib}, we adopt the following definition. A world model is a computational system that learns an internal representation of an environment and models how that environment evolves over time and under intervention. Rather than modeling observations alone, it represents task-relevant latent state and the transition processes that generate future observations. This requires three tightly coupled operations: 

Motivated by previous work on learned world models in reinforcement learning~\citep{worldmodels,planet,dreamerv3}, we characterize world modeling by three tightly coupled operations over task-relevant latent state rather than observations alone: (i) \textbf{state inference}, which estimates latent environmental state from observations; (ii) \textbf{forward simulation}, which predicts state evolution under autonomous dynamics, physical constraints, and external actions; and (iii) \textbf{interactive response}, which adapts predicted futures coherently under alternative actions or interventions. Let $o_t$ denote an observation, $c_t$ an action or intervention, and $h_t=(o_{\leq t},c_{<t})$ the interaction history. A minimal, modality-agnostic abstraction is

\begin{equation}
\label{eq:world-model}
    z_t \sim q(z_t\mid h_t),
    \qquad
    z_{t+1} \sim p(z_{t+1}\mid z_t,c_t),
    \qquad
    \hat{o}_{t+1} \sim g(\cdot\mid z_{t+1}).
\end{equation}

Here $z_t$ denotes a task-relevant environmental state, while $q$, $p$, and $g$ denote state inference, transition, and observation models. Repeated application of $p$ yields predictive rollouts under alternative control sequences; once $o_{t+1}$ is observed, the state is re-inferred as $z_{t+1}\sim q(\cdot\mid h_{t+1})$ before subsequent predictions, closing the loop between prediction and feedback, following the latent-dynamics perspective used in model-based planning~\citep{planet}. This formulation is functional rather than architectural: $z_t$ may be recurrent, object-centric, geometric, token-based, or otherwise structured. Coupling these generic world-modeling operations into a single persistent process likewise yields our foundational definition of generative world models:

\begin{definitionbox}
\textbf{A Generative World Model} is a learned, history- and control-conditioned latent dynamical system that infers and maintains a persistent environmental state, evolves that state under endogenous dynamics and external interventions, and assimilates feedback to generate coherent future trajectories.
\end{definitionbox}

%Under this definition, world modeling requires more than high-fidelity observation synthesis. A model must retain task-relevant history, predict coherent trajectories from a shared state under alternative interventions, carry their consequences forward, and revise its state as new observations become available. The defining property is the functional coupling of persistent state, intervention-conditioned dynamics, and feedback-driven updating. These requirements motivate the capability progression developed next, which identifies the constituent capabilities of world modeling and clarifies how their coordination supports coherent prediction and interaction.
With this definition, a generative world model synthesizes possible future trajectories of observations grounded in a persistent, history-conditioned environmental state. Generation under alternative interventions branches from a shared state, with the consequences of each intervention constraining subsequent outputs. Newly acquired observations update the maintained state and inform further generation. The defining property is therefore the coupling of observation synthesis with persistent state, intervention-conditioned dynamics, and observational feedback. These requirements motivate the capability progression below, which characterizes how constituent capabilities jointly support coherent world generation and interaction.

\subsection{What World Models Can Do: Capability Progression}
\label{subsec:capability}
\begin{figure}[H]
    \centering
    \includegraphics[width=\linewidth]{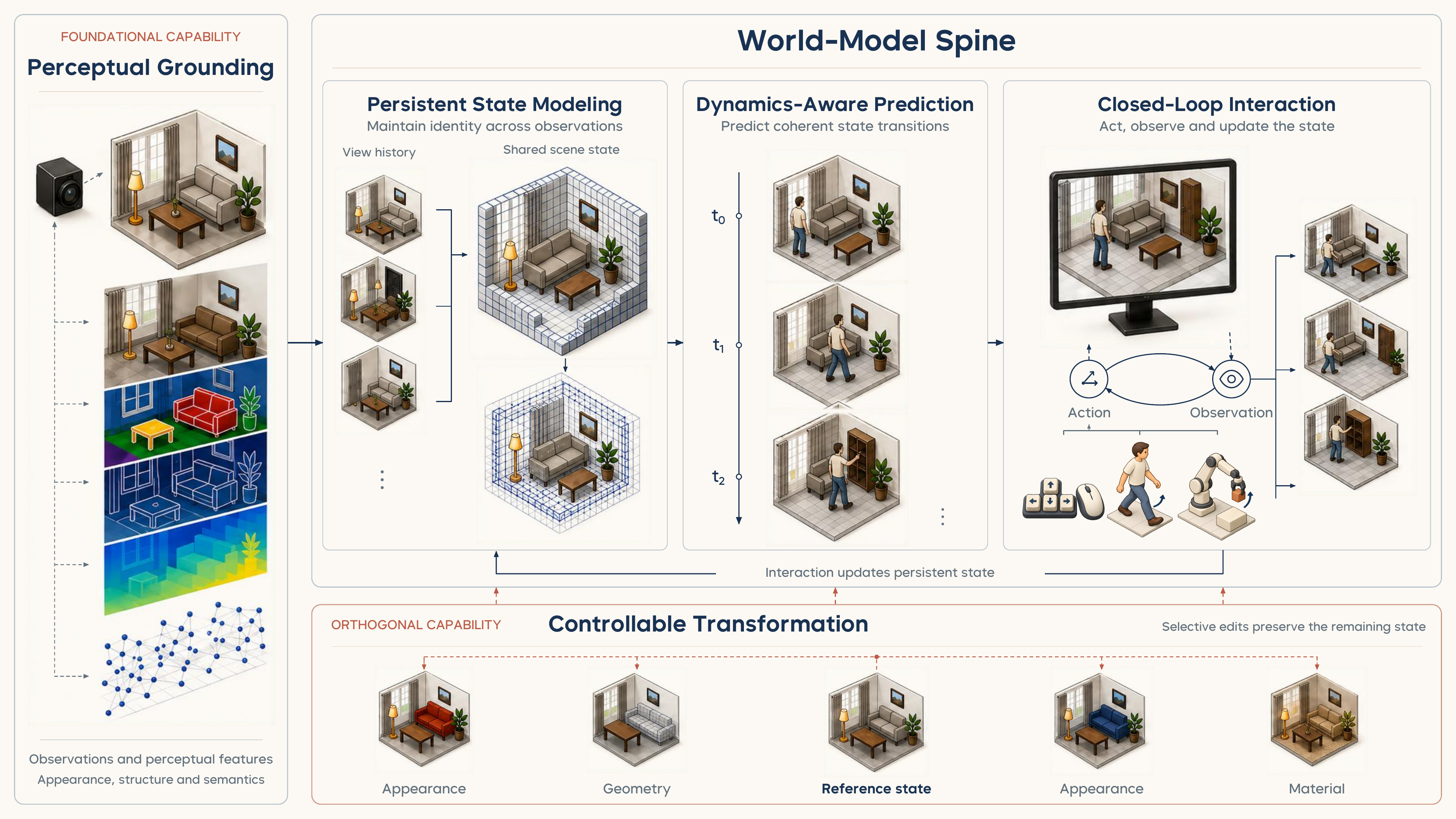}
    \caption{\textbf{Capability progression from generative modeling toward world modeling}. Perceptual grounding provides the foundation; persistent state modeling, dynamics-aware prediction, and closed-loop interaction form the world-model spine; controllable transformation provides an orthogonal capability for manipulation.}
    \label{fig:capacity-level}
\end{figure}

To situate existing generative systems with respect to this definition, we organize them along a capability progression in which generation becomes increasingly grounded in an evolving environmental state. \textbf{Perceptual Grounding} constitutes the foundation: the capability to interpret, reconstruct, or synthesize observations without maintaining a persistent environmental state. The \textbf{World-Model Spine} comprises three successive stages, each building on its predecessor: \textbf{Persistent State Modeling} infers and maintains a structured environmental state across observations; \textbf{Dynamics-Aware Prediction} predicts physically and causally coherent transitions of that state under endogenous dynamics and external interventions; and \textbf{Closed-Loop Interaction} sustains repeated cycles of action, observation, and state update, in which each observation revises the state that conditions subsequent predictions. \textbf{Controllable Transformation}, the capability to modify represented entities, attributes, or configurations as specified while preserving content outside the scope of the edit, is orthogonal to the spine and can complement any of its stages; by itself, however, it entails none of the spine capabilities. Rather than assigning each system to a single discrete level, the progression characterizes it by a capability profile: which capabilities it exhibits and the degree to which they are integrated.
\vspace{1em}

\begin{tcolorbox}[
    enhanced,
    colback=blue!3,
    colframe=blue!55!black,
    boxrule=0.7pt,
    arc=2mm,
    outer arc=2mm,
    left=8pt,
    right=8pt,
    top=8pt,
    bottom=7pt,
    before skip=9pt,
    after skip=7pt,
    fontupper=\small,
    fonttitle=\bfseries,
    coltitle=black,
    title={Foundational Capability: Perceptual Grounding},
    attach boxed title to top left={
        xshift=8pt,
        yshift=-2mm
    },
    boxed title style={
        colback=blue!10,
        colframe=blue!55!black,
        boxrule=0.6pt,
        arc=1.5mm,
        outer arc=1.5mm,
        left=5pt,
        right=5pt,
        top=1.5pt,
        bottom=1.5pt
    }
]
\textbf{Perceptual Grounding (Perception)} is the capability to interpret, reconstruct, or synthesize observable content. It captures structure at the observation level, supplying
the perceptual basis for state inference without itself maintaining a persistent representation of the underlying environment.
\end{tcolorbox}

\begin{tcolorbox}[
    enhanced,
    colback=red!2,
    colframe=red!55!black,
    boxrule=0.7pt,
    arc=2mm,
    outer arc=2mm,
    left=7pt,
    right=7pt,
    top=11pt,
    bottom=8pt,
    before skip=9pt,
    after skip=7pt,
    fonttitle=\bfseries,
    coltitle=black,
    title={World-Model Spine},
    attach boxed title to top left={
        xshift=8pt,
        yshift=-2mm
    },
    boxed title style={
        colback=red!8,
        colframe=red!55!black,
        boxrule=0.6pt,
        arc=1.5mm,
        outer arc=1.5mm,
        left=5pt,
        right=5pt,
        top=1.5pt,
        bottom=1.5pt
    }
]

\begin{center}
{\footnotesize\itshape\color{red!45!black}
Persistent State
\(\rightarrow\)
Dynamics Prediction
\(\rightarrow\)
Interactive Feedback}
\end{center}

\vspace{-2pt}

% ------------------------------------------------------------
% Spine I
% ------------------------------------------------------------

\begin{tcolorbox}[
    enhanced,
    colback=white,
    colframe=red!45!black,
    boxrule=0.7pt,
    arc=2mm,
    outer arc=2mm,
    left=8pt,
    right=8pt,
    top=8pt,
    bottom=7pt,
    before skip=9pt,
    after skip=6pt,
    fontupper=\small,
    fonttitle=\bfseries,
    coltitle=black,
    title={Spine I: Persistent State Modeling},
    attach boxed title to top left={
        xshift=8pt,
        yshift=-2mm
    },
    boxed title style={
        colback=red!5,
        colframe=red!45!black,
        boxrule=0.6pt,
        arc=1.5mm,
        outer arc=1.5mm,
        left=5pt,
        right=5pt,
        top=1.5pt,
        bottom=1.5pt
    }
]
\textbf{Persistent State Modeling (Modeling)} is the capability to infer, maintain, and recursively update a structured representation of environmental state from sequential observations. This representation tracks task-relevant entities and their identities, geometry, spatial relations, and latent attributes across viewpoint changes and occlusions under partial observability.
\end{tcolorbox}

\begin{center}
\color{red!45!black}
\footnotesize
\(\downarrow\)
\end{center}

% ------------------------------------------------------------
% Spine II
% ------------------------------------------------------------

\begin{tcolorbox}[
    enhanced,
    colback=white,
    colframe=red!45!black,
    boxrule=0.7pt,
    arc=2mm,
    outer arc=2mm,
    left=8pt,
    right=8pt,
    top=8pt,
    bottom=7pt,
    before skip=9pt,
    after skip=6pt,
    fontupper=\small,
    fonttitle=\bfseries,
    coltitle=black,
    title={Spine II: Dynamics-Aware Prediction},
    attach boxed title to top left={
        xshift=8pt,
        yshift=-2mm
    },
    boxed title style={
        colback=red!5,
        colframe=red!45!black,
        boxrule=0.6pt,
        arc=1.5mm,
        outer arc=1.5mm,
        left=5pt,
        right=5pt,
        top=1.5pt,
        bottom=1.5pt
    }
]
\textbf{Dynamics-Aware Prediction (Dynamics)} is the capability to predict environmental state transitions in accordance with physical and causal regularities. It propagates the effects of endogenous processes, entity interactions, and external interventions through the evolving state, producing temporally coherent trajectories in which earlier events constrain later predictions through their persisting consequences.
\end{tcolorbox}

\begin{center}
\color{red!45!black}
\footnotesize
\(\downarrow\)
\end{center}

% ------------------------------------------------------------
% Spine III
% ------------------------------------------------------------

\begin{tcolorbox}[
    enhanced,
    colback=white,
    colframe=red!45!black,
    boxrule=0.7pt,
    arc=2mm,
    outer arc=2mm,
    left=8pt,
    right=8pt,
    top=8pt,
    bottom=7pt,
    before skip=9pt,
    after skip=6pt,
    fontupper=\small,
    fonttitle=\bfseries,
    coltitle=black,
    title={Spine III: Closed-Loop Interaction},
    attach boxed title to top left={
        xshift=8pt,
        yshift=-2mm
    },
    boxed title style={
        colback=red!5,
        colframe=red!45!black,
        boxrule=0.6pt,
        arc=1.5mm,
        outer arc=1.5mm,
        left=5pt,
        right=5pt,
        top=1.5pt,
        bottom=1.5pt
    }
]
\textbf{Closed-Loop Interaction (Interaction)} is the capability to sustain recurrent cycles of action, environmental transition, observation, and state update within a persistent environment. Feedback from observed action outcomes revises the maintained state, which in turn conditions subsequent predictions and actions, preserving causal continuity across successive interactions.
\end{tcolorbox}

\end{tcolorbox}

\begin{tcolorbox}[
    enhanced,
    colback=orange!3,
    colframe=orange!55!black,
    boxrule=0.7pt,
    arc=2mm,
    outer arc=2mm,
    left=8pt,
    right=8pt,
    top=8pt,
    bottom=7pt,
    before skip=9pt,
    after skip=7pt,
    fontupper=\small,
    fonttitle=\bfseries,
    coltitle=black,
    title={Orthogonal Capability: Controllable Transformation},
    attach boxed title to top left={
        xshift=8pt,
        yshift=-2mm
    },
    boxed title style={
        colback=orange!9,
        colframe=orange!55!black,
        boxrule=0.6pt,
        arc=1.5mm,
        outer arc=1.5mm,
        left=5pt,
        right=5pt,
        top=1.5pt,
        bottom=1.5pt
    }
]
\textbf{Controllable Transformation (Manipulation)} is the capability to modify represented content or state according to specified controls while preserving properties and structural relations outside the intended scope of the edit. As an orthogonal capability, it can complement any stage of the progression, but on its own establishes neither persistent state modeling nor coherent dynamics or closed-loop interaction.
\end{tcolorbox}

\begin{center}
\begin{minipage}{0.94\linewidth}
\centering
\small\itshape\color{oecapink!62}
%Perceptual Grounding supplies the observational basis; Persistent State Modeling, Dynamics-Aware Prediction, and Closed-Loop Interaction form the world-model spine; and Controllable Transformation remains orthogonal to that progression.
\end{minipage}
\end{center}

Taken together, these capabilities characterize world intelligence through their integration rather than their individual strength: observations update persistent state,
state constrains predictions, and action outcomes inform subsequent inference and interaction. Current systems occupy this progression unevenly. Video generators achieve strong perceptual grounding with largely implicit state; 3D and 4D methods maintain explicit spatial structure but rarely predict its evolution under actions; interactive world models close the loop over short horizons and have begun to address longer-range memory; and embodied policies act within persistent environments under narrow task and embodiment coverage. Evaluating where a system sits therefore requires evidence about the cycle itself, not only about output quality. Existing benchmarks test complementary segments of this progression: visual and physical fidelity~\citep{vbench2,worldmodelbench}, camera-controlled world generation~\citep{worldscore}, interactive memory~\citep{iworldbench}, and downstream embodied utility~\citep{worldarena,worldarena2}.

\subsection{Taxonomy: Foundational Generative Models and Generative World Models}
\label{subsec:taxonomy_classes}
%Building on the definition and capability progression above, we draw a single boundary through contemporary generative systems: whether a system closes the world-modeling loop. Models that primarily generate or transform observations count as foundational generative models, even with partial, local, or implicit dynamics, unless they maintain persistent state, support action-conditioned dynamics, and produce counterfactually valid rollouts. Generative world models, by contrast, meet all three conditions, maintaining and evolving latent world states for prediction, intervention, and interaction.
Building on the definition and capability progression above, we draw a single boundary through contemporary generative systems: whether a system closes the world-modeling loop. Models that primarily generate or transform observations count as foundational generative models, even with partial, local, or implicit dynamics, unless they maintain persistent state, support intervention-conditioned dynamics, and assimilate observational feedback. Generative world models, by contrast, meet all three conditions: they maintain latent world states that evolve under intervention and are revised by incoming observations.

%Building on the definition and the capability progression introduced above, we draw a single boundary through contemporary generative system: whether a system closes the world-modeling loop. Models that primarily generate or transform observations are classified as foundational generative models, even when they exhibit partial, local, or implicit dynamics, unless they maintain persistent state, support action-conditioned dynamics, and produce counterfactually valid rollouts. By contrast, generative world models are systems that meet all three conditions and therefore maintain and evolve latent world states for prediction, intervention, and interaction.

\textbf{Foundational Generative Models} learn visual, geometric, or spatiotemporal priors for synthesis, reconstruction, and controllable manipulation, but do not integrate state, dynamics, and feedback into a complete world-modeling loop. Their defining limitation is the inability to maintain an evolving environmental state that preserves intervention effects and incorporates observational feedback into subsequent predictions. Strong individual capabilities therefore do not suffice to sustain a coherent, revisable model of the world.

\textbf{Generative World Models}, by contrast, integrate persistent state, intervention-conditioned dynamics, and observational feedback into a complete world-modeling loop. Their defining property is functional coupling: intervention effects persist across state transitions, while each new observation revises the state governing subsequent predictions. Generation is thereby made accountable to an evolving environmental model, supplying a functional basis for world intelligence, the capacity to anticipate the consequences of intervention and to revise those expectations through interaction.

\section{WorldFoundry Design}
\label{sec:framework}

Assessing the capabilities defined above requires reproducible experiments that tie observed behavior to explicitly specified tasks and evaluation conditions. Existing implementations, however, rely on incompatible environments, input conventions, execution procedures, artifact formats, and evaluation protocols, making it difficult to tell whether observed differences arise from model capabilities or from the surrounding experimental pipelines. This fragmentation also blurs the distinctions among catalog inclusion, successful inference, partial validation, and benchmark reproduction, which are often conflated under a single notion of ``model support.'' Related infrastructures address complementary aspects of this problem, including OpenWorldLib for world-model inference~\citep{openworldlib}, Cosmos for physical-AI model development~\citep{cosmos}, and StarVLA for modular embodied-policy research~\citep{starvla}. Common to all of these settings is a central design challenge: preserving model-specific execution semantics while making explicit what has been executed, under what conditions, and which conclusions the resulting evidence supports.

We introduce \textbf{\textsc{WorldFoundry}} as open-source infrastructure organized around a common lifecycle of discovery, runtime preparation, execution, inspection, and evaluation. It serves three functions. \textbf{(I)} It integrates declarative model and benchmark catalogs with explicit runtime, dependency, and asset management, allowing researchers to discover, configure, and execute foundational generative models for video, 3D, and 4D alongside generative world models, embodied simulators, and vision-language-action, vision-action, and world-action models, while preserving native execution semantics. \textbf{(II)} It decouples execution from evaluation through durable artifacts and serializable request and result contracts, so that generated videos, spatial representations, actions, trajectories, and execution traces can be inspected, reused, compared, and scored without re-instantiating the originating model or runtime; run manifests, sample results, metric summaries, reports, and scorecards record the provenance and evidential scope of each experiment. \textbf{(III)} It provides interactive interfaces for configuring runs, issuing control inputs, visualizing multimodal outputs, comparing results across model families, and diagnosing failures throughout generation, interaction, and evaluation.

\subsection{System Overview}

\textbf{WorldFoundry} is organized around a user-oriented, capability-aware research lifecycle that brings heterogeneous visual generation and world-modeling systems into a coherent operational framework. Rather than requiring users to navigate fragmented, model-specific inference scripts and disconnected evaluation pipelines, it establishes shared conventions for model discovery, runtime preparation, execution, artifact inspection, and evidence-based evaluation while preserving model-specific semantics. The framework comprises five components (Figure~\ref{fig:overview}), each described in a subsequent subsection. \textbf{(1)} \emph{Surface Layer} provides consistent user-facing access through a terminal interface, command-line tools, a browser workspace, a Python API, and programmatic interfaces. \textbf{(2)} \emph{Unified Inference and Interaction} resolves model configurations, runtime profiles, and required assets, and supports both stateless generation and stateful interaction through shared request and execution pathways without altering model-native inputs, parameters, or execution semantics. \textbf{(3)} \emph{Artifact and Representation Contracts} specify how requests, results, generated outputs, state traces, and execution records are materialized as durable, inspectable artifacts with explicit identity, semantics, metadata, and provenance. \textbf{(4)} \emph{Evaluation and Evidence} applies compatible metrics and benchmark protocols to these artifacts and records per-sample results, aggregate scores, coverage, validation status, unresolved blockers, and comparison eligibility. \textbf{(5)} \emph{Model and Benchmark Zoo} maintains manifest-driven registries of models, benchmarks, tasks, runtime profiles, assets, evaluators, readiness states, and known blockers.

Across these components, persistent artifacts form the explicit boundary between execution and assessment. This separation allows computationally expensive outputs to be inspected, compared, rescored, and audited without re-executing the originating model, and keeps catalog inclusion, runnable integration, bounded validation, and complete benchmark reproduction as distinct claims rather than a single notion of support.

\subsection{Surface Layer}
The Surface Layer offers multiple user-facing interfaces that share a single orchestration path, ensuring consistent execution and evaluation across access modes. The terminal user interface (TUI) guides discovery by querying model and benchmark catalogs, checking requirements and readiness, and resolving an executable command before resources are allocated. The command-line interface (CLI) covers discovery, preparation, inference, evaluation, validation, and reporting, and is designed for automated experimentation and reproducible benchmarking. The browser-based Workspace supports multi-model job configuration, model-specific input specification, runtime selection, execution monitoring, artifact inspection, and cross-run comparison, while the single-model Studio offers tailored controls for interactive use and detailed inspection. Python and MCP interfaces extend this functionality to external programs and agents. Every surface translates user intent into normalized requests that are resolved against shared catalogs, runtime profiles, Pipelines, Operators, and result contracts; this shared resolution keeps results comparable regardless of how a run is launched. Figures~\ref{fig:surface_cli}--\ref{fig:surface_tui} illustrate the complementary roles of the three primary human-facing surfaces.
\vspace{-0.25em}

\begin{center}
\begin{tcolorbox}[
    enhanced,
    width=\linewidth,
    colback=white,
    colframe=oepanelborder,
    colbacktitle=oeaccent,
    coltitle=oeredark,
    boxrule=0.7pt,
    arc=2mm,
    outer arc=2mm,
    boxsep=0pt,
    left=5pt,
    right=5pt,
    top=5pt,
    bottom=5pt,
    toptitle=4pt,
    bottomtitle=4pt,
    fonttitle=\sffamily\bfseries\fontsize{8.6}{10}\selectfont,
    title={
        \textcolor{oered}{\rule{2.2pt}{1.55ex}}
        \hspace{5pt}WORLDFOUNDRY CLI
        \hfill
        \textcolor{oemuted}{DISCOVER, VERIFY, EXECUTE, AND REPORT}
    }
]
    \begin{tcolorbox}[
        width=\linewidth,
        colback=oeink,
        colframe=oeline,
        boxrule=0.5pt,
        arc=1mm,
        boxsep=0pt,
        left=1pt,
        right=1pt,
        top=1pt,
        bottom=1pt
    ]
        \centering
        \includegraphics[width=0.60\linewidth]{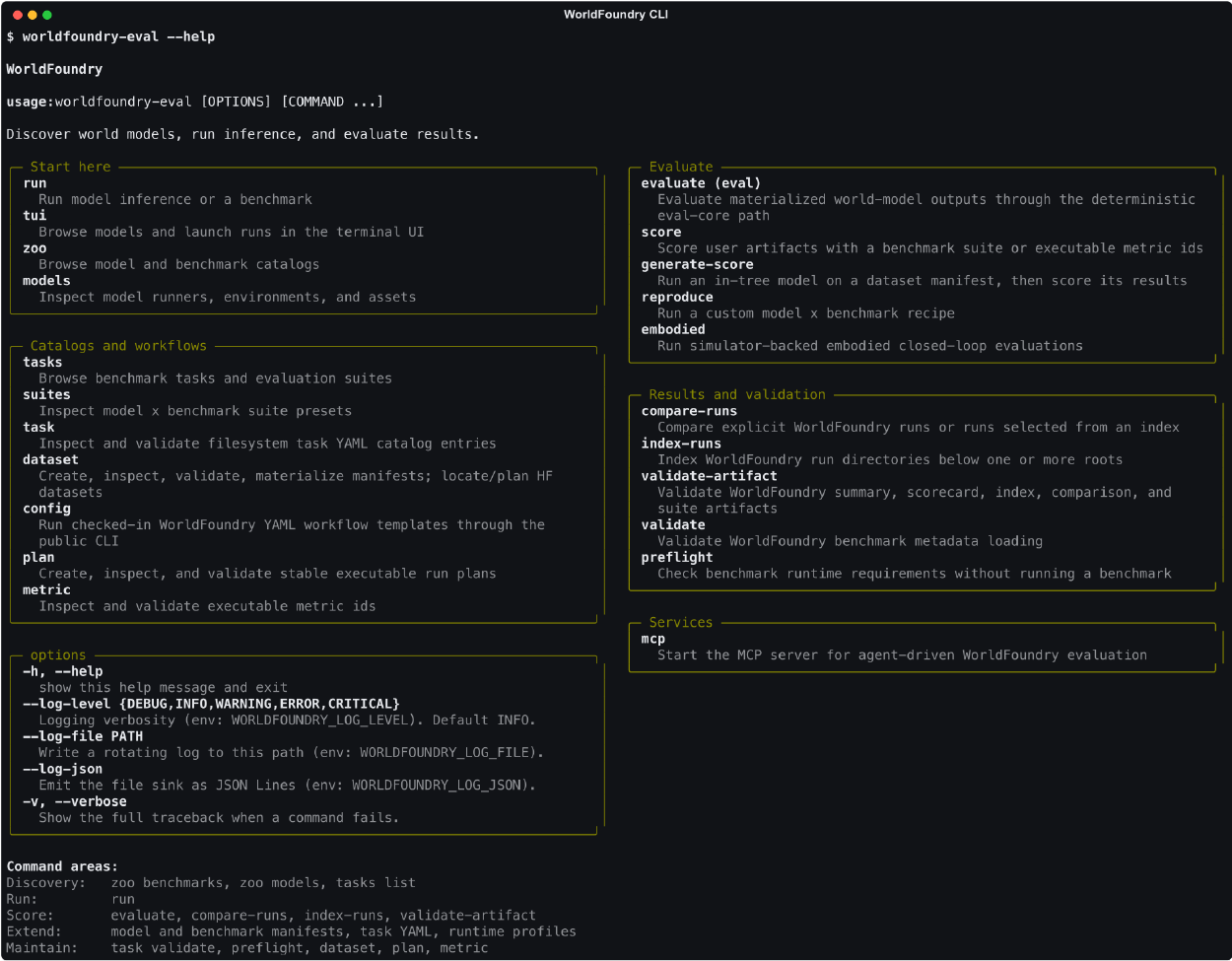}
        \par
    \end{tcolorbox}
\end{tcolorbox}

\captionof{figure}{
    \textbf{Command-line interface (CLI).}
    WorldFoundry CLI is a unified command-line interface for configuring
    and executing model inference and evaluation workflows.
    Screenshot from the
    \href{https://openenvision.github.io/WorldFoundry/docs/reference/cli}
    {CLI reference}.
}
\label{fig:surface_cli}
\end{center}

% ============================================================
% WORKSPACE / STUDIO
% ============================================================
\begin{center}
\begin{tcolorbox}[
    enhanced,
    width=0.98\linewidth,
    colback=white,
    colframe=oepanelborder,
    colbacktitle=oeaccent,
    coltitle=oeredark,
    boxrule=0.7pt,
    arc=2mm,
    outer arc=2mm,
    boxsep=0pt,
    left=5pt,
    right=5pt,
    top=5pt,
    bottom=5pt,
    toptitle=4pt,
    bottomtitle=4pt,
    fonttitle=\sffamily\bfseries\fontsize{8.6}{10}\selectfont,
    title={
        \textcolor{oered}{\rule{2.2pt}{1.55ex}}
        \hspace{5pt}WORLDFOUNDRY WORKSPACE / STUDIO
        \hfill
        \textcolor{oemuted}{MULTI-MODEL VISUAL OPERATION}
    }
]
    \begin{tcolorbox}[
        width=\linewidth,
        colback=oeink,
        colframe=oeline,
        boxrule=0.5pt,
        arc=1mm,
        boxsep=0pt,
        left=1pt,
        right=1pt,
        top=1pt,
        bottom=1pt,
        before skip=0pt,
        after skip=4pt
    ]
        \centering
        \includegraphics[width=0.85\linewidth]{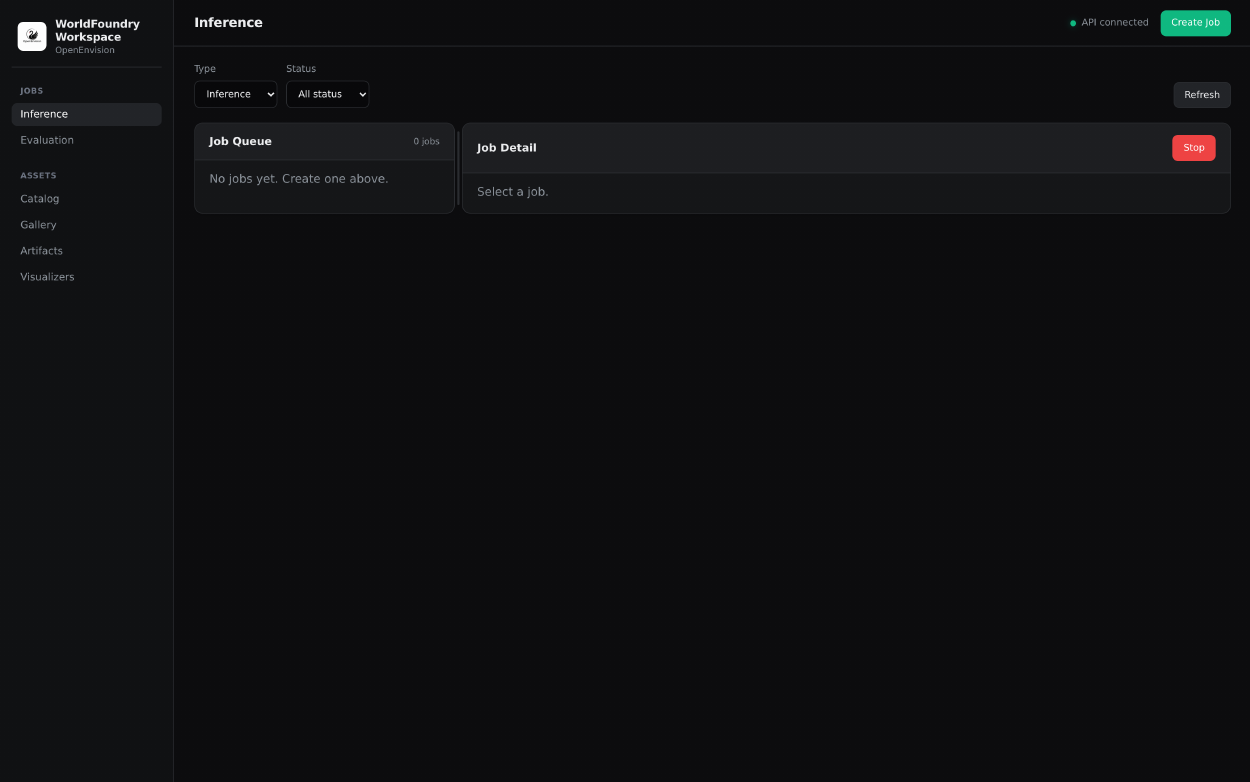}
        \par
    \end{tcolorbox}

    \begin{tcolorbox}[
        width=\linewidth,
        colback=oeink,
        colframe=oeline,
        boxrule=0.5pt,
        arc=1mm,
        boxsep=0pt,
        left=1pt,
        right=1pt,
        top=1pt,
        bottom=1pt,
        before skip=0pt,
        after skip=0pt
    ]
        \centering
        \includegraphics[width=0.85\linewidth]{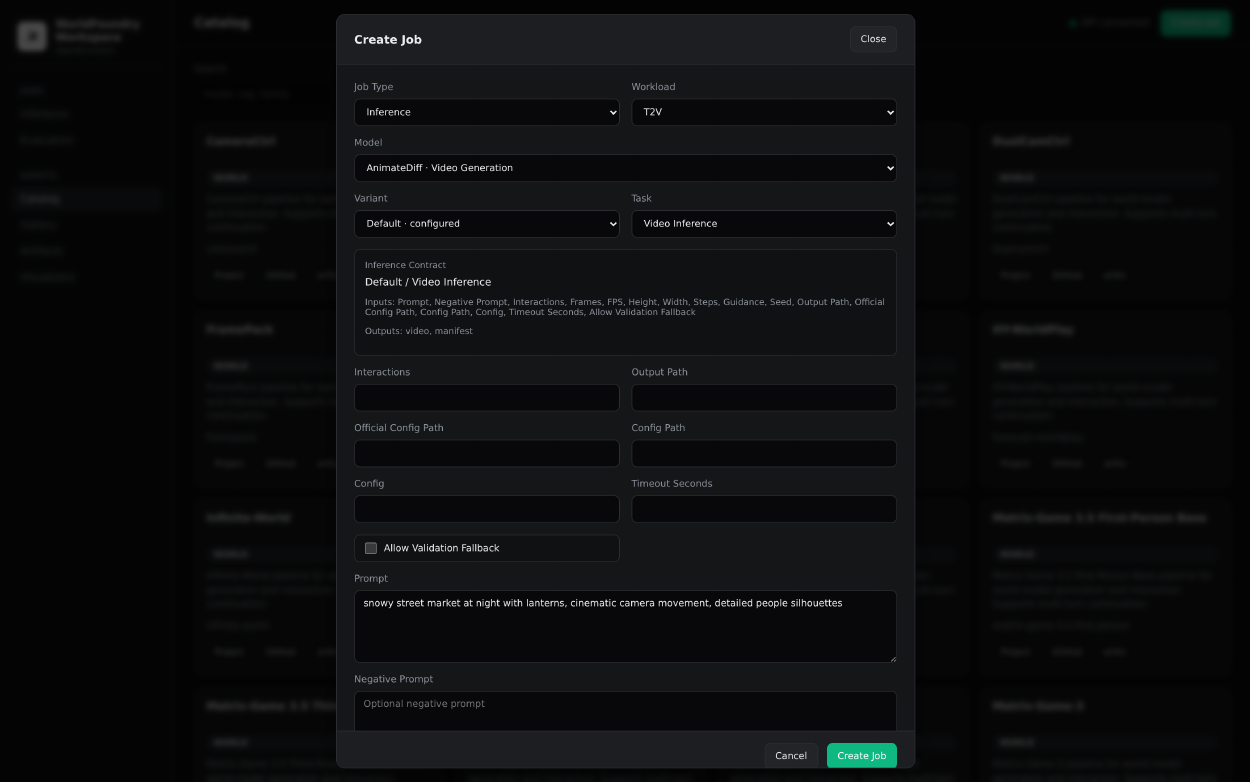}
        \par
    \end{tcolorbox}
\end{tcolorbox}

\captionof{figure}{
    \textbf{Workspace and Studio.}
    WorldFoundry Workspace provides a unified browser interface for
    managing model inference and evaluation workflows, while Studio
    offers interactive environments tailored to different modalities.
    Screenshots from the
    \href{https://openenvision.github.io/WorldFoundry/docs/guides/studio}
    {Studio guide}.
}
\label{fig:surface_workspace}
\end{center}

\begin{center}
\begin{tcolorbox}[
    enhanced,
    width=\linewidth,
    colback=white,
    colframe=oepanelborder,
    colbacktitle=oeaccent,
    coltitle=oeredark,
    boxrule=0.7pt,
    arc=2mm,
    outer arc=2mm,
    boxsep=0pt,
    left=5pt,
    right=5pt,
    top=5pt,
    bottom=5pt,
    toptitle=4pt,
    bottomtitle=4pt,
    fonttitle=\sffamily\bfseries\fontsize{8.6}{10}\selectfont,
    title={
        \textcolor{oered}{\rule{2.2pt}{1.55ex}}
        \hspace{5pt}WORLDFOUNDRY TUI
        \hfill
        \textcolor{oemuted}{GUIDED DISCOVERY AND COMMAND GENERATION}
    }
]
    \begin{tcolorbox}[
        width=\linewidth,
        colback=oeink,
        colframe=oeline,
        boxrule=0.5pt,
        arc=1mm,
        boxsep=0pt,
        left=1pt,
        right=1pt,
        top=1pt,
        bottom=1pt
    ]
        \centering
        \includegraphics[width=0.85\linewidth]{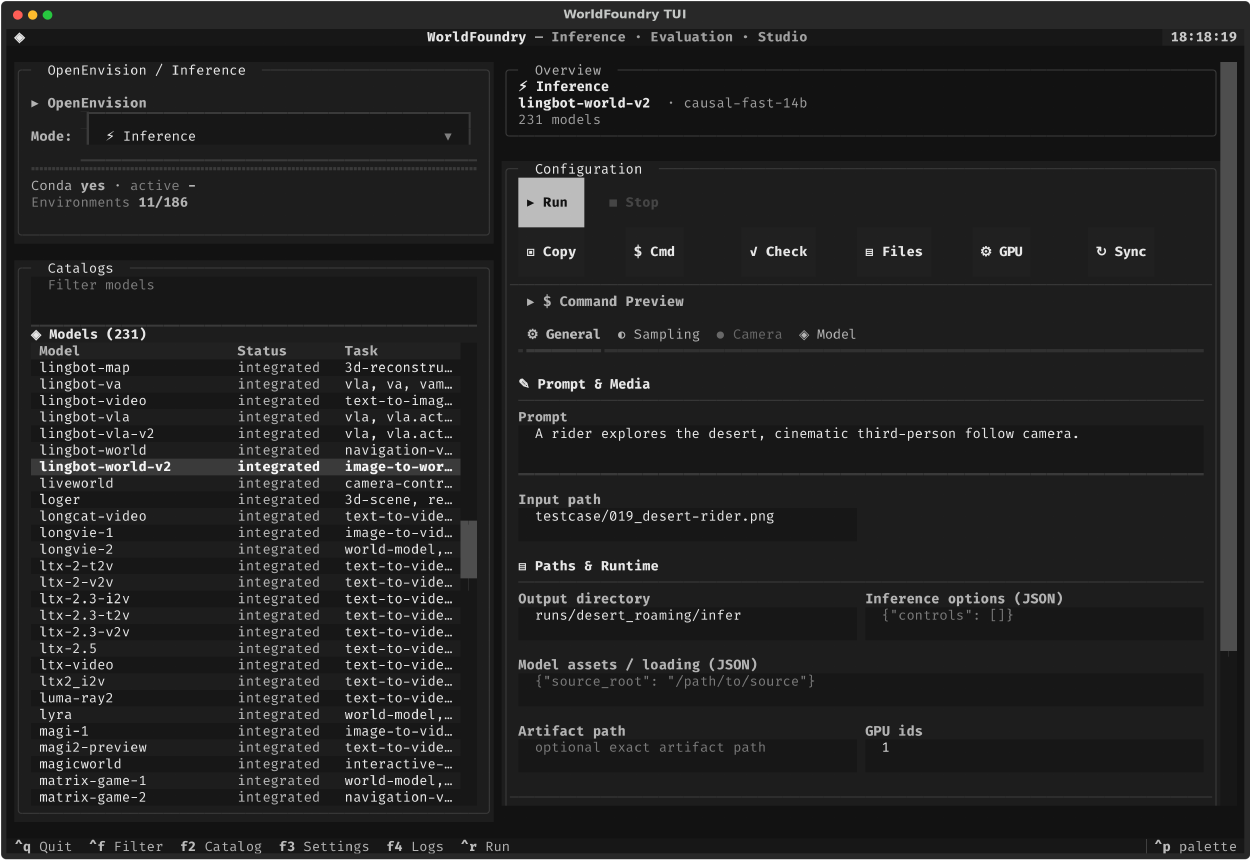}
        \par
    \end{tcolorbox}
\end{tcolorbox}

\captionof{figure}{
    \textbf{Terminal user interface (TUI).}
    WorldFoundry TUI is an interactive terminal interface for configuring
    and executing model inference and evaluation workflows through
    guided menus.
    Screenshot from the
    \href{https://openenvision.github.io/WorldFoundry/docs/guides/tui}
    {TUI guide}.
}
\label{fig:surface_tui}
\end{center}
Studio additionally provides modality-aware visualization and interaction. It displays generated videos, temporal previews, galleries, and artifact references; renders point clouds and meshes in dedicated Viser views; and presents Gaussian splats in the Spark viewer. For interactive world models, a resident model session allows navigation, camera, and text controls to update the environment state without reinitializing the Pipeline for each input. Execution status, runtime logs, generation progress, and performance telemetry help users diagnose model-loading failures, interaction latency, generation instability, and invalid artifacts. Embodied systems that require native simulation connect through an external simulator bridge, which isolates simulator-specific dependencies from the browser-based Workspace. Together, these features unify discovery, configuration, execution control, qualitative inspection, and runtime diagnosis, while preserving generated outputs as durable artifacts for subsequent evaluation.

\subsection{Unified Inference and Interaction}
\label{sec:unified_inference}
\textbf{WorldFoundry} resolves inference and interaction requests through a manifest-driven execution sequence. Each model identifier or alias is mapped to a canonical catalog entry whose manifest specifies provenance, supported tasks and variants, input and output modalities, required assets, runtime configuration, execution binding, and readiness constraints. Given the requested task and variant, the resolver verifies environment and asset compatibility and availability, reports blockers, and selects an execution backend. Once resolution succeeds, the input is encoded as a \texttt{GenerationRequest}, which defines common fields for sample identity, task context, conditioning inputs, control signals, generation parameters, and output specifications while retaining model-specific payloads. This contract standardizes orchestration without imposing uniform input semantics: video systems retain temporal and media-conditioning semantics; 3D and 4D systems retain camera, coordinate-frame, geometry, and rendering semantics; embodied systems retain observation--action schemas and simulator contracts; and interactive world models retain session state and sequential controls. The corresponding \texttt{GenerationResult} records execution status, timing, errors, the resolved configuration, and artifact references, establishing a common basis for traceability and downstream evaluation.

Given a resolved \texttt{GenerationRequest}, the \texttt{WorldModelRunner} coordinates execution through two complementary abstractions: an \emph{Operator} and a \emph{Pipeline}. The Operator handles task-specific adaptation, including input validation, preprocessing, translation into model-native arguments, postprocessing, and conversion of native outputs into framework-compatible artifacts. The Pipeline manages the model lifecycle and inference runtime, including model construction, checkpoint loading, device placement, runtime configuration, distributed execution, and invocation of native inference procedures. This division separates model adaptation from runtime management while accommodating domain-specific dependencies and simulator integration. Interactive systems may retain a resident Pipeline and session state, allowing successive actions or conditions to update the environment without repeated checkpoint loading or runtime initialization. Artifact references and execution metadata populate the \texttt{GenerationResult}; benchmark runners subsequently consume the materialized outputs and provenance records, decoupling evaluation from model loading and inference.

\subsection{Artifact and Representation Contracts}
\label{sec:artifact_contracts}

\textbf{WorldFoundry} defines a two-level contract system that separates sample-level execution from run-level organization. A \texttt{GenerationRequest} represents a single execution unit: \texttt{inputs} contains text, source media, or task-specific observations; \texttt{controls} specifies action sequences, camera trajectories, or other control signals; \texttt{generation\_kwargs} retains model-native parameters such as random seeds and frame rates; and \texttt{output\_schema} declares expected artifact types and structural requirements. The corresponding \texttt{GenerationResult} records stable sample and model identifiers, execution status, errors, timing, resolved metadata, and output references. Each materialized output is described by an \texttt{ArtifactRef}, which identifies externally stored content by its semantic role and URI, with optional MIME type, media properties, file size, and content hash. Being JSON-serializable, these contracts persist, transfer, and can be inspected across process and environment boundaries independently of the originating model implementation.

%Shared artifact types and metadata support discovery, visualization, validation, transfer, and compatibility checks.

The contracts standardize artifact identification, provenance, references, and run organization while preserving native encodings and modality-specific semantics across media, geometry, dynamic 4D representations, actions, and state traces. At the run level, \texttt{execution\_plan.json} specifies intended stages and sample identifiers; \texttt{requests.jsonl} and \texttt{results.jsonl} retain corresponding requests and outcomes; \texttt{sample\_ledger.jsonl} tracks per-stage status and failures; \texttt{artifacts.jsonl} indexes materialized outputs; and \texttt{run\_manifest.json} records the resolved run identity, configuration, runtime context, and output locations. Failed requests remain explicit in result and ledger records, preserving traceability and coverage accounting. This persistent record structure decouples artifact inspection and evaluation from model execution, enabling outputs to be transferred, rescored, compared, and audited without reloading the originating model. It also distinguishes structural conformance, artifact availability, qualitative validity, and benchmark eligibility as separate claims requiring appropriate evidence.

\subsection{Evaluation and Evidence}
\label{sec:evaluation_evidence}

\textbf{WorldFoundry} treats evaluation as an artifact-centered process, allowing scoring to proceed independently of the originating model runtime once the required outputs are materialized. The evaluation pipeline resolves the model, benchmark, metrics, data source, runtime profile, and output directory, and maps task or dataset entries to \texttt{GenerationRequest} records with stable sample identifiers, split assignments, and input keys. It supports scoring existing \texttt{GenerationResult} records, generating outputs for registered metrics, executing a complete model--benchmark evaluation, and applying benchmark-specific evaluators to existing artifact directories. Metrics and benchmarks carry distinct responsibilities: metrics compute per-sample measurements and aggregate statistics from compatible artifacts, spanning semantic alignment, quality, distributional similarity, perceptual comparison, and editing or layout assessment, whereas benchmarks additionally prescribe data, sample coverage, preprocessing, artifact layout, evaluator assets, judge configuration, and aggregation rules, with benchmark runners implementing protocol-specific logic and normalizing external evaluator outputs into a shared result structure. Dependencies and evaluator checkpoints are declared explicitly, and unavailable resources yield explicit skipped or failed states rather than undeclared substitutions. 

The evidence layer records evaluation outcomes and the conditions under which they support scientific comparisons. Each evaluation preserves its execution plan, configuration, environment and code revision, request and result records, sample coverage, stage-level errors, metric outputs, artifact references, and unresolved blockers. Per-sample measurements are stored in \texttt{metrics/per\_sample.jsonl}, aggregate values and failed or skipped sample identifiers in \texttt{metrics/summary.json}, and run metadata and a reproducibility fingerprint in \texttt{run\_manifest.json}. These records inform \texttt{scorecard.json}, the authoritative machine-readable record of score validity, provenance, protocol fidelity, coverage, and comparison eligibility. Derived \texttt{summary.json} and \texttt{report.md} files provide concise machine-readable and human-readable views consistent with the scorecard. Evidence is classified as diagnostic, benchmark-comparable, or an exact benchmark reproduction according to fidelity to the prescribed generation settings, data, evaluator, runtime, and reference configuration. Importing official results validates result normalization, while a bounded evaluator run demonstrates integration operability; neither alone establishes complete benchmark reproduction. Leaderboard eligibility further requires prescribed data and coverage, a verified evaluator, compatible runtime evidence, successful metric computation, and no disqualifying blockers. These distinctions constrain reported claims to the available evidence, preventing successful execution or score production from being mistaken for benchmark validity.

\subsection{Model and Benchmark Zoo}
\label{sec:model_benchmark_zoo}
The Model Zoo is a manifest-driven registry that links model discovery to explicit, auditable integration states. Entries fall into five operational categories: video models for conditional generation, editing, animation, and audiovisual synthesis; generative world models for interactive environments, navigation, and simulation; 3D and 4D models for reconstruction, depth estimation, and static or dynamic scene representations; vision-language-action, vision-action, and world-action models for embodied prediction and control; and provider-hosted models accessed remotely. These categories reflect integration requirements rather than mutually exclusive capabilities. Each entry has a stable identifier and optional aliases, with a manifest documenting official sources, licensing and access conditions, tasks and variants, checkpoint references, input and artifact contracts, runtime bindings, hardware requirements, and blockers. Linked records specify runtime contracts, execution backends, and readiness; environment requirements, including Python and CUDA versions, package constraints, validation imports, and environment isolation; and the concrete Pipeline implementation. Evidence records distinguish source-reported results from those obtained through WorldFoundry, while checkpoints, credentials, and other protected or machine-dependent assets remain outside the repository and are resolved through explicit references. Catalog inclusion therefore establishes discoverability, not executable support: source availability, implementation status, asset readiness, runtime validation, and execution evidence are tracked independently, and entries remain planned or blocked until the relevant requirements are met.

The Benchmark Zoo is a parallel registry that links benchmark discovery to task construction, evaluator execution, result normalization, and evidence reporting, covering video quality, semantic alignment, physical and dynamical consistency, camera and spatial behavior, 3D and 4D understanding, interactive world modeling, safety, memory, and embodied control. Each entry has a stable identifier and optional aliases and documentation of target domains, official sources, datasets and splits, task definitions, sample coverage, artifact requirements, metrics, evaluator assets, judge requirements, result formats, runtime profiles, and blockers. Task manifests define how dataset records or prompts become reproducible sample requests, while benchmark manifests specify the evaluation protocol and the supported execution path that most closely follows it. This path may use an in-framework evaluator, an external official runtime, an official-result importer, a hosted judge, or a simulator interface. Integration verification is recorded separately from eligibility for benchmark comparison: importing a schema-conformant official result verifies parsing and normalization, whereas reproducing an official evaluation also requires the prescribed data, artifact coverage, dependencies, assets, credentials, runtime conditions, and protocol settings. Models and benchmarks are paired only when their declared task semantics and artifact requirements are compatible. Each pairing is then classified as diagnostic, benchmark-comparable, or eligible for an exact-reproduction claim based on provenance, coverage, protocol fidelity, and verification evidence.
\begin{table}[H]
    \centering
    \caption{
        Representative coverage of the WorldFoundry Model Zoo,
        Benchmark Zoo, and Metric Registry. Entries are illustrative
        rather than exhaustive or uniformly mature.
    }
    \label{tab:zoo_representatives}

    \begingroup
    \setlength{\tabcolsep}{6.5pt}
    \renewcommand{\arraystretch}{1.22}
    \arrayrulecolor{oeline}

    \begin{tcolorbox}[
        enhanced,
        colback=white,
        colframe=oepanelborder,
        boxrule=0.7pt,
        arc=2mm,
        outer arc=2mm,
        left=0pt,
        right=0pt,
        top=0pt,
        bottom=0pt,
        boxsep=0pt,
        before skip=2pt,
        after skip=3pt
    ]

    \fontsize{8}{9.6}\selectfont

    \begin{tabularx}{\linewidth}{
        >{\raggedright\arraybackslash}p{0.19\linewidth}
        Y
        >{\raggedright\arraybackslash}p{0.255\linewidth}
    }

        % ================================================================
        % MODEL ZOO
        % ================================================================

        \rowcolor{tabhead}
        \multicolumn{3}{@{}l@{}}{
            \hspace{6.5pt}
            \raisebox{-0.35ex}{
                \textcolor{oered}{\rule{2.4pt}{2.05ex}}
            }
            \hspace{6pt}
            {\sffamily\bfseries\color{oeredark}MODEL ZOO}
            \hspace{7pt}
            {\sffamily\color{oemuted}
                Executable systems and native artifacts}
        } \\

        \rowcolor{white}
        {\sffamily\bfseries\color{oemuted}Family}
        &
        {\sffamily\bfseries\color{oemuted}Representative integrations}
        &
        {\sffamily\bfseries\color{oemuted}Native tasks}
        \\

        \hline
        \rowcolor{tabalt}
        {\sffamily\bfseries Video generation}
        &
        CogVideoX
        \,\textperiodcentered\, HunyuanVideo series
        \,\textperiodcentered\, LTX series
        \,\textperiodcentered\, MiniMax H3
        \,\textperiodcentered\, SANA Video series
        \,\textperiodcentered\, Wan series
        &
        \textcolor{oemuted}{
            Text-, visual-, and audio-conditioned video synthesis;
            video editing
        }
        \\

        \hline
        {\sffamily\bfseries Interactive world models}
        &
        Matrix-Game series
        \,\textperiodcentered\, LingBot-World series
        \,\textperiodcentered\, Hunyuan GameCraft series
        \,\textperiodcentered\, Cosmos 3
        \,\textperiodcentered\, SANA WM
        \,\textperiodcentered\, YUME-1.5
        \,\textperiodcentered\, SCOPE
        &
        \textcolor{oemuted}{
            Action- and camera-conditioned rollouts;
            navigation and scene interaction
        }
        \\

        \hline
        \rowcolor{tabalt}
        {\sffamily\bfseries 3D \& 4D}
        &
        DUSt3R
        \,\textperiodcentered\, VGGT / VGGT-Omega
        \,\textperiodcentered\, Splatt3R
        \,\textperiodcentered\, FlashWorld
        \,\textperiodcentered\, MonST3R
        \,\textperiodcentered\, NeoVerse
        \,\textperiodcentered\, Lyra
        &
        \textcolor{oemuted}{
            3D reconstruction and generation;
            dynamic scene modeling
        }
        \\

        \hline
        {\sffamily\bfseries Embodied action}
        &
        ACT
        \,\textperiodcentered\, OpenVLA
        \,\textperiodcentered\, Lingbot-VLA / VA
        \,\textperiodcentered\, MolmoAct2
        \,\textperiodcentered\, GigaBrain-0
        \,\textperiodcentered\, Being-H0.5
        \,\textperiodcentered\, StarVLA
        &
        \textcolor{oemuted}{
            Action prediction; policy execution;
            language-conditioned control
        }
        \\

        \hline
        \rowcolor{tabalt}
        {\sffamily\bfseries Hosted APIs}
        &
        Seedance 2 / 2.5
        \,\textperiodcentered\, Kling API
        \,\textperiodcentered\, Runway Gen-4.5
        \,\textperiodcentered\, Sora2
        \,\textperiodcentered\, Veo3
        \,\textperiodcentered\, World Labs API
        &
        \textcolor{oemuted}{
            API-based video and world generation
        }
        \\

        % ================================================================
        % BENCHMARK ZOO
        % ================================================================

        \rowcolor{tabhead}
        \multicolumn{3}{@{}l@{}}{
            \hspace{6.5pt}
            \raisebox{-0.35ex}{
                \textcolor{oered}{\rule{2.4pt}{2.05ex}}
            }
            \hspace{6pt}
            {\sffamily\bfseries\color{oeredark}BENCHMARK ZOO}
            \hspace{7pt}
            {\sffamily\color{oemuted}
                Evaluation protocols and evidence}
        } \\

        \rowcolor{white}
        {\sffamily\bfseries\color{oemuted}Domain}
        &
        {\sffamily\bfseries\color{oemuted}Representative benchmarks}
        &
        {\sffamily\bfseries\color{oemuted}Evaluation focus}
        \\

        \hline
        \rowcolor{tabalt}
        {\sffamily\bfseries Video generation}
        &
        VBench series
        \,\textperiodcentered\, T2V-CompBench
        \,\textperiodcentered\, ChronoMagic-Bench
        \,\textperiodcentered\, VideoScore
        \,\textperiodcentered\, VideoPhy2
        \,\textperiodcentered\, T2VSafetyBench
        &
        \textcolor{oemuted}{
            Perceptual quality; semantic alignment; compositionality;
            temporal dynamics; physical plausibility; content safety
        }
        \\

        \hline
        {\sffamily\bfseries World models}
        &
        WorldBench
        \,\textperiodcentered\, WorldModelBench
        \,\textperiodcentered\, WorldScore
        \,\textperiodcentered\, 4DWorldBench
        \,\textperiodcentered\, iWorld-Bench
        \,\textperiodcentered\, WorldAtlas
        &
        \textcolor{oemuted}{
            Dynamics fidelity; 3D/4D consistency; memory retention;
            controllability; interactive task performance
        }
        \\

        \hline
        \rowcolor{tabalt}
        {\sffamily\bfseries Embodied AI}
        &
        AI2-THOR
        \,\textperiodcentered\, BEHAVIOR-1K
        \,\textperiodcentered\, CALVIN
        \,\textperiodcentered\, LIBERO
        \,\textperiodcentered\, RoboTwin series
        \,\textperiodcentered\, WorldArena series
        \,\textperiodcentered\, RoboDojo
        &
        \textcolor{oemuted}{
            Closed-loop task completion in simulated environments
        }
        \\

        % ================================================================
        % METRIC REGISTRY
        % ================================================================

        \rowcolor{tabhead}
        \multicolumn{3}{@{}l@{}}{
            \hspace{6.5pt}
            \raisebox{-0.35ex}{
                \textcolor{oered}{\rule{2.4pt}{2.05ex}}
            }
            \hspace{6pt}
            {\sffamily\bfseries\color{oeredark}METRIC REGISTRY}
            \hspace{7pt}
            {\sffamily\color{oemuted}
                In-tree scorers and compute APIs}
        } \\

        \rowcolor{white}
        {\sffamily\bfseries\color{oemuted}Family}
        &
        {\sffamily\bfseries\color{oemuted}Representative metrics}
        &
        {\sffamily\bfseries\color{oemuted}Measurement focus}
        \\

        \hline
        \rowcolor{tabalt}
        {\sffamily\bfseries Scorers \& quality}
        &
        CLIPScore
        \,\textperiodcentered\, VQAScore
        \,\textperiodcentered\, ITMScore
        \,\textperiodcentered\, FaceScore
        \,\textperiodcentered\, ArtScore
        \,\textperiodcentered\, OpenS2V
        &
        \textcolor{oemuted}{
            Semantic alignment; identity consistency;
            aesthetic quality; subject-to-video quality
        }
        \\

        \hline
        {\sffamily\bfseries Distribution}
        &
        FID
        \,\textperiodcentered\, IS
        \,\textperiodcentered\, KID
        \,\textperiodcentered\, FVD
        \,\textperiodcentered\, CMMD
        \,\textperiodcentered\, Clean-FID
        \,\textperiodcentered\, JEDi
        &
        \textcolor{oemuted}{
            Image/video distributional fidelity and sample diversity
        }
        \\

        \hline
        \rowcolor{tabalt}
        {\sffamily\bfseries Perceptual pairwise}
        &
        LPIPS
        \,\textperiodcentered\, SSIM / MS-SSIM
        \,\textperiodcentered\, PSNR
        \,\textperiodcentered\, DINO / DreamSim
        \,\textperiodcentered\, mask accuracy
        \,\textperiodcentered\, LQS
        &
        \textcolor{oemuted}{
            Perceptual and structural similarity;
            mask accuracy; layout quality
        }
        \\

        \hline
        {\sffamily\bfseries Editing \& layout}
        &
        SemSR
        \,\textperiodcentered\, IRS
        \,\textperiodcentered\, CAS
        \,\textperiodcentered\, manipulation direction
        \,\textperiodcentered\, VS similarity
        \,\textperiodcentered\, object-wise consistency
        &
        \textcolor{oemuted}{
            Semantic edit success; manipulation direction;
            object-level consistency
        }
        \\

    \end{tabularx}
    \end{tcolorbox}

    \parbox{\linewidth}{
        \fontsize{7.5}{9.2}\selectfont
        \color{oemuted}
        {\sffamily\bfseries Reading the table.}
        Catalog inclusion, runnable integration, and runner/evidence
        verification are distinct maturity claims.
        The live
        \href{https://openenvision.github.io/WorldFoundry/docs/guides/supported-models}
        {Model recipes},
        \href{https://openenvision.github.io/WorldFoundry/docs/evaluation/benchmark-hub}
        {Benchmark Hub}, and
        \href{https://openenvision.github.io/WorldFoundry/docs/evaluation/metrics}
        {Metrics registry}
        remain authoritative for updates.
    }

    \endgroup
\end{table}

\section{Empirical Validation and Analysis}
\label{sec:analysis}

%Following the infrastructure design, 
We evaluate \textbf{WorldFoundry} using representative models for video generation, interactive world modeling, 3D and 4D generation, and embodied action prediction and control. The evaluation first verifies whether the framework preserves native input and output semantics and produces results consistent with native implementations under matched conditions. It then examines whether equivalent requests issued through the CLI, Python API, and web interface yield consistent configurations and execution behavior. Latency and GPU memory overhead are profiled last, together with the effort required to integrate a new model.

\subsection{Experimental Setting}
Experiments are conducted on 128 NVIDIA A100/H100 GPUs with 80 GB of memory each; evaluation on other hardware is left to future work. As model families differ in objectives, inputs, outputs, and execution semantics, we evaluate infrastructure capabilities rather than ranking performance across domains. Specifically, we assess whether the framework resolves declarative catalog entries into executable configurations, preserves model-native semantics across interfaces, stores heterogeneous outputs as persistent artifacts, and supports reported claims with traceable evidence. All workflows share a lifecycle of discovery, preparation, execution, inspection, and evaluation while retaining domain-specific inputs, controls, runtime requirements, and output representations. The evaluation examines whether this shared infrastructure supports reproducible and auditable workflows without imposing a common task, representation, or evaluation metric.
\subsection{Experimental Results}
We report results by model family: video generation (\S\ref{sec:video_gen}), interactive world models (\S\ref{sec:interactive_wm}), 3D and 4D generation (\S\ref{sec:3d}), and embodied action generation (\S\ref{sec:embodied}). For each family, we describe the task setting, inputs and outputs, and representative systems executed through \textbf{WorldFoundry}. We then present inference validation results with visualizations of the generated outputs and summarize the empirical insights obtained from these executions, comparing qualitative behavior with published findings where noted.

\subsubsection{Video Generation.}
\label{sec:video_gen}

\begin{figure}[H]
    \centering
    \includegraphics[width=\linewidth]{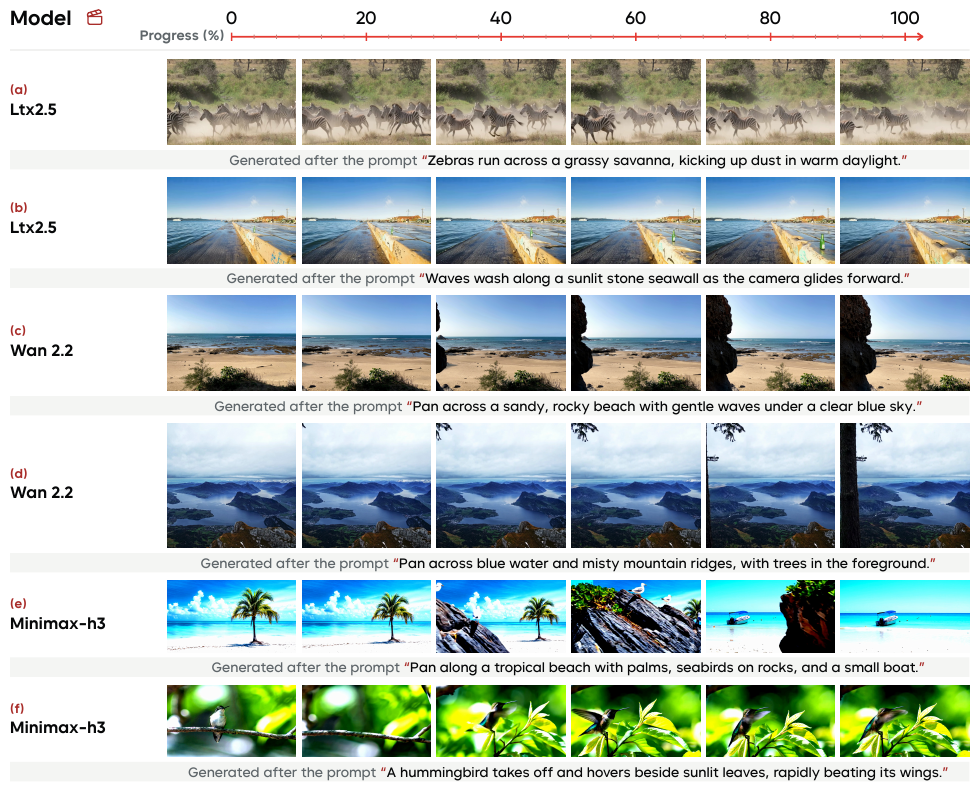}
    \caption{Demonstration of inference results from representative video generation models.}
    \label{fig:video_generation}
\end{figure}

For video generation, we consider text-to-video synthesis, image-to-video synthesis, and video-to-video editing. These tasks assess whether a model can generate plausible visual dynamics, preserve subject identity and scene structure over time, and satisfy task-specific conditions, capabilities central to modeling how the visual world evolves. Successful generation therefore requires both fidelity to the supplied conditions and coherent temporal progression in appearance, spatial relationships, and motion. Depending on the task, inputs consist of a textual description alone or of visual conditions, such as reference images or source videos, paired with instructions specifying the desired content, motion, or transformation; additional reference modalities and structural constraints are supplied where the model supports them. Representative systems include CogVideoX~\citep{cogvideox}, Vchitect-2.0~\citep{vchitect2}, HunyuanVideo 1.5~\citep{hunyuanvideo15}, LTX-2.5~\citep{wfmissing_ltx25}, Wan2.1~\citep{wan}, Wan2.1 VACE~\citep{vace}, Wan2.2~\citep{wan22}, LingBot-Video~\citep{lingbotvideo}, and MiniMax H3~\citep{minimaxh3}, which span diverse conditioning mechanisms and generation pipelines. WorldFoundry preserves each system's native input processing and inference settings, so that results reflect its intended operation. Figure~\ref{fig:video_generation} shows representative text-to-video results from LTX-2.5, Wan2.2, and MiniMax H3, covering both camera motion through largely static scenes and object-centric dynamics such as running animals and a hovering bird.

Qualitative analysis highlights distinct strengths across generation tasks. In text-conditioned synthesis, CogVideoX and Vchitect-2.0 capture scene composition and broad motion patterns, but complex motion and multi-object interactions expose inconsistencies in identity, fine detail, and spatial relationships. FLAG3D++~\cite{tang2025flag3dpp} provides detailed and professional language instructions to describe how to perform a specific activity, as well as the corresponding accurate and dense 3D human pose captured from an advanced MoCap system to handle complex activities and large movements. HunyuanVideo 1.5 and Wan2.1 produce coherent short sequences, particularly with image conditioning, although appearance preservation does not consistently translate into plausible motion under occlusion or substantial deformation. Published results for Wan2.2~\citep{wan22} report improvements in motion generation and aesthetic control over Wan2.1, extending the range of visually convincing sequences. LingBot-Video provides a complementary advance: its reported post-training comparisons~\citep{lingbotvideo} show reduced object deformation, interpenetration, and premature object release, indicating gains in physical interaction beyond perceptual quality alone. MiniMax H3 broadens reference-based generation and editing, with official demonstrations combining appearance, motion, and audio cues from different sources~\citep{minimaxh3}; these examples demonstrate compositional flexibility but do not establish reliable physical dynamics. Together, these observations suggest that progress unfolds along several distinct dimensions: visual refinement, preservation of conditioning information, and consistency of the depicted events. A sequence may remain visually smooth and recognizable while violating contact relations or failing to complete the requested transformation, which makes event-level consistency a necessary complement to frame-level quality in video-generation analysis.

\subsubsection{Interactive World Model.}
\label{sec:interactive_wm}
\begin{figure}[H]
    \centering
    \includegraphics[width=\linewidth]{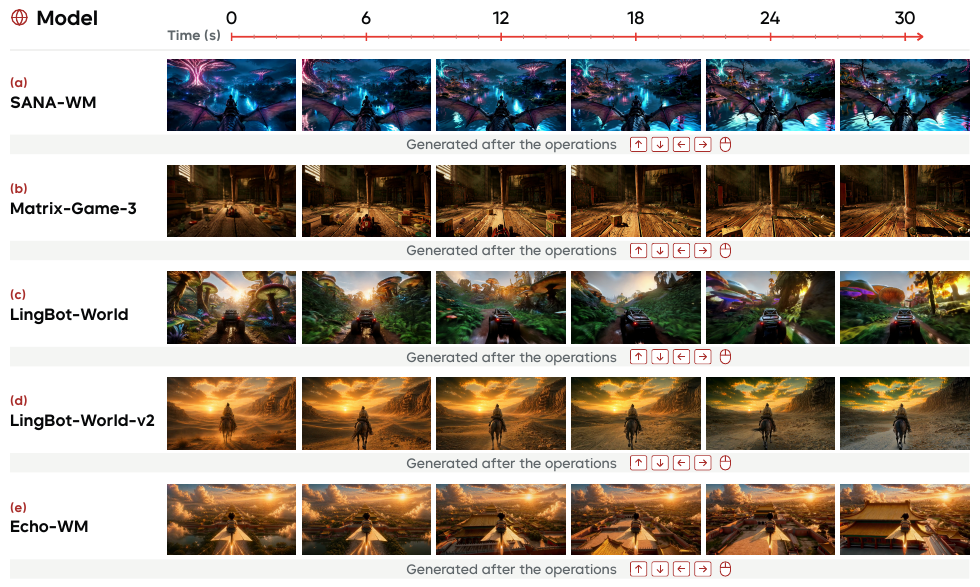}
    \caption{Demonstration of inference results from representative world models.}
    \label{fig:interactive_wm}
\end{figure}

\begingroup
\setlength{\emergencystretch}{2em}
For interactive world modeling, we consider navigation, camera-controlled exploration, and action- or instruction-driven interaction within generated environments. These tasks assess whether a model can maintain a coherent environment as actions unfold, retain information beyond the immediate visual context, and carry the consequences of earlier interactions into subsequent predictions. Unlike offline video synthesis, the model produces an evolving observation stream conditioned on successive interventions, allowing later actions to respond to previously generated observations. Inputs comprise an initial image or image sequence, the available interaction history, and model-supported controls, including directional movement, camera rotation, discrete game actions, and textual instructions. Successful generation therefore requires both immediate responsiveness to controls and persistent consistency in scene geometry, object identity, and interaction outcomes, including when previously observed regions are revisited. Representative systems include the Matrix-Game series up to the recent 3.5 release~\citep{matrixgame35}, LingBot-World 1.0 and 2.0~\citep{lingbotworld,lingbotworld2}, Hunyuan GameCraft and GameCraft-2~\citep{gamecraft,gamecraft2}, YUME-1.5~\citep{yume15}, HY-WorldPlay~\citep{hyworldplay}, SCOPE~\citep{scope}, SANA-WM~\citep{wfmodel_sanawm}, and Echo-WM~\citep{wfmissing_echo_wm}. WorldFoundry preserves each system's native action vocabulary, history-conditioning mechanism, and generation schedule, and records controls, observations, and runtime configurations to support reproducible trajectory-level analysis. Figure~\ref{fig:interactive_wm} shows 30-second rollouts from five representative systems driven by directional movement and mouse-based camera control across diverse scenes.\par
\endgroup

Qualitative comparison highlights progress along two complementary dimensions: scene persistence and interaction expressiveness. Matrix-Game 2.0~\citep{matrixgame2} enables responsive navigation but exhibits appearance drift over extended rollouts, whereas LingBot-World 1.0, Hunyuan GameCraft, YUME-1.5, and HY-WorldPlay produce visually coherent navigation sequences in the examined cases. Recent memory-augmented systems address consistency beyond the immediate context: Matrix-Game 3.0~\citep{matrixgame3} introduces long-horizon memory, while published results for Matrix-Game 3.5~\citep{matrixgame35} and ReWorld~\citep{reworld} demonstrate scene recall through geometry-aware retrieval and pose-indexed landmarks, respectively. These advances make revisitation more informative than uninterrupted forward motion for assessing whether a model retains previously observed content. In parallel, LingBot-World 2.0~\citep{lingbotworld2} and Hunyuan-GameCraft-2~\citep{gamecraft2} broaden interaction to diverse actions and instruction-driven events, while SCOPE~\citep{scope} demonstrates localized firing and reloading responses alongside continuous movement and camera control. However, scene recall and action responsiveness are distinct from persistent interaction outcomes: recovering a familiar view does not establish that an opened door remains open or that a displaced object retains its new position after leaving the field of view. The resulting evaluation priority is therefore to distinguish memory of scene appearance from memory of changes caused by actions. Reliable interactive world modeling requires both: stable recovery of unchanged content and consistent propagation of action-dependent state changes across subsequent observations.

\subsubsection{3D \& 4D Generation.}
\label{sec:3d}
\begin{figure}[H]
    \centering
    \includegraphics[width=\linewidth]{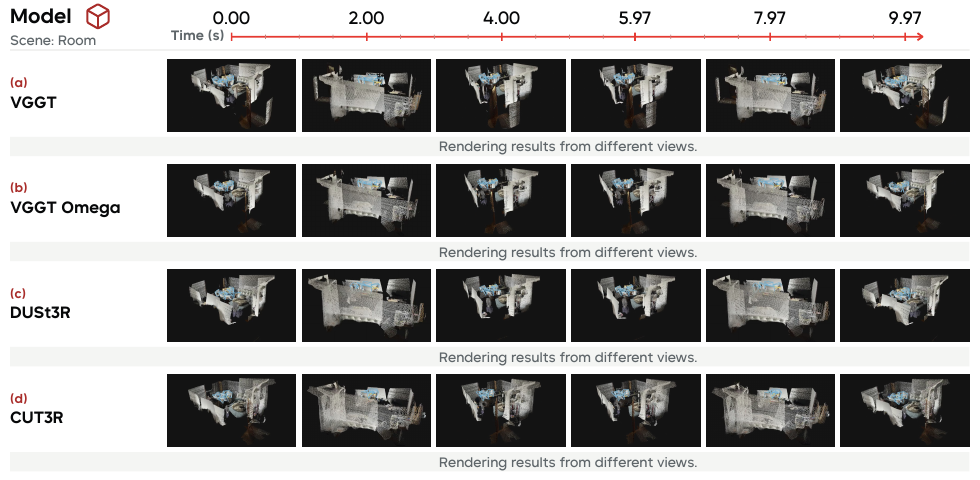}
    \caption{Demonstration of inference results from representative 3D scene generation models.}
    \label{fig:3d_4d}
\end{figure}

For 3D and 4D generation, we consider static scene reconstruction, scene synthesis, and dynamic scene modeling. These tasks assess whether a model can recover or generate spatial structure, maintain consistency across viewpoints, and represent changes over time. Successful reconstruction requires geometry and camera estimates consistent with the observed views, while conditional synthesis requires spatially coherent content that follows the supplied visual or textual conditions. Dynamic scene modeling additionally requires distinguishing camera motion from scene motion and preserving temporal correspondence through deformation, occlusion, and reappearance. Depending on the task, inputs comprise single images, multi-view observations, video sequences, or textual descriptions, and outputs include point maps, camera parameters, point clouds, meshes, Gaussian representations, and time-dependent geometry. Representative systems include DUSt3R~\citep{dust3r}, CUT3R~\citep{wfmodel_cut3r}, VGGT~\citep{vggt}, VGGT-Omega~\citep{vggtomega}, Splatt3R~\citep{splatt3r}, and FlashWorld~\citep{flashworld} for spatial reconstruction or synthesis, alongside MonST3R~\citep{monst3r}, NeoVerse~\citep{neoverse}, and Lyra~\citep{lyra} for complementary video-based and spatiotemporal workflows. Evaluation criteria include camera consistency, reconstruction completeness, novel-view quality, inference efficiency, and, where reference geometry is available, geometric accuracy, with additional attention to temporal stability in dynamic scenes. WorldFoundry preserves each method's native inputs and representations, organizes comparisons within compatible task families, and records configurations, rendered views, and exported artifacts for reproducible spatial and temporal analysis. Figure~\ref{fig:3d_4d} compares VGGT, VGGT-Omega, DUSt3R, and CUT3R on the same indoor scene, rendering each reconstruction from a common sequence of viewpoints.

Qualitative analysis highlights a distinction between reconstructing observed structure and generating plausible spatial completion. DUSt3R recovers geometry through pairwise point-map prediction, while VGGT jointly estimates camera parameters, depth, point maps, and point tracks from multiple views; VGGT-Omega further develops this feed-forward reconstruction paradigm. Splatt3R complements geometric reconstruction with Gaussian appearance attributes for novel-view rendering, although convincing rendered views do not establish accurate geometry in sparsely observed or occluded regions. FlashWorld enables efficient scene synthesis through direct Gaussian generation, while Lyra transfers video-generation priors into feed-forward 3D and 4D reconstruction, illustrating how learned priors can supply plausible content beyond the available observations. In dynamic scenes, MonST3R accommodates scene motion when estimating geometry and cameras, and NeoVerse extends monocular reconstruction to time-varying scenes and novel camera trajectories. The central ambiguity is whether image changes arise from camera motion, object motion, or newly visible surfaces: temporal smoothness alone cannot establish correct geometric correspondence. For extended exploration, Lyra 2.0~\citep{lyra2} reports improved consistency through geometry-guided retrieval and training that mitigates accumulated generation drift. Together, these developments motivate distinguishing three properties: agreement with observed evidence, cross-view consistency of inferred content, and temporal persistence of scene structure. A representation can be visually convincing and internally consistent while remaining inaccurate in unobserved regions; reconstructing a dynamic sequence likewise does not establish the ability to predict its evolution under alternative actions.

\subsubsection{Embodied Action Generation.}
\label{sec:embodied}

\begin{figure}[H]
    \centering
    \includegraphics[width=\linewidth]{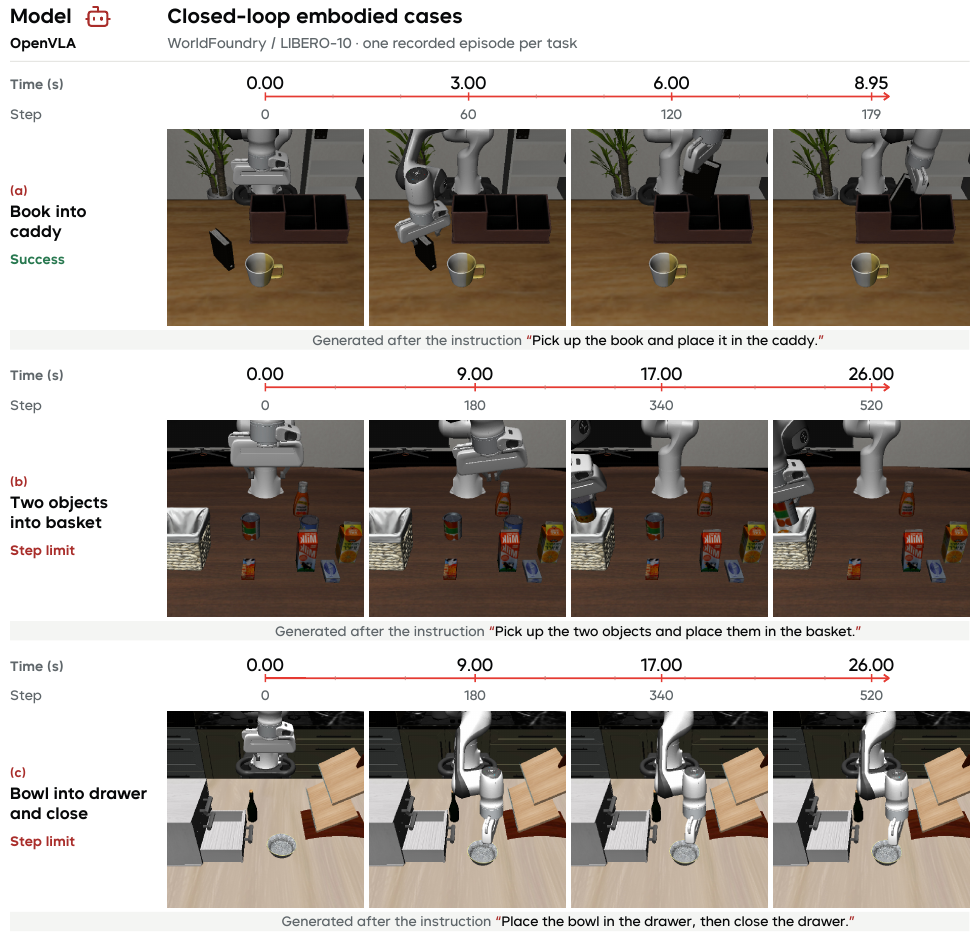}
    \caption{\textbf{Representative closed-loop embodied results.} OpenVLA executed through WorldFoundry on LIBERO-10, with one recorded episode per task. The policy completes task (a) within 179 steps, whereas tasks (b) and (c) reach the 520-step limit before completion.}
    \label{fig:embodied}
\end{figure}

For embodied action generation, we consider visual policy learning, language-conditioned manipulation, and control informed by predicted environmental dynamics. These tasks assess whether a model can translate observations and task specifications into executable actions, coordinate perception with control, and adapt to feedback during execution. Inputs comprise visual observations and proprioceptive states, supplemented by natural-language instructions or predicted future observations where supported. Outputs comprise discrete actions, continuous control commands, or action sequences expressed in the target embodiment's native action space. The integrated systems follow three complementary paradigms: Vision-Action (VA) policies, which generate actions from visual and robot-state observations; Vision-Language-Action (VLA) models, which additionally incorporate linguistic instructions; and World-Action Models (WAMs), which couple action generation with predictions of environmental evolution. Representative integrations include Action Chunking Transformer~\citep{act}, OpenVLA~\citep{openvla}, RoboFlamingo~\citep{roboflamingo}, MolmoAct2~\citep{molmoact2}, GigaBrain-0~\citep{gigabrain0}, Being-H0.5~\citep{beingh05}, StarVLA~\citep{starvla}, LingBot-VLA 1.0 and 2.0~\citep{lingbotvla,lingbotvla2}, and LingBot-VA 1.0 and 2.0~\citep{lingbotva,lingbotva2}. Evaluation combines offline action prediction with closed-loop execution in compatible manipulation environments, including LIBERO and RoboTwin~\citep{libero,robotwin2}; criteria include task completion, instruction adherence, control accuracy, trajectory smoothness, execution latency, and recovery from intermediate errors. WorldFoundry preserves each system's observation history, camera configuration, action coordinates, normalization statistics, control frequency, and action-chunk horizon, with comparisons restricted to compatible embodiments and task protocols. Figure~\ref{fig:embodied} shows closed-loop LIBERO-10 episodes of OpenVLA executed through WorldFoundry: the policy completes a single-object pick-and-place task but reaches the step limit on two multi-stage tasks, illustrating why closed-loop completion, rather than offline action accuracy, is the relevant measure.

Published empirical findings indicate that semantic grounding, embodiment transfer, and predictive control contribute to embodied competence through distinct mechanisms. SuperMimic \cite{liao2026supermimic} demonstrates that a shared differentiable world model can efficiently scale humanoid whole-body control through short imagined rollouts and analytic policy gradients, substantially reducing the computational cost of policy learning. OpenVLA~\citep{openvla} and RoboFlamingo~\citep{roboflamingo} demonstrate that pretrained vision-language representations can support language-conditioned manipulation, although transfer remains dependent on the target embodiment and adaptation data. MolmoAct2~\citep{molmoact2} combines embodied reasoning with a dedicated action expert, exposing a trade-off between additional spatial reasoning and control latency. GigaBrain-0~\citep{gigabrain0} improves generalization through world-model-generated training data, whereas Being-H0.5~\citep{beingh05} aligns human and robot motion within a shared action representation. StarVLA~\citep{starvla} provides complementary policy variants for examining how action decoding and training choices affect execution, rather than defining a single control mechanism. LingBot-VLA 1.0~\citep{lingbotvla} demonstrates effective adaptation across robot platforms, while LingBot-VLA 2.0~\citep{lingbotvla2} extends this approach to whole-body control and incorporates future semantic and geometric prediction as auxiliary supervision. LingBot-VA 1.0~\citep{lingbotva} instead couples video and action prediction within a causal sequence model; LingBot-VA 2.0~\citep{lingbotva2} further introduces action-aligned visual representations and asynchronous inference that incorporates the latest observations during execution. These distinctions suggest that the value of world modeling depends on where prediction enters the control pipeline: as training-data augmentation, representation supervision, or an online component of action generation. In particular, plausible future observations are useful only insofar as they support executable actions and timely correction when reality departs from prediction. The decisive evidence is therefore improved closed-loop completion and recovery under comparable data, embodiment, and latency constraints, rather than predictive visual quality or offline action accuracy alone.

\section{Limitations and Future Work}
\label{sec:limitations}

Although \textbf{WorldFoundry} provides a unified operational foundation for diverse generative world-modeling systems, its broad coverage encompasses different levels of execution maturity. Catalog inclusion, runnable integration, local asset readiness, official runtime validation, and benchmark eligibility remain distinct states, reflecting the external checkpoints, licensed datasets, hosted services, simulator assets, and dedicated environments required by individual models and benchmarks. Differences in architecture and hardware requirements also prevent a single optimization strategy from benefiting every video, spatial, interactive, and embodied model equally. In addition, the present framework focuses on inference, interaction, artifact management, and evaluation. A general training infrastructure remains a direction for future development, and upstream training code is intentionally separated from current runtime packages. This scope provides a consistent and auditable foundation for experimentation, while leaving opportunities to extend lifecycle reproducibility as public datasets and reproducible training recipes become available across domains.

The inference core already exposes a broad set of optimization mechanisms: configurable attention backends, including FlashAttention 2~\citep{flashattention2} and 3~\citep{flashattention3}, SageAttention~\citep{sageattention}, xFormers, and SDPA; Triton kernel registration; compilation and inference caching; NVFP4 quantization where supported; hardware-aware GPU selection; and multi-GPU context and sequence parallelism. Their effects, however, have not yet been characterized systematically across model families, and validation has primarily used A100 and H100 GPUs; limited access to B200- and B300-class systems constrains assessment on newer accelerators. Future work will combine broader hardware validation with runtime profiling, automatic backend selection, improved compilation and cache reuse, fidelity-aware quantization, and more efficient memory management and distributed execution. We will also integrate dedicated serving engines through modular backend adapters and runtime profiles that preserve native input semantics, artifact contracts, and evaluation protocols. Improvements will be assessed by end-to-end latency, throughput, memory use, scaling efficiency, and reproducibility, since kernel-level acceleration alone does not establish application-level gains. An optimization will be adopted only if its outputs remain within a declared fidelity tolerance of the unoptimized route.

Beyond inference optimization, we plan to extend \textbf{WorldFoundry} to support data preparation and training. Initial development will focus on representative baselines and public datasets, reflecting the uneven availability of reproducible training recipes across domains. The extension will support versioned dataset manifests, distributed training, fine-tuning and post-training, checkpoint lineage, experiment recovery, and direct checkpoint registration in the Model Zoo. Existing contracts will link data and optimization configurations to checkpoints, inference artifacts, and benchmark results. Since reproducible training alone does not guarantee consistent deployment behavior, we will explicitly align preprocessing, conditioning, normalization, coordinate conventions, and checkpoint interpretation across training and inference. Reusable configurations, guided setup, and actionable diagnostics will ease adoption while retaining domain-specific controls. Training, execution, and evaluation will remain distinct responsibilities within an auditable research lifecycle, preserving architectural flexibility and traceability from data preparation to reported results.

\section{Conclusion}
\label{sec:conclusion}

We introduce \textbf{WorldFoundry} as open-source infrastructure built on an operational definition of generative world models as systems with persistent state, intervention-conditioned dynamics, and observational feedback. This definition shifts the criterion for world modeling from the quality of individual outputs to the behavior of the loop that produces them: a generated observation counts as evidence of a world model only insofar as it is constrained by a state that past interventions have altered and that new observations can revise. WorldFoundry relies on this view across video, 3D and 4D, interactive world, and embodied action systems, cataloging 270 model entries, 74 benchmarks, and 62 metrics under shared contracts that preserve native semantics. By recording catalog inclusion, executability, and validated performance as distinct claims, it keeps every reported result traceable to the artifacts and conditions that support it. Our cross-family analysis indicates that visual fidelity, scene consistency, and action responsiveness, although complementary, cannot by themselves certify world modeling; what current evaluations seldom measure is whether intervention effects persist and whether incoming observations correct the state. Future work will develop evaluations that directly probe intervention persistence and feedback assimilation, alongside broader inference optimization, hardware validation, and reproducible training with explicit training--inference consistency. By connecting capability definitions, executable workflows, and traceable evidence, WorldFoundry provides a shared foundation for measuring progress toward world intelligence.

\clearpage

\phantomsection
\addcontentsline{toc}{section}{References}
\begingroup
\normalem
\hypersetup{urlcolor=oeredark,linkcolor=oeredark}
\bibliographystyle{unsrtnat}
\setlength{\bibsep}{4pt plus 1pt minus 1pt}
\bibliography{main}
\endgroup

\clearpage
\beginappendix
\section{Author Contributions}
\label{sec:appendix}
\newcommand{\contribrole}[1]{%
    \textcolor{oeredark}{\textbf{#1}}%
}

\begin{itemize}[leftmargin=*, itemsep=0.5\baselineskip]

    \item \textbf{OpenEnvision Team}:
    \contribrole{Project Team.}

    \item \textbf{Juanxi Tian (NTU)}:
    \contribrole{Project Leader \& Core Contributor.} Leads the project, designs and implements WorldFoundry,
and writes the technical report.

    \item \textbf{Jiadong Hou (NTU)}:
    \contribrole{Core Contributor.}
    Co-develops WorldFoundry and coordinates its updates.

    \item \textbf{Baiang Li (Princeton)}:
    \contribrole{Contributor.}
    Reviews and helps refine WorldFoundry's video generation pipeline.

    \item \textbf{Xin Lin (UCSD)}:
    \contribrole{Contributor.}
    Reviews and helps refine WorldFoundry's 3/4D generation pipeline.

    \item \textbf{Zhengbo Jiao (HKUST)}:
    \contribrole{Contributor.} Tests WorldFoundry's pipelines and supports outreach for the framework.

    \item \textbf{Weijia Zhang (Yale)}:
    \contribrole{Contributor.} Tests WorldFoundry's pipelines and supports outreach for the framework.
    
    \item \textbf{Yeying Jin (NUS)}: 
    \contribrole{Advisor.}
   Provides advice on writing the technical report.

    \item \textbf{Truong Nguyen (UCSD)}:
    \contribrole{Advisor.}
    Provides advice on writing the technical report.

    \item \textbf{Haitian Zheng (Adobe)}:
    \contribrole{Advisor.}
    Offers advice on WorldFoundry's design and development.
    
    \item \textbf{Zhifei Zhang (Adobe)}: 
    \contribrole{Advisor.}
    Offers advice on WorldFoundry's design and development.

    \item \textbf{Jianfei Yang (NTU)}: 
    \contribrole{Advisor.} Offers advice on WorldFoundry's design and development.

    \item \textbf{Ming-Hsuan Yang (UC Merced)}: \contribrole{Advisor.} Offers advice on WorldFoundry's design and development.
    
    \item \textbf{Felix Heide (Princeton)}: \contribrole{Project Supervisor.} 
    Supervises the WorldFoundry project and provides guidance on the technical report.

    \item \textbf{Wei Xiong (Nvidia)}: \contribrole{Project Supervisor.}
    Supervises the WorldFoundry project and provides guidance on the technical report.

    \item \textbf{Enze Xie (Nvidia)}: \contribrole{Project Supervisor.}
    Supervises the WorldFoundry project and provides guidance on the technical report.

    \item \textbf{Wanhua Li (NTU)}: \contribrole{Project Supervisor.} Supervises the WorldFoundry project and provides guidance on the technical report.
    %\item \textbf{Yeying Jin (NUS)}:

    %\item \textbf{Zhifei Zhang (Adobe)}:

    %\item \textbf{Wei Xiong (Nvidia)}:

    %\item \textbf{Truong Nguyen (UCSD)}:

    %\item \textbf{Felix Heide (PrincetonU)}:

    %\item \textbf{Ming-Hsuan Yang (UCM)}:

    %\item \textbf{Wanhua Li (NTU)}:

    %\item \textbf{Enze Xie (Nvidia)}:

\end{itemize}

\clearpage

\section{Supported Models}
\label{sec:supported-models}

The \href{https://openenvision.org/WorldFoundry/docs/guides/supported-models}{WorldFoundry Model Zoo} comprises \textbf{270 distinct registry entries}, organized according to the five families in Table~\ref{tab:zoo_representatives}: 64 video generation models (including multimodal generation), 109 world models, 34 3D/4D models, 53 embodied models, and 10 hosted API models. Tables~\ref{tab:models-video}--\ref{tab:models-hosted-api} enumerate every entry with its stable registry ID, upstream reference, and readiness status. Checkpoint variants and aliases are consolidated under a single entry unless the registry assigns them separate IDs. Model names link to the corresponding papers, technical reports, official releases, model cards, or API documentation. Where no upstream publication exists, the source is the official release or a registry record.

Readiness is evidence-aware; all 270 entries are marked \emph{Runner verified} in this appendix. Verification scope is specific to the corresponding execution route, model variant, and configuration.

\begingroup
\footnotesize
\setlength{\tabcolsep}{4pt}
\renewcommand{\arraystretch}{1.08}
\begin{longtable}{@{}>{\raggedright\arraybackslash}p{0.43\linewidth}>{\raggedright\arraybackslash}p{0.31\linewidth}>{\raggedright\arraybackslash}p{0.19\linewidth}@{}}
\caption{Video generation models (64 entries).}\label{tab:models-video}\\
\toprule
Model & Registry ID & Readiness \\
\midrule
\endfirsthead
\multicolumn{3}{l}{\textit{Table \thetable{} continued: video generation models}}\\
\toprule
Model & Registry ID & Readiness \\
\midrule
\endhead
\midrule
\multicolumn{3}{r}{\textit{Continued on next page}}\\
\endfoot
\bottomrule
\endlastfoot
% model-id: allegro
\href{https://arxiv.org/abs/2410.15458}{Allegro} & \nolinkurl{allegro} & Runner verified \\
% model-id: animatediff
\href{https://arxiv.org/abs/2307.04725}{AnimateDiff} & \nolinkurl{animatediff} & Runner verified \\
% model-id: bernini
\href{https://arxiv.org/abs/2605.22344}{Bernini} & \nolinkurl{bernini} & Runner verified \\
% model-id: bernini-r-1.3b
\href{https://arxiv.org/abs/2605.22344}{Bernini-R 1.3B} & \nolinkurl{bernini-r-1.3b} & Runner verified \\
% model-id: bernini-r-14b
\href{https://arxiv.org/abs/2605.22344}{Bernini-R 14B} & \nolinkurl{bernini-r-14b} & Runner verified \\
% model-id: cogvideox
\href{https://arxiv.org/abs/2408.06072}{CogVideoX} & \nolinkurl{cogvideox} & Runner verified \\
% model-id: dynamicrafter
\href{https://arxiv.org/abs/2310.12190}{DynamiCrafter} & \nolinkurl{dynamicrafter} & Runner verified \\
% model-id: easyanimate
\href{https://arxiv.org/abs/2405.18991}{EasyAnimate} & \nolinkurl{easyanimate} & Runner verified \\
% model-id: emu3.5
\href{https://arxiv.org/abs/2510.26583}{Emu3.5} & \nolinkurl{emu3.5} & Runner verified \\
% model-id: fastvideo-causal-wan2.2
\href{https://huggingface.co/FastVideo/CausalWan2.2-I2V-A14B-Preview-Diffusers}{FastVideo CausalWan2.2} & \nolinkurl{fastvideo-causal-wan2.2} & Runner verified \\
% model-id: framepack
\href{https://arxiv.org/abs/2504.12626}{FramePack} & \nolinkurl{framepack} & Runner verified \\
% model-id: helios
\href{https://arxiv.org/abs/2603.04379}{Helios} & \nolinkurl{helios} & Runner verified \\
% model-id: hunyuanvideo
\href{https://arxiv.org/abs/2412.03603}{HunyuanVideo} & \nolinkurl{hunyuanvideo} & Runner verified \\
% model-id: hunyuanvideo-1.5
\href{https://arxiv.org/abs/2511.18870}{HunyuanVideo 1.5} & \nolinkurl{hunyuanvideo-1.5} & Runner verified \\
% model-id: hyperflow
\href{https://github.com/Video-Rebirth/hyperflow}{HyperFlow for MiniMax-H3} & \nolinkurl{hyperflow} & Runner verified \\
% model-id: i2vgen-xl
\href{https://arxiv.org/abs/2311.04145}{I2VGen-XL} & \nolinkurl{i2vgen-xl} & Runner verified \\
% model-id: joyai-echo-longvideo
\href{https://arxiv.org/abs/2608.23383}{JoyAI-Echo 1.5 / Echo-LongVideo} & \nolinkurl{joyai-echo-longvideo} & Runner verified \\
% model-id: kandinsky6-lite
\href{https://arxiv.org/abs/2610.05608}{Kandinsky 6.0 lite} & \nolinkurl{kandinsky6-lite} & Runner verified \\
% model-id: kandinsky6-lite-distill
\href{https://arxiv.org/abs/2610.05608}{Kandinsky 6.0 lite-distill} & \nolinkurl{kandinsky6-lite-distill} & Runner verified \\
% model-id: kandinsky6-lite-pretrain
\href{https://arxiv.org/abs/2610.05608}{Kandinsky 6.0 lite-pretrain} & \nolinkurl{kandinsky6-lite-pretrain} & Runner verified \\
% model-id: kandinsky6-pro
\href{https://arxiv.org/abs/2610.05608}{Kandinsky 6.0 pro} & \nolinkurl{kandinsky6-pro} & Runner verified \\
% model-id: kandinsky6-pro-distill
\href{https://arxiv.org/abs/2610.05608}{Kandinsky 6.0 pro-distill} & \nolinkurl{kandinsky6-pro-distill} & Runner verified \\
% model-id: kandinsky6-pro-pretrain
\href{https://arxiv.org/abs/2610.05608}{Kandinsky 6.0 pro-pretrain} & \nolinkurl{kandinsky6-pro-pretrain} & Runner verified \\
% model-id: krea-realtime-video
\href{https://github.com/krea-ai/realtime-video}{Krea Realtime Video 14B} & \nolinkurl{krea-realtime-video} & Runner verified \\
% model-id: lingbot-video
\href{https://arxiv.org/abs/2607.07675}{LingBot-Video} & \nolinkurl{lingbot-video} & Runner verified \\
% model-id: longcat-video
\href{https://arxiv.org/abs/2510.22200}{LongCat-Video} & \nolinkurl{longcat-video} & Runner verified \\
% model-id: longvie-1
\href{https://arxiv.org/abs/2508.03694}{LongVie} & \nolinkurl{longvie-1} & Runner verified \\
% model-id: longvie-2
\href{https://arxiv.org/abs/2512.13604}{LongVie 2} & \nolinkurl{longvie-2} & Runner verified \\
% model-id: ltx-2.5
\href{https://github.com/Lightricks/LTX-2}{LTX-2.5 Distilled} & \nolinkurl{ltx-2.5} & Runner verified \\
% model-id: ltx-2.x
\href{https://arxiv.org/abs/2601.03233}{LTX-2.x} & \nolinkurl{ltx-2.x} & Runner verified \\
% model-id: ltx-video
\href{https://arxiv.org/abs/2501.00103}{LTX-Video} & \nolinkurl{ltx-video} & Runner verified \\
% model-id: magi-1
\href{https://arxiv.org/abs/2505.13211}{MAGI-1} & \nolinkurl{magi-1} & Runner verified \\
% model-id: magi2-preview
\href{https://github.com/SandAI-org/MAGI-2-preview}{MAGI-2 Preview} & \nolinkurl{magi2-preview} & Runner verified \\
% model-id: minimax-h3
\href{https://www.minimax.cn/blog/minimax-h3}{MiniMax H3} & \nolinkurl{minimax-h3} & Runner verified \\
% model-id: mmaudio
\href{https://arxiv.org/abs/2412.15322}{MMAudio} & \nolinkurl{mmaudio} & Runner verified \\
% model-id: mochi-1
\href{https://github.com/genmoai/models}{Mochi-1} & \nolinkurl{mochi-1} & Runner verified \\
% model-id: modelscope-t2v
\href{https://arxiv.org/abs/2308.06571}{ModelScopeT2V} & \nolinkurl{modelscope-t2v} & Runner verified \\
% model-id: omnivinci
\href{https://arxiv.org/abs/2510.15870}{OmniVinci} & \nolinkurl{omnivinci} & Runner verified \\
% model-id: open-magvit2
\href{https://arxiv.org/abs/2409.04410}{Open-MAGVIT2} & \nolinkurl{open-magvit2} & Runner verified \\
% model-id: open-sora
\href{https://arxiv.org/abs/2503.09642}{Open-Sora} & \nolinkurl{open-sora} & Runner verified \\
% model-id: open-sora-plan
\href{https://arxiv.org/abs/2412.00131}{Open-Sora-Plan} & \nolinkurl{open-sora-plan} & Runner verified \\
% model-id: pusa-vidgen
\href{https://arxiv.org/abs/2410.03160}{Pusa VidGen} & \nolinkurl{pusa-vidgen} & Runner verified \\
% model-id: qwen2.5-omni
\href{https://arxiv.org/abs/2503.20215}{Qwen2.5-Omni} & \nolinkurl{qwen2.5-omni} & Runner verified \\
% model-id: sama-14b
\href{https://arxiv.org/abs/2603.19228}{SAMA} & \nolinkurl{sama-14b} & Runner verified \\
% model-id: sana
\href{https://arxiv.org/abs/2410.10629}{Sana} & \nolinkurl{sana} & Runner verified \\
% model-id: show-o
\href{https://arxiv.org/abs/2408.12528}{Show-o} & \nolinkurl{show-o} & Runner verified \\
% model-id: skyreels-v3
\href{https://arxiv.org/abs/2601.17323}{SkyReels V3} & \nolinkurl{skyreels-v3} & Runner verified \\
% model-id: skyreels-v2
\href{https://arxiv.org/abs/2504.13074}{SkyReels-V2} & \nolinkurl{skyreels-v2} & Runner verified \\
% model-id: spatial-ladder
\href{https://arxiv.org/abs/2510.08531}{SpatialLadder} & \nolinkurl{spatial-ladder} & Runner verified \\
% model-id: spatial-reasoner
\href{https://huggingface.co/ccvl/SpatialReasoner}{SpatialReasoner} & \nolinkurl{spatial-reasoner} & Runner verified \\
% model-id: stable-video-infinity
\href{https://arxiv.org/abs/2510.09212}{Stable Video Infinity 2.0} & \nolinkurl{stable-video-infinity} & Runner verified \\
% model-id: step-video-t2v
\href{https://arxiv.org/abs/2502.10248}{Step-Video-T2V} & \nolinkurl{step-video-t2v} & Runner verified \\
% model-id: t2v_turbo_t2v
\href{https://arxiv.org/abs/2405.18750}{T2V Turbo T2V} & \nolinkurl{t2v_turbo_t2v} & Runner verified \\
% model-id: unianimate-dit
\href{https://arxiv.org/abs/2504.11289}{UniAnimate-DiT} & \nolinkurl{unianimate-dit} & Runner verified \\
% model-id: vchitect-2-t2v
\href{https://arxiv.org/abs/2501.08453}{Vchitect-2.0 T2V} & \nolinkurl{vchitect-2-t2v} & Runner verified \\
% model-id: videocrafter
\href{https://arxiv.org/abs/2401.09047}{VideoCrafter} & \nolinkurl{videocrafter} & Runner verified \\
% model-id: videocrafter1-i2v
\href{https://arxiv.org/abs/2310.19512}{VideoCrafter1 I2V} & \nolinkurl{videocrafter1-i2v} & Runner verified \\
% model-id: videocrafter1-t2v
\href{https://arxiv.org/abs/2310.19512}{VideoCrafter1 T2V} & \nolinkurl{videocrafter1-t2v} & Runner verified \\
% model-id: videocrafter2-t2v
\href{https://arxiv.org/abs/2401.09047}{VideoCrafter2 T2V} & \nolinkurl{videocrafter2-t2v} & Runner verified \\
% model-id: vmem
\href{https://arxiv.org/abs/2506.18903}{VMem} & \nolinkurl{vmem} & Runner verified \\
% model-id: wan2.1
\href{https://arxiv.org/abs/2503.20314}{Wan2.1} & \nolinkurl{wan2.1} & Runner verified \\
% model-id: wan2.1-vace
\href{https://arxiv.org/abs/2503.07598}{Wan2.1 VACE} & \nolinkurl{wan2.1-vace} & Runner verified \\
% model-id: wan2.2
\href{https://github.com/Wan-Video/Wan2.2}{Wan2.2} & \nolinkurl{wan2.2} & Runner verified \\
% model-id: zeroscope
\href{https://huggingface.co/cerspense/zeroscope_v2_576w}{ZeroScope} & \nolinkurl{zeroscope} & Runner verified \\
\end{longtable}
\endgroup

\begingroup
\footnotesize
\setlength{\tabcolsep}{4pt}
\renewcommand{\arraystretch}{1.08}
\begin{longtable}{@{}>{\raggedright\arraybackslash}p{0.43\linewidth}>{\raggedright\arraybackslash}p{0.31\linewidth}>{\raggedright\arraybackslash}p{0.19\linewidth}@{}}
\caption{World models (109 entries).}\label{tab:models-world}\\
\toprule
Model & Registry ID & Readiness \\
\midrule
\endfirsthead
\multicolumn{3}{l}{\textit{Table \thetable{} continued: world models}}\\
\toprule
Model & Registry ID & Readiness \\
\midrule
\endhead
\midrule
\multicolumn{3}{r}{\textit{Continued on next page}}\\
\endfoot
\bottomrule
\endlastfoot
% model-id: abot-world-0-5b-lf
\href{https://arxiv.org/abs/2607.19191}{ABot-World-0-5B-LF} & \nolinkurl{abot-world-0-5b-lf} & Runner verified \\
% model-id: ac3d
\href{https://arxiv.org/abs/2411.18673}{AC3D} & \nolinkurl{ac3d} & Runner verified \\
% model-id: adaworld
\href{https://arxiv.org/abs/2503.18938}{AdaWorld} & \nolinkurl{adaworld} & Runner verified \\
% model-id: alayaworld
\href{https://arxiv.org/abs/2607.06291}{AlayaWorld} & \nolinkurl{alayaworld} & Runner verified \\
% model-id: alayaworld-v1.1
\href{https://arxiv.org/abs/2608.13492}{AlayaWorld v1.1} & \nolinkurl{alayaworld-v1.1} & Runner verified \\
% model-id: astra
\href{https://arxiv.org/abs/2512.08931}{Astra} & \nolinkurl{astra} & Runner verified \\
% model-id: astronex-world
\href{https://world.astronex.com.cn/assets/paper/astronex-world-technical-report.pdf}{Astronex-World} & \nolinkurl{astronex-world} & Runner verified \\
% model-id: ati-wan21-14b
\href{https://arxiv.org/abs/2505.22944}{ATI-Wan2.1-14B} & \nolinkurl{ati-wan21-14b} & Runner verified \\
% model-id: biwm-wan21
\href{https://arxiv.org/abs/2606.10135}{BiWM Wan2.1} & \nolinkurl{biwm-wan21} & Runner verified \\
% model-id: biwm-wan22
\href{https://arxiv.org/abs/2606.10135}{BiWM Wan2.2} & \nolinkurl{biwm-wan22} & Runner verified \\
% model-id: cameractrl
\href{https://arxiv.org/abs/2404.02101}{CameraCtrl} & \nolinkurl{cameractrl} & Runner verified \\
% model-id: causal-forcing
\href{https://arxiv.org/abs/2602.02214}{Causal-Forcing} & \nolinkurl{causal-forcing} & Runner verified \\
% model-id: causal-rcm
\href{https://arxiv.org/abs/2510.08431}{Causal-rCM} & \nolinkurl{causal-rcm} & Runner verified \\
% model-id: cosmos-predict-2
\href{https://arxiv.org/abs/2501.03575}{Cosmos Predict 2} & \nolinkurl{cosmos-predict-2} & Runner verified \\
% model-id: cosmos-predict-2.5
\href{https://arxiv.org/abs/2511.00062}{Cosmos Predict 2.5} & \nolinkurl{cosmos-predict-2.5} & Runner verified \\
% model-id: cosmos-transfer-2.5
\href{https://arxiv.org/abs/2511.00062}{Cosmos Transfer 2.5} & \nolinkurl{cosmos-transfer-2.5} & Runner verified \\
% model-id: cosmos3
\href{https://arxiv.org/abs/2606.02800}{Cosmos3} & \nolinkurl{cosmos3} & Runner verified \\
% model-id: ctrl-world
\href{https://arxiv.org/abs/2510.10125}{Ctrl-World} & \nolinkurl{ctrl-world} & Runner verified \\
% model-id: diamond
\href{https://arxiv.org/abs/2405.12399}{DIAMOND} & \nolinkurl{diamond} & Runner verified \\
% model-id: dino-wm
\href{https://arxiv.org/abs/2411.04983}{DINO-WM} & \nolinkurl{dino-wm} & Runner verified \\
% model-id: dreamdojo
\href{https://arxiv.org/abs/2602.06949}{DreamDojo} & \nolinkurl{dreamdojo} & Runner verified \\
% model-id: dreamx-world-5b
\href{https://arxiv.org/abs/2606.16993}{DreamX-World 5B} & \nolinkurl{dreamx-world-5b} & Runner verified \\
% model-id: dualcamctrl
\href{https://arxiv.org/abs/2511.23127}{DualCamCtrl} & \nolinkurl{dualcamctrl} & Runner verified \\
% model-id: echo-infinity
\href{https://arxiv.org/abs/2606.04527}{Echo-Infinity} & \nolinkurl{echo-infinity} & Runner verified \\
% model-id: echo-memory-context-k1
\href{https://arxiv.org/abs/2606.09803}{Echo-Memory} & \nolinkurl{echo-memory-context-k1} & Runner verified \\
% model-id: joyai-echo-wm
\href{https://arxiv.org/abs/2608.23189}{Echo-WM} & \nolinkurl{joyai-echo-wm} & Runner verified \\
% model-id: egowm
\href{https://arxiv.org/abs/2601.15284}{EgoWM} & \nolinkurl{egowm} & Runner verified \\
% model-id: evoke
\href{https://arxiv.org/abs/2608.13546}{Evoke} & \nolinkurl{evoke} & Runner verified \\
% model-id: gamma-world
\href{https://arxiv.org/abs/2605.28816}{Gamma-World} & \nolinkurl{gamma-world} & Runner verified \\
% model-id: gen3c
\href{https://arxiv.org/abs/2503.03751}{GEN3C} & \nolinkurl{gen3c} & Runner verified \\
% model-id: genie-envisioner
\href{https://arxiv.org/abs/2508.05635}{Genie Envisioner} & \nolinkurl{genie-envisioner} & Runner verified \\
% model-id: giga-world-0
\href{https://arxiv.org/abs/2511.19861}{GigaWorld-0} & \nolinkurl{giga-world-0} & Runner verified \\
% model-id: happyoyster
\href{https://github.com/OpenEnvision/WorldFoundry/blob/main/worldfoundry/data/models/catalog/world_models/happyoyster.yaml}{HappyOyster} & \nolinkurl{happyoyster} & Runner verified \\
% model-id: helixworld
\href{https://arxiv.org/abs/2609.38123}{HelixWorld Preview v1} & \nolinkurl{helixworld} & Runner verified \\
% model-id: hma
\href{https://arxiv.org/abs/2502.04296}{HMA} & \nolinkurl{hma} & Runner verified \\
% model-id: hunyuan-game-craft
\href{https://arxiv.org/abs/2506.17201}{Hunyuan Game Craft} & \nolinkurl{hunyuan-game-craft} & Runner verified \\
% model-id: hunyuanworld-1
\href{https://arxiv.org/abs/2507.21809}{HunyuanWorld 1.0} & \nolinkurl{hunyuanworld-1} & Runner verified \\
% model-id: hunyuanworld-mirror
\href{https://arxiv.org/abs/2510.10726}{HunyuanWorld-Mirror} & \nolinkurl{hunyuanworld-mirror} & Runner verified \\
% model-id: hunyuanworld-voyager
\href{https://arxiv.org/abs/2506.04225}{HunyuanWorld-Voyager} & \nolinkurl{hunyuanworld-voyager} & Runner verified \\
% model-id: hy-world-2.0
\href{https://arxiv.org/abs/2604.14268}{HY-World 2.0} & \nolinkurl{hy-world-2.0} & Runner verified \\
% model-id: hyworld-worldgen
\href{https://arxiv.org/abs/2604.14268}{HY-World 2.0 Trajectory Renderer} & \nolinkurl{hyworld-worldgen} & Runner verified \\
% model-id: hy-worldplay
\href{https://arxiv.org/abs/2512.14614}{HY-WorldPlay} & \nolinkurl{hy-worldplay} & Runner verified \\
% model-id: hydra
\href{https://arxiv.org/abs/2603.25716}{HyDRA} & \nolinkurl{hydra} & Runner verified \\
% model-id: infinite-world
\href{https://arxiv.org/abs/2602.02393}{Infinite-World} & \nolinkurl{infinite-world} & Runner verified \\
% model-id: inspatio-world-1p5
\href{https://arxiv.org/abs/2604.07209}{InSpatio World 1.5} & \nolinkurl{inspatio-world-1p5} & Runner verified \\
% model-id: inspatio-world
\href{https://arxiv.org/abs/2604.07209}{InSpatio-World} & \nolinkurl{inspatio-world} & Runner verified \\
% model-id: worldfm
\href{https://arxiv.org/abs/2603.11911}{InSpatio-WorldFM} & \nolinkurl{worldfm} & Runner verified \\
% model-id: irasim
\href{https://arxiv.org/abs/2406.14540}{IRASim} & \nolinkurl{irasim} & Runner verified \\
% model-id: kairos-sensenova
\href{https://arxiv.org/abs/2606.16533}{Kairos Sensenova} & \nolinkurl{kairos-sensenova} & Runner verified \\
% model-id: leworldmodel
\href{https://arxiv.org/abs/2603.19312}{LeWorldModel} & \nolinkurl{leworldmodel} & Runner verified \\
% model-id: lingbot-world
\href{https://arxiv.org/abs/2601.20540}{LingBot-World} & \nolinkurl{lingbot-world} & Runner verified \\
% model-id: lingbot-world-act
\href{https://arxiv.org/abs/2601.20540}{LingBot-World Base Act} & \nolinkurl{lingbot-world-act} & Runner verified \\
% model-id: lingbot-world-v2
\href{https://arxiv.org/abs/2607.07534}{LingBot-World-V2} & \nolinkurl{lingbot-world-v2} & Runner verified \\
% model-id: liveworld
\href{https://arxiv.org/abs/2603.07145}{LiveWorld} & \nolinkurl{liveworld} & Runner verified \\
% model-id: magicworld
\href{https://arxiv.org/abs/2511.18886}{MagicWorld} & \nolinkurl{magicworld} & Runner verified \\
% model-id: matrix-game-1
\href{https://arxiv.org/abs/2506.18701}{Matrix-Game 1.0} & \nolinkurl{matrix-game-1} & Runner verified \\
% model-id: matrix-game-2
\href{https://arxiv.org/abs/2508.13009}{Matrix-Game 2.0} & \nolinkurl{matrix-game-2} & Runner verified \\
% model-id: matrix-game-3
\href{https://arxiv.org/abs/2604.08995}{Matrix-Game 3.0} & \nolinkurl{matrix-game-3} & Runner verified \\
% model-id: matrix-game-3.5-first-person
\href{https://arxiv.org/abs/2608.29910}{Matrix-Game 3.5} & \nolinkurl{matrix-game-3.5-first-person} & Runner verified \\
% model-id: mineworld
\href{https://arxiv.org/abs/2504.08388}{MineWorld} & \nolinkurl{mineworld} & Runner verified \\
% model-id: minwm-hy-action2v
\href{https://arxiv.org/abs/2605.30263}{minWM} & \nolinkurl{minwm-hy-action2v} & Runner verified \\
% model-id: mira
\href{https://arxiv.org/abs/2607.05352}{MIRA} & \nolinkurl{mira} & Runner verified \\
% model-id: mosaicmem
\href{https://arxiv.org/abs/2603.17117}{MosaicMem} & \nolinkurl{mosaicmem} & Runner verified \\
% model-id: motionbricks
\href{https://arxiv.org/abs/2604.24833}{MotionBricks} & \nolinkurl{motionbricks} & Runner verified \\
% model-id: motionctrl
\href{https://arxiv.org/abs/2312.03641}{MotionCtrl} & \nolinkurl{motionctrl} & Runner verified \\
% model-id: moverse
\href{https://arxiv.org/abs/2606.13376}{MoVerse} & \nolinkurl{moverse} & Runner verified \\
% model-id: multiworld
\href{https://arxiv.org/abs/2604.18564}{MultiWorld} & \nolinkurl{multiworld} & Runner verified \\
% model-id: nwm
\href{https://arxiv.org/abs/2412.03572}{Navigation World Models} & \nolinkurl{nwm} & Runner verified \\
% model-id: oasis-500m
\href{https://github.com/etched-ai/open-oasis}{Oasis 500M} & \nolinkurl{oasis-500m} & Runner verified \\
% model-id: omniforcing
\href{https://arxiv.org/abs/2603.11647}{OmniForcing} & \nolinkurl{omniforcing} & Runner verified \\
% model-id: open-dreamer
\href{https://arxiv.org/abs/2509.24527}{Open Dreamer} & \nolinkurl{open-dreamer} & Runner verified \\
% model-id: pandora
\href{https://arxiv.org/abs/2406.09455}{Pandora} & \nolinkurl{pandora} & Runner verified \\
% model-id: pointworld
\href{https://arxiv.org/abs/2601.03782}{PointWorld} & \nolinkurl{pointworld} & Runner verified \\
% model-id: rolling-forcing
\href{https://arxiv.org/abs/2509.25161}{RollingForcing} & \nolinkurl{rolling-forcing} & Runner verified \\
% model-id: sana-wm
\href{https://arxiv.org/abs/2605.15178}{SANA-WM} & \nolinkurl{sana-wm} & Runner verified \\
% model-id: scope
\href{https://arxiv.org/abs/2605.23345}{SCOPE} & \nolinkurl{scope} & Runner verified \\
% model-id: self-forcing
\href{https://arxiv.org/abs/2506.08009}{Self-Forcing} & \nolinkurl{self-forcing} & Runner verified \\
% model-id: shotstream
\href{https://arxiv.org/abs/2603.25746}{ShotStream} & \nolinkurl{shotstream} & Runner verified \\
% model-id: simworld
\href{https://arxiv.org/abs/2512.01078}{SimWorld} & \nolinkurl{simworld} & Runner verified \\
% model-id: solaris
\href{https://arxiv.org/abs/2602.22208}{Solaris} & \nolinkurl{solaris} & Runner verified \\
% model-id: solarwm
\href{https://arxiv.org/abs/2609.02886}{SolarWM} & \nolinkurl{solarwm} & Runner verified \\
% model-id: spatia
\href{https://arxiv.org/abs/2512.15716}{Spatia} & \nolinkurl{spatia} & Runner verified \\
% model-id: starwm
\href{https://arxiv.org/abs/2602.14857}{StarWM} & \nolinkurl{starwm} & Runner verified \\
% model-id: tesseract
\href{https://arxiv.org/abs/2504.20995}{TesserAct} & \nolinkurl{tesseract} & Runner verified \\
% model-id: uni3c
\href{https://arxiv.org/abs/2504.14899}{Uni3C} & \nolinkurl{uni3c} & Runner verified \\
% model-id: uwm
\href{https://arxiv.org/abs/2504.02792}{Unified World Models} & \nolinkurl{uwm} & Runner verified \\
% model-id: uniworld-view
\href{https://arxiv.org/abs/2608.04701}{UniWorld-View} & \nolinkurl{uniworld-view} & Runner verified \\
% model-id: versecrafter
\href{https://arxiv.org/abs/2601.05138}{VerseCrafter} & \nolinkurl{versecrafter} & Runner verified \\
% model-id: vggt-world
\href{https://arxiv.org/abs/2603.12655}{VGGT-World} & \nolinkurl{vggt-world} & Runner verified \\
% model-id: vid2world
\href{https://arxiv.org/abs/2505.14357}{Vid2World} & \nolinkurl{vid2world} & Runner verified \\
% model-id: wan21-fun-1p3b-cam
\href{https://github.com/aigc-apps/VideoX-Fun}{Wan2.1-Fun V1.1 1.3B Control Camera} & \nolinkurl{wan21-fun-1p3b-cam} & Runner verified \\
% model-id: wan21-fun-14b-cam
\href{https://github.com/aigc-apps/VideoX-Fun}{Wan2.1-Fun V1.1 14B Control Camera} & \nolinkurl{wan21-fun-14b-cam} & Runner verified \\
% model-id: wan22-fun-5b-cam
\href{https://github.com/aigc-apps/VideoX-Fun}{Wan2.2-Fun 5B Control Camera} & \nolinkurl{wan22-fun-5b-cam} & Runner verified \\
% model-id: wan22-fun-a14b-cam
\href{https://github.com/aigc-apps/VideoX-Fun}{Wan2.2-Fun A14B Control Camera} & \nolinkurl{wan22-fun-a14b-cam} & Runner verified \\
% model-id: warp-as-history
\href{https://arxiv.org/abs/2605.15182}{Warp-as-History} & \nolinkurl{warp-as-history} & Runner verified \\
% model-id: wilddet3d
\href{https://arxiv.org/abs/2604.08626}{WildDet3D} & \nolinkurl{wilddet3d} & Runner verified \\
% model-id: wildworld
\href{https://arxiv.org/abs/2603.23497}{WildWorld} & \nolinkurl{wildworld} & Runner verified \\
% model-id: worldlabs-marble-1.1
\href{https://docs.worldlabs.ai/marble/models}{World Labs Marble 1.1} & \nolinkurl{worldlabs-marble-1.1} & Runner verified \\
% model-id: worldcam
\href{https://arxiv.org/abs/2603.16871}{WorldCam} & \nolinkurl{worldcam} & Runner verified \\
% model-id: worldcrafter
\href{https://arxiv.org/abs/2609.24984}{WorldCrafter} & \nolinkurl{worldcrafter} & Runner verified \\
% model-id: worldcrafter-base
\href{https://arxiv.org/abs/2609.24984}{WorldCrafter Base} & \nolinkurl{worldcrafter-base} & Runner verified \\
% model-id: worldcrafter-fast
\href{https://arxiv.org/abs/2609.24984}{WorldCrafter Fast} & \nolinkurl{worldcrafter-fast} & Runner verified \\
% model-id: worldgrow
\href{https://arxiv.org/abs/2510.21682}{WorldGrow} & \nolinkurl{worldgrow} & Runner verified \\
% model-id: worldmem
\href{https://arxiv.org/abs/2504.12369}{WorldMem} & \nolinkurl{worldmem} & Runner verified \\
% model-id: worldplay2
\href{https://arxiv.org/abs/2609.35560}{WorldPlay2} & \nolinkurl{worldplay2} & Runner verified \\
% model-id: wow
\href{https://arxiv.org/abs/2509.22642}{WoW} & \nolinkurl{wow} & Runner verified \\
% model-id: xgen-jing
\href{https://github.com/XGEN-Labs/XGEN-JING}{XGEN-JING Flash v1} & \nolinkurl{xgen-jing} & Runner verified \\
% model-id: yume
\href{https://arxiv.org/abs/2507.17744}{YUME} (\href{https://arxiv.org/abs/2512.22096}{1.5 paper}) & \nolinkurl{yume} & Runner verified \\
% model-id: zing
\href{https://arxiv.org/abs/2609.17909}{Zing-0.5} & \nolinkurl{zing} & Runner verified \\
\end{longtable}
\endgroup

\begingroup
\footnotesize
\setlength{\tabcolsep}{4pt}
\renewcommand{\arraystretch}{1.08}
\begin{longtable}{@{}>{\raggedright\arraybackslash}p{0.43\linewidth}>{\raggedright\arraybackslash}p{0.31\linewidth}>{\raggedright\arraybackslash}p{0.19\linewidth}@{}}
\caption{3D and 4D models (34 entries).}\label{tab:models-3d4d}\\
\toprule
Model & Registry ID & Readiness \\
\midrule
\endfirsthead
\multicolumn{3}{l}{\textit{Table \thetable{} continued: 3d and 4d models}}\\
\toprule
Model & Registry ID & Readiness \\
\midrule
\endhead
\midrule
\multicolumn{3}{r}{\textit{Continued on next page}}\\
\endfoot
\bottomrule
\endlastfoot
% model-id: cut3r
\href{https://arxiv.org/abs/2501.12387}{CUT3R} & \nolinkurl{cut3r} & Runner verified \\
% model-id: dap
\href{https://arxiv.org/abs/2512.16913}{Depth Any Panoramas} & \nolinkurl{dap} & Runner verified \\
% model-id: depth-anything-v3
\href{https://arxiv.org/abs/2511.10647}{Depth Anything 3} & \nolinkurl{depth-anything-v3} & Runner verified \\
% model-id: depth-anything-v3-prior
\href{https://arxiv.org/abs/2511.10647}{Depth Anything 3 Metric Prior} & \nolinkurl{depth-anything-v3-prior} & Runner verified \\
% model-id: depth-anything-v1
\href{https://arxiv.org/abs/2401.10891}{Depth Anything V1} & \nolinkurl{depth-anything-v1} & Runner verified \\
% model-id: depth-anything-v2-prior
\href{https://arxiv.org/abs/2406.09414}{Depth Anything V2} & \nolinkurl{depth-anything-v2-prior} & Runner verified \\
% model-id: dust3r
\href{https://arxiv.org/abs/2312.14132}{DUSt3R} & \nolinkurl{dust3r} & Runner verified \\
% model-id: dvlt
\href{https://arxiv.org/abs/2605.30215}{D\'ej\`a View Looping Transformer} & \nolinkurl{dvlt} & Runner verified \\
% model-id: fantasyworld
\href{https://arxiv.org/abs/2509.21657}{FantasyWorld} & \nolinkurl{fantasyworld} & Runner verified \\
% model-id: flashworld
\href{https://arxiv.org/abs/2510.13678}{FlashWorld} & \nolinkurl{flashworld} & Runner verified \\
% model-id: geocalib-prior
\href{https://arxiv.org/abs/2409.06704}{GeoCalib} & \nolinkurl{geocalib-prior} & Runner verified \\
% model-id: infinite-vggt
\href{https://arxiv.org/abs/2507.11539}{Infinite VGGT} & \nolinkurl{infinite-vggt} & Runner verified \\
% model-id: lagernvs
\href{https://arxiv.org/abs/2603.20176}{LagrNVS} & \nolinkurl{lagernvs} & Runner verified \\
% model-id: lingbot-map
\href{https://arxiv.org/abs/2604.14141}{LingBot-Map} & \nolinkurl{lingbot-map} & Runner verified \\
% model-id: loger
\href{https://arxiv.org/abs/2603.03269}{LoGeR} & \nolinkurl{loger} & Runner verified \\
% model-id: lyra
\href{https://research.nvidia.com/labs/toronto-ai/lyra/}{Lyra} (\href{https://arxiv.org/abs/2604.13036}{2.0 paper}) & \nolinkurl{lyra} & Runner verified \\
% model-id: metric3d-prior
\href{https://arxiv.org/abs/2404.15506}{Metric3D} & \nolinkurl{metric3d-prior} & Runner verified \\
% model-id: monst3r
\href{https://arxiv.org/abs/2410.03825}{MonST3R} & \nolinkurl{monst3r} & Runner verified \\
% model-id: mvdiffusion
\href{https://arxiv.org/abs/2307.01097}{MVDiffusion} & \nolinkurl{mvdiffusion} & Runner verified \\
% model-id: neoverse
\href{https://arxiv.org/abs/2601.00393}{NeoVerse} & \nolinkurl{neoverse} & Runner verified \\
% model-id: pi3
\href{https://arxiv.org/abs/2507.13347}{Pi3} & \nolinkurl{pi3} & Runner verified \\
% model-id: pixelsplat
\href{https://arxiv.org/abs/2312.12337}{pixelSplat} & \nolinkurl{pixelsplat} & Runner verified \\
% model-id: prior-depth-anything
\href{https://arxiv.org/abs/2505.10565}{Prior Depth Anything} & \nolinkurl{prior-depth-anything} & Runner verified \\
% model-id: recammaster
\href{https://arxiv.org/abs/2503.11647}{ReCamMaster} & \nolinkurl{recammaster} & Runner verified \\
% model-id: track-anything-prior
\href{https://arxiv.org/abs/2305.06558}{Segment and Track Anything} & \nolinkurl{track-anything-prior} & Runner verified \\
% model-id: splatt3r
\href{https://arxiv.org/abs/2408.13912}{Splatt3R} & \nolinkurl{splatt3r} & Runner verified \\
% model-id: stable-virtual-camera
\href{https://arxiv.org/abs/2503.14489}{Stable Virtual Camera} & \nolinkurl{stable-virtual-camera} & Runner verified \\
% model-id: unidepth-v2-prior
\href{https://arxiv.org/abs/2502.20110}{UniDepth V2} & \nolinkurl{unidepth-v2-prior} & Runner verified \\
% model-id: unik3d-prior
\href{https://arxiv.org/abs/2503.16591}{UniK3D} & \nolinkurl{unik3d-prior} & Runner verified \\
% model-id: vggt
\href{https://arxiv.org/abs/2503.11651}{VGGT} & \nolinkurl{vggt} & Runner verified \\
% model-id: vggt-omega
\href{https://arxiv.org/abs/2605.15195}{VGGT-Omega} & \nolinkurl{vggt-omega} & Runner verified \\
% model-id: video-depth-anything-prior
\href{https://arxiv.org/abs/2501.12375}{Video Depth Anything} & \nolinkurl{video-depth-anything-prior} & Runner verified \\
% model-id: wonderjourney
\href{https://arxiv.org/abs/2312.03884}{WonderJourney} & \nolinkurl{wonderjourney} & Runner verified \\
% model-id: worldgen
\href{https://arxiv.org/abs/2511.16825}{WorldGen} & \nolinkurl{worldgen} & Runner verified \\
\end{longtable}
\endgroup

\begingroup
\footnotesize
\setlength{\tabcolsep}{4pt}
\renewcommand{\arraystretch}{1.08}
\begin{longtable}{@{}>{\raggedright\arraybackslash}p{0.43\linewidth}>{\raggedright\arraybackslash}p{0.31\linewidth}>{\raggedright\arraybackslash}p{0.19\linewidth}@{}}
\caption{Embodied models (53 entries).}\label{tab:models-embodied}\\
\toprule
Model & Registry ID & Readiness \\
\midrule
\endfirsthead
\multicolumn{3}{l}{\textit{Table \thetable{} continued: embodied models}}\\
\toprule
Model & Registry ID & Readiness \\
\midrule
\endhead
\midrule
\multicolumn{3}{r}{\textit{Continued on next page}}\\
\endfoot
\bottomrule
\endlastfoot
% model-id: a1
\href{https://arxiv.org/abs/2604.05672}{A1} & \nolinkurl{a1} & Runner verified \\
% model-id: abot-m0
\href{https://arxiv.org/abs/2602.11236}{ABot-M0} & \nolinkurl{abot-m0} & Runner verified \\
% model-id: act
\href{https://arxiv.org/abs/2304.13705}{Action Chunking Transformer} & \nolinkurl{act} & Runner verified \\
% model-id: ahawam
\href{https://arxiv.org/abs/2606.09811}{AHA-WAM} & \nolinkurl{ahawam} & Runner verified \\
% model-id: being-h05
\href{https://arxiv.org/abs/2601.12993}{Being-H0.5} & \nolinkurl{being-h05} & Runner verified \\
% model-id: being-h07
\href{https://arxiv.org/abs/2605.00078}{Being-H0.7} & \nolinkurl{being-h07} & Runner verified \\
% model-id: cogact
\href{https://arxiv.org/abs/2411.19650}{CogACT} & \nolinkurl{cogact} & Runner verified \\
% model-id: db-cogact
\href{https://arxiv.org/abs/2510.23511}{DB-CogACT / Dexbotic} & \nolinkurl{db-cogact} & Runner verified \\
% model-id: dexora-1b
\href{https://arxiv.org/abs/2605.18722}{Dexora 1B} & \nolinkurl{dexora-1b} & Runner verified \\
% model-id: diffusion-policy
\href{https://arxiv.org/abs/2303.04137}{Diffusion Policy} & \nolinkurl{diffusion-policy} & Runner verified \\
% model-id: dm0
\href{https://arxiv.org/abs/2510.23511}{DM0} & \nolinkurl{dm0} & Runner verified \\
% model-id: eventvla
\href{https://arxiv.org/abs/2606.20092}{EventVLA} & \nolinkurl{eventvla} & Runner verified \\
% model-id: fastwam
\href{https://arxiv.org/abs/2603.16666}{FastWAM} & \nolinkurl{fastwam} & Runner verified \\
% model-id: galaxea-vla
\href{https://arxiv.org/abs/2509.00576}{Galaxea G0Plus} & \nolinkurl{galaxea-vla} & Runner verified \\
% model-id: giga-brain-0
\href{https://arxiv.org/abs/2510.19430}{GigaBrain-0} & \nolinkurl{giga-brain-0} & Runner verified \\
% model-id: giga-world-policy-0.5
\href{https://arxiv.org/abs/2607.13960}{GigaWorld-Policy-0.5} & \nolinkurl{giga-world-policy-0.5} & Runner verified \\
% model-id: go1
\href{https://arxiv.org/abs/2503.06669}{GO-1} & \nolinkurl{go1} & Runner verified \\
% model-id: h-rdt
\href{https://arxiv.org/abs/2507.23523}{H-RDT} & \nolinkurl{h-rdt} & Runner verified \\
% model-id: hy-embodied-vla
\href{https://arxiv.org/abs/2606.14409}{Hy-Embodied-0.5-VLA} & \nolinkurl{hy-embodied-vla} & Runner verified \\
% model-id: internvla-a1
\href{https://arxiv.org/abs/2601.02456}{InternVLA-A1-3B} & \nolinkurl{internvla-a1} & Runner verified \\
% model-id: last-r1
\href{https://arxiv.org/abs/2604.28192}{LaST-R1} & \nolinkurl{last-r1} & Runner verified \\
% model-id: lapa
\href{https://arxiv.org/abs/2410.11758}{Latent Action Pretraining} & \nolinkurl{lapa} & Runner verified \\
% model-id: lda-1b
\href{https://arxiv.org/abs/2602.12215}{LDA-1B} & \nolinkurl{lda-1b} & Runner verified \\
% model-id: lingbot-va
\href{https://arxiv.org/abs/2601.21998}{LingBot-VA} & \nolinkurl{lingbot-va} & Runner verified \\
% model-id: lingbot-vla
\href{https://arxiv.org/abs/2601.18692}{LingBot-VLA} & \nolinkurl{lingbot-vla} & Runner verified \\
% model-id: lingbot-vla-v2
\href{https://arxiv.org/abs/2607.06403}{LingBot-VLA 2.0} & \nolinkurl{lingbot-vla-v2} & Runner verified \\
% model-id: mem-0
\href{https://arxiv.org/abs/2603.01229}{Mem-0} & \nolinkurl{mem-0} & Runner verified \\
% model-id: mme-vla
\href{https://arxiv.org/abs/2603.04639}{MME-VLA} & \nolinkurl{mme-vla} & Runner verified \\
% model-id: molmoact2
\href{https://arxiv.org/abs/2605.02881}{MolmoAct2} & \nolinkurl{molmoact2} & Runner verified \\
% model-id: molmobot
\href{https://arxiv.org/abs/2603.16861}{MolmoBot} & \nolinkurl{molmobot} & Runner verified \\
% model-id: dreamzero
\href{https://arxiv.org/abs/2602.15922}{NVIDIA DreamZero} & \nolinkurl{dreamzero} & Runner verified \\
% model-id: gr00t
\href{https://arxiv.org/abs/2503.14734}{NVIDIA GR00T} & \nolinkurl{gr00t} & Runner verified \\
% model-id: octo
\href{https://arxiv.org/abs/2405.12213}{Octo} & \nolinkurl{octo} & Runner verified \\
% model-id: openpi
\href{https://arxiv.org/abs/2504.16054}{OpenPI} & \nolinkurl{openpi} & Runner verified \\
% model-id: openvla
\href{https://arxiv.org/abs/2406.09246}{OpenVLA} & \nolinkurl{openvla} & Runner verified \\
% model-id: openvla-oft
\href{https://arxiv.org/abs/2502.19645}{OpenVLA-OFT} & \nolinkurl{openvla-oft} & Runner verified \\
% model-id: pi0
\href{https://arxiv.org/abs/2410.24164}{pi0} & \nolinkurl{pi0} & Runner verified \\
% model-id: pi0-fast
\href{https://arxiv.org/abs/2501.09747}{pi0-FAST} & \nolinkurl{pi0-fast} & Runner verified \\
% model-id: pi05
\href{https://arxiv.org/abs/2504.16054}{pi0.5} & \nolinkurl{pi05} & Runner verified \\
% model-id: rdt-1b
\href{https://arxiv.org/abs/2410.07864}{RDT-1B} & \nolinkurl{rdt-1b} & Runner verified \\
% model-id: rise
\href{https://arxiv.org/abs/2602.11075}{RISE} & \nolinkurl{rise} & Runner verified \\
% model-id: roboflamingo
\href{https://arxiv.org/abs/2311.01378}{RoboFlamingo} & \nolinkurl{roboflamingo} & Runner verified \\
% model-id: rt-1
\href{https://arxiv.org/abs/2212.06817}{RT-1} & \nolinkurl{rt-1} & Runner verified \\
% model-id: smolvla
\href{https://arxiv.org/abs/2506.01844}{SmolVLA} & \nolinkurl{smolvla} & Runner verified \\
% model-id: spatial-forcing
\href{https://arxiv.org/abs/2510.12276}{Spatial-Forcing} & \nolinkurl{spatial-forcing} & Runner verified \\
% model-id: spirit-v1.5
\href{https://www.spirit-ai.com/en/blog/spirit-v1-5}{Spirit-VLA v1.5} & \nolinkurl{spirit-v1.5} & Runner verified \\
% model-id: starvla
\href{https://arxiv.org/abs/2604.05014}{StarVLA} & \nolinkurl{starvla} & Runner verified \\
% model-id: tinyvla
\href{https://arxiv.org/abs/2409.12514}{TinyVLA} & \nolinkurl{tinyvla} & Runner verified \\
% model-id: vlanext
\href{https://arxiv.org/abs/2602.18532}{VLANeXt} & \nolinkurl{vlanext} & Runner verified \\
% model-id: xvla
\href{https://arxiv.org/abs/2510.10274}{X-VLA} & \nolinkurl{xvla} & Runner verified \\
% model-id: x-wam
\href{https://arxiv.org/abs/2604.26694}{X-WAM} & \nolinkurl{x-wam} & Runner verified \\
% model-id: xiaomi-robotics-0
\href{https://arxiv.org/abs/2602.12684}{Xiaomi-Robotics-0} & \nolinkurl{xiaomi-robotics-0} & Runner verified \\
% model-id: xiaomi-robotics-1
\href{https://arxiv.org/abs/2607.15330}{Xiaomi-Robotics-1} & \nolinkurl{xiaomi-robotics-1} & Runner verified \\
\end{longtable}
\endgroup

\begingroup
\footnotesize
\setlength{\tabcolsep}{4pt}
\renewcommand{\arraystretch}{1.08}
\begin{longtable}{@{}>{\raggedright\arraybackslash}p{0.43\linewidth}>{\raggedright\arraybackslash}p{0.31\linewidth}>{\raggedright\arraybackslash}p{0.19\linewidth}@{}}
\caption{Hosted API models (10 entries).}\label{tab:models-hosted-api}\\
\toprule
Model & Registry ID & Readiness \\
\midrule
\endfirsthead
\multicolumn{3}{l}{\textit{Table \thetable{} continued: hosted API models}}\\
\toprule
Model & Registry ID & Readiness \\
\midrule
\endhead
\midrule
\multicolumn{3}{r}{\textit{Continued on next page}}\\
\endfoot
\bottomrule
\endlastfoot
% model-id: hailuo-2p3
\href{https://www.minimax.io/news/minimax-hailuo-23}{Hailuo 2.3} & \nolinkurl{hailuo-2p3} & Runner verified \\
% model-id: kling-api
\href{https://kling.ai/document-api/apiReference/model/textToVideo}{Kling API} & \nolinkurl{kling-api} & Runner verified \\
% model-id: luma-ray2
\href{https://lumalabs.ai/changelog/introducing-ray2}{Luma Ray2} & \nolinkurl{luma-ray2} & Runner verified \\
% model-id: runway-gen4p5
\href{https://runway.com/research/introducing-runway-gen-4.5}{Runway Gen-4.5} & \nolinkurl{runway-gen4p5} & Runner verified \\
% model-id: sora2
\href{https://openai.com/index/sora-2/}{Sora2} & \nolinkurl{sora2} & Runner verified \\
% model-id: veo3
\href{https://storage.googleapis.com/deepmind-media/Model-Cards/Veo-3-Model-Card.pdf}{Veo3} & \nolinkurl{veo3} & Runner verified \\
% model-id: wan-2p5
\href{https://www.alibabacloud.com/help/en/model-studio/text-to-video-api-reference}{Wan 2.5 API} & \nolinkurl{wan-2p5} & Runner verified \\
% model-id: wan-2p6
\href{https://www.alibabacloud.com/help/en/model-studio/text-to-video-api-reference}{Wan 2.6 API} & \nolinkurl{wan-2p6} & Runner verified \\
% model-id: wan-2p7
\href{https://www.alibabacloud.com/help/en/model-studio/text-to-video-api-reference}{Wan 2.7 API} & \nolinkurl{wan-2p7} & Runner verified \\
% model-id: worldlabs
\href{https://www.worldlabs.ai/blog/announcing-the-world-api}{World Labs API} & \nolinkurl{worldlabs} & Runner verified \\
\end{longtable}
\endgroup

\clearpage

\section{Supported Benchmarks}
\label{sec:supported-benchmarks}
The \href{https://openenvision.org/WorldFoundry/docs/evaluation/benchmark-hub}{WorldFoundry Benchmark Hub documentation} lists \textbf{74 benchmarks}. Following the evaluation domains in Table~\ref{tab:zoo_representatives}, we organize them into 23 embodied AI, 33 video generation, and 18 world model benchmarks. Tables~\ref{tab:benchmarks-embodied}--\ref{tab:benchmarks-world} summarize each benchmark's evaluation focus and reported status. Benchmark names link to their sources.

All 74 entries are marked \emph{Integrated \& Verified} in this appendix. Verification scope is specific to the corresponding benchmark recipe and configuration, including its required datasets, checkpoints, judges, and execution environments.

\begingroup
\small
\setlength{\tabcolsep}{4pt}
\renewcommand{\arraystretch}{1.16}
\begin{longtable}{@{}>{\raggedright\arraybackslash}p{0.25\linewidth}>{\raggedright\arraybackslash}p{0.51\linewidth}>{\raggedright\arraybackslash}p{0.20\linewidth}@{}}
\caption{Supported embodied AI benchmarks (23 entries).}\label{tab:benchmarks-embodied}\\
\toprule Benchmark & Evaluation focus & Status \\ \midrule
\endfirsthead
\multicolumn{3}{l}{\small\itshape Table~\ref{tab:benchmarks-embodied} continued}\\
\toprule Benchmark & Evaluation focus & Status \\ \midrule
\endhead
\midrule\multicolumn{3}{r}{\small\itshape Continued on next page}\\
\endfoot
\bottomrule\endlastfoot
\href{https://arxiv.org/abs/1712.05474}{AI2-THOR} & Interactive household navigation and manipulation & Integrated \& Verified \\
\href{https://arxiv.org/abs/2403.09227}{BEHAVIOR-1K} & 1,000 household activities and BDDL goals & Integrated \& Verified \\
\href{https://arxiv.org/abs/2308.12952}{BridgeData V2} & Real robot trajectories; language and goal image conditioning & Integrated \& Verified \\
\href{https://arxiv.org/abs/2112.03227}{CALVIN} & Language instructed manipulation chains (length 1--5) & Integrated \& Verified \\
\href{https://arxiv.org/abs/2410.23208}{Kinetix} & Open ended 2D physics control tasks & Integrated \& Verified \\
\href{https://arxiv.org/abs/2604.11689}{LARYBench} & Latent action extraction, classification, and regression & Integrated \& Verified \\
\href{https://arxiv.org/abs/2306.03310}{LIBERO} & Spatial, Object, Goal, and LIBERO-100 lifelong manipulation & Integrated \& Verified \\
\href{https://arxiv.org/abs/2511.11478}{LIBERO-Mem} & Object motion, sequence, relation, and occlusion memory & Integrated \& Verified \\
\href{https://arxiv.org/abs/2510.13626}{LIBERO-Plus} & Seven perturbation factors; robustness at five difficulty levels & Integrated \& Verified \\
\href{https://arxiv.org/abs/2510.03827}{LIBERO-Pro} & Object, configuration, instruction, task, and environment robustness & Integrated \& Verified \\
\href{https://arxiv.org/abs/2302.04659}{ManiSkill} & GPU parallel SAPIEN manipulation skill families & Integrated \& Verified \\
\href{https://arxiv.org/abs/2302.04659}{ManiSkill2} & 20 SAPIEN manipulation task families; multiple embodiments & Integrated \& Verified \\
\href{https://arxiv.org/abs/1910.10897}{Meta-World} & 50 manipulation tasks; multitask and meta learning suites & Integrated \& Verified \\
\href{https://arxiv.org/abs/2502.10550}{MiKASA-Robo} & Object, spatial, sequential, and capacity memory & Integrated \& Verified \\
\href{https://arxiv.org/abs/2603.16861}{MolmoSpaces-Bench} & Pick and place variants; stage and final task success & Integrated \& Verified \\
\href{https://arxiv.org/abs/1909.12271}{RLBench} & 100 vision guided manipulation tasks and procedural variations & Integrated \& Verified \\
\href{https://arxiv.org/abs/2406.02523}{RoboCasa} & Atomic and long horizon kitchen manipulation & Integrated \& Verified \\
\href{https://arxiv.org/abs/2506.06677}{RoboCerebra} & Long horizon manipulation with memory and disturbances & Integrated \& Verified \\
\href{https://arxiv.org/abs/2603.04639}{RoboMME} & Counting, permanence, reference, and imitation memory & Integrated \& Verified \\
\href{https://arxiv.org/abs/2506.18088}{RoboTwin 2.0} & 50 dual arm tasks; clean and randomized settings & Integrated \& Verified \\
\href{https://arxiv.org/abs/2405.05941}{SimplerEnv} & Real to simulation evaluation on Google Robot and WidowX & Integrated \& Verified \\
\href{https://openaccess.thecvf.com/content/ICCV2025/html/Zhang_VLABench_A_Large-Scale_Benchmark_for_Language-Conditioned_Robotics_Manipulation_with_Long-Horizon_ICCV_2025_paper.html}{VLABench} & Language conditioned manipulation with long horizon reasoning & Integrated \& Verified \\
\href{https://arxiv.org/abs/2602.08971}{WorldArena} & Embodied perceptual quality and functional utility & Integrated \& Verified \\
\end{longtable}
\endgroup

\begingroup
\small
\setlength{\tabcolsep}{4pt}
\renewcommand{\arraystretch}{1.16}
\begin{longtable}{@{}>{\raggedright\arraybackslash}p{0.25\linewidth}>{\raggedright\arraybackslash}p{0.51\linewidth}>{\raggedright\arraybackslash}p{0.20\linewidth}@{}}
\caption{Supported video generation benchmarks (33 entries).}\label{tab:benchmarks-video}\\
\toprule Benchmark & Evaluation focus & Status \\ \midrule
\endfirsthead
\multicolumn{3}{l}{\small\itshape Table~\ref{tab:benchmarks-video} continued}\\
\toprule Benchmark & Evaluation focus & Status \\ \midrule
\endhead
\midrule\multicolumn{3}{r}{\small\itshape Continued on next page}\\
\endfoot
\bottomrule\endlastfoot
\href{https://arxiv.org/abs/2401.01651}{AIGCBench} & Image to video fidelity, alignment, motion, and consistency & Integrated \& Verified \\
\href{https://arxiv.org/abs/2504.15376}{CameraBench} & Camera motion classification, VQA, retrieval, and captioning & Integrated \& Verified \\
\href{https://arxiv.org/abs/2406.18522}{ChronoMagic-Bench} & Time lapse coherence and metamorphic amplitude & Integrated \& Verified \\
\href{https://arxiv.org/abs/2407.01094}{DEVIL Dynamics} & Dynamics range, controllability, and quality & Integrated \& Verified \\
\href{https://arxiv.org/abs/2310.11440}{EvalCrafter} & Visual quality, alignment, motion, and temporal consistency & Integrated \& Verified \\
\href{https://proceedings.neurips.cc/paper_files/paper/2023/hash/c481049f7410f38e788f67c171c64ad5-Abstract-Datasets_and_Benchmarks.html}{FETV} & Fine grained content, attribute control, and prompt complexity & Integrated \& Verified \\
\href{https://arxiv.org/abs/2406.04485}{GenAI-Bench} & Human preferences for image/video generation and editing & Integrated \& Verified \\
\href{https://arxiv.org/abs/2503.14378}{IPV-Bench} & Generation and recognition of impossible events & Integrated \& Verified \\
\href{https://arxiv.org/abs/2510.11512}{LikePhys} & Likelihood preference for valid versus impossible physics & Integrated \& Verified \\
\href{https://arxiv.org/abs/2606.27537}{MemoBench} & Visible--disappeared--reappeared visual memory & Integrated \& Verified \\
\href{https://arxiv.org/abs/2602.08025}{MIND} & Interactive memory consistency and action control & Integrated \& Verified \\
\href{https://arxiv.org/abs/2407.06358}{MiraBench} & Long video consistency, motion, quality, and caption alignment & Integrated \& Verified \\
\href{https://arxiv.org/abs/2601.00943}{PhyEduVideo} & Physics education across six scientific domains & Integrated \& Verified \\
\href{https://arxiv.org/abs/2603.14375}{PhyFPS-Bench-Gen} & Physical FPS alignment and temporal stability (100 prompts) & Integrated \& Verified \\
\href{https://arxiv.org/abs/2410.05363}{PhyGenBench} & Physical commonsense and adherence to physical laws & Integrated \& Verified \\
\href{https://arxiv.org/abs/2605.10806}{PhyGround} & Criteria grounded physics in text/image to video generation & Integrated \& Verified \\
\href{https://arxiv.org/abs/2501.09038}{Physics-IQ Original} & Five second real physics continuation; original protocol & Integrated \& Verified \\
\href{https://arxiv.org/abs/2606.18943}{Physics-IQ Verified} & Physics continuation under the distinct verified protocol & Integrated \& Verified \\
\href{https://arxiv.org/abs/2507.15824}{PhysVidBench} & Caption based physical commonsense QA and tool use & Integrated \& Verified \\
\href{https://arxiv.org/abs/2601.15282}{RBench} & Robot image to video task and embodiment fidelity & Integrated \& Verified \\
\href{https://arxiv.org/abs/2407.14505}{T2V-CompBench} & Seven compositional axes; attribute/action binding and numeracy & Integrated \& Verified \\
\href{https://arxiv.org/abs/2407.05965}{T2VSafetyBench} & Harmful content and temporal safety risks & Integrated \& Verified \\
\href{https://arxiv.org/abs/2311.17982}{VBench} & 16 dimensions of quality, temporal coherence, and semantics & Integrated \& Verified \\
\href{https://arxiv.org/abs/2503.21755}{VBench-2.0} & 18 abilities across five intrinsic faithfulness pillars & Integrated \& Verified \\
\href{https://arxiv.org/abs/2411.13503}{VBench++} & Image to video, long video, and trustworthiness extensions & Integrated \& Verified \\
\href{https://arxiv.org/abs/2504.04907}{Video-Bench} & Human preference aligned quality and condition alignment & Integrated \& Verified \\
\href{https://arxiv.org/abs/2406.03520}{VideoPhy} & Semantic adherence and physical commonsense & Integrated \& Verified \\
\href{https://arxiv.org/abs/2503.06800}{VideoPhy2} & Action centric physical commonsense and rule judgements & Integrated \& Verified \\
\href{https://arxiv.org/abs/2512.02942}{VideoScience-Bench} & Scientific phenomena, dynamics, and spatiotemporal coherence & Integrated \& Verified \\
\href{https://arxiv.org/abs/2406.15252}{VideoScore} & Learned five aspect video quality/reward scoring & Integrated \& Verified \\
\href{https://arxiv.org/abs/2510.08398}{VideoVerse} & Event level temporal causality and world knowledge & Integrated \& Verified \\
\href{https://arxiv.org/abs/2603.14375}{Visual Chronometer} & Physical FPS prediction on arbitrary videos & Integrated \& Verified \\
\href{https://openaccess.thecvf.com/content/ICCV2025/html/Ling_VMBench_A_Benchmark_for_Perception-Aligned_Video_Motion_Generation_ICCV_2025_paper.html}{VMBench} & Five perception aligned video motion dimensions & Integrated \& Verified \\
\end{longtable}
\endgroup

\begingroup
\small
\setlength{\tabcolsep}{4pt}
\renewcommand{\arraystretch}{1.16}
\begin{longtable}{@{}>{\raggedright\arraybackslash}p{0.25\linewidth}>{\raggedright\arraybackslash}p{0.51\linewidth}>{\raggedright\arraybackslash}p{0.20\linewidth}@{}}
\caption{Supported world model benchmarks (18 entries).}\label{tab:benchmarks-world}\\
\toprule Benchmark & Evaluation focus & Status \\ \midrule
\endfirsthead
\multicolumn{3}{l}{\small\itshape Table~\ref{tab:benchmarks-world} continued}\\
\toprule Benchmark & Evaluation focus & Status \\ \midrule
\endhead
\midrule\multicolumn{3}{r}{\small\itshape Continued on next page}\\
\endfoot
\bottomrule\endlastfoot
\href{https://arxiv.org/abs/2511.19836}{4DWorldBench} & Perceptual quality, condition alignment, physics, and 4D consistency & Integrated \& Verified \\
\href{https://arxiv.org/abs/2607.16401}{Apple-PI} & Law grounded perception, formulation, and deduction & Integrated \& Verified \\
\href{https://arxiv.org/abs/2505.09694}{EWMBench} & Embodied scene, motion, semantic consistency, and diversity & Integrated \& Verified \\
\href{https://arxiv.org/abs/2608.16859}{HarnessEval-W} & Transitions, drift resistance, revisits, and offscreen evolution & Integrated \& Verified \\
\href{https://arxiv.org/abs/2605.03941}{iWorld-Bench} & Interactive generation, trajectory following, and memory & Integrated \& Verified \\
\href{https://arxiv.org/abs/2608.27345}{PAWBench} & Stochastic dynamics calibration and outcome coverage & Integrated \& Verified \\
\href{https://arxiv.org/abs/2512.01989}{Physical AI Bench (PAI-Bench)} & Generation and conditional generation tracks only & Integrated \& Verified \\
\href{https://arxiv.org/abs/2605.15178}{SANA-WM 80-Scene Benchmark} & 80 scenes per split; minute scale camera control and revisits & Integrated \& Verified \\
\href{https://arxiv.org/abs/2603.13215}{STEVO-Bench} & Control, physics, and unobserved state evolution & Integrated \& Verified \\
\href{https://arxiv.org/abs/2507.18107}{T2VWorldBench} & Six domains of world knowledge (1,200 prompts) & Integrated \& Verified \\
\href{https://arxiv.org/abs/2605.25874}{WBench} & Multi turn quality, setting, interaction, consistency, and physics & Integrated \& Verified \\
\href{https://arxiv.org/abs/2510.18135}{World-in-World} & Closed loop recognition, QA, navigation, and manipulation utility & Integrated \& Verified \\
\href{https://arxiv.org/abs/2601.21282}{WorldBench} & Diagnostic physical video segmentation and text QA & Integrated \& Verified \\
\href{https://arxiv.org/abs/2502.20694}{WorldModelBench} & Instruction following, commonsense, and physics adherence & Integrated \& Verified \\
\href{https://arxiv.org/abs/2606.11129}{WorldOlympiad} & Physical, interaction, and geometry triathlon & Integrated \& Verified \\
\href{https://arxiv.org/abs/2605.10434}{WorldReasonBench} & Process aware reasoning and multidimensional quality & Integrated \& Verified \\
\href{https://arxiv.org/abs/2504.00983}{WorldScore} & Camera conditioned controllability, quality, and dynamics & Integrated \& Verified \\
\href{https://arxiv.org/abs/2606.20545}{WRBench} & Camera control, visual integrity, and persistent state & Integrated \& Verified \\
\end{longtable}
\endgroup

\clearpage

\section{Supported Metrics}
\label{sec:supported-metrics}

Table~\ref{tab:supported-metrics} lists all \textbf{62 canonical entries} in the WorldFoundry metric registry. The table is compiled from the \href{https://openenvision.org/WorldFoundry/docs/evaluation/metrics}{metric documentation} together with the \href{https://github.com/OpenEnvision/WorldFoundry/blob/main/worldfoundry/evaluation/tasks/metrics/registry.py}{registry and its module declarations}, and therefore includes entries absent from the documentation summary, such as \nolinkurl{laion_aesthetic}. Metrics are grouped by use into multimodal/quality scorers (10), image/feature distribution metrics (26), video distribution metrics (3), perceptual/layout metrics (11), editing metrics (7), and existing-result utilities (5). Aliases and variants map to their canonical entries and are not counted separately.

The registry specifies scoring implementations; benchmark scorecard dimensions and official evaluator outputs are defined by their respective benchmark protocols. $\uparrow$ and $\downarrow$ denote the preferred direction declared for each assessment; ``--'' denotes a result utility with no universal quality ordering. Numeric-field direction follows the underlying reported score. Metric IDs link to their sources or implementations. Learned metrics require their declared pretrained backbones, FVD uses I3D weights, JEDi uses compatible video-encoder/probe weights, and \nolinkurl{natural_score} requires an API judge. Precomputed-feature and layout routines instead consume the specified embeddings, probabilities, masks, or boxes. Support for a registry entry does not by itself establish equivalence to every published evaluator configuration.

\begingroup
\small
\setlength{\tabcolsep}{4pt}
\setlength{\LTpost}{0pt}
\renewcommand{\arraystretch}{1.12}
\begin{longtable}{@{}>{\raggedright\arraybackslash}p{0.28\linewidth}>{\raggedright\arraybackslash}p{0.58\linewidth}>{\centering\arraybackslash}p{0.05\linewidth}@{}}
\caption{Complete supported metric catalog. Metrics are grouped by evaluation use; the underlying registry families comprise 30 distribution, 14 scorer, 13 perceptual, and 5 existing-result entries.}\label{tab:supported-metrics}\\
\toprule
\textbf{Canonical metric ID} & \textbf{Assessment and required inputs} & \textbf{Dir.} \\
\midrule
\endfirsthead
\multicolumn{3}{l}{\small\textit{Table~\ref{tab:supported-metrics} continued}}\\
\toprule
\textbf{Canonical metric ID} & \textbf{Assessment and required inputs} & \textbf{Dir.} \\
\midrule
\endhead
\midrule
\multicolumn{3}{r}{\small\textit{Continued on next page}}\\
\endfoot
\bottomrule
\endlastfoot
\multicolumn{3}{@{}l}{\textbf{Multimodal scorers and quality (10)}}\\*
\addlinespace[2pt]
\href{https://arxiv.org/abs/2104.08718}{\nolinkurl{clip_score}} & CLIPScore: image/video--text embedding alignment & $\uparrow$ \\
\href{https://arxiv.org/abs/2404.01291}{\nolinkurl{vqa_score}} & VQAScore: affirmative-answer probability for a visual question & $\uparrow$ \\
\href{https://arxiv.org/abs/2301.12597}{\nolinkurl{itm_score}} & Image/video--text matching with BLIP-2 or related ITM backbones & $\uparrow$ \\
\href{https://arxiv.org/abs/2406.17100}{\nolinkurl{facescore}} & Learned facial realism and quality reward & $\uparrow$ \\
\href{https://arxiv.org/abs/2305.04923}{\nolinkurl{artscore}} & Learned artness score for generated images & $\uparrow$ \\
\href{https://arxiv.org/abs/2505.20292}{\nolinkurl{facesim_cur}} & Subject-to-video face identity similarity with CurricularFace & $\uparrow$ \\
\href{https://arxiv.org/abs/2505.20292}{\nolinkurl{gme_score}} & GME-Qwen2-VL text--video relevance & $\uparrow$ \\
\href{https://arxiv.org/abs/2505.20292}{\nolinkurl{nexus_score}} & Subject consistency with YOLO-World detection and GME & $\uparrow$ \\
\href{https://arxiv.org/abs/2505.20292}{\nolinkurl{natural_score}} & API-based video naturalness rating & $\uparrow$ \\
\href{https://github.com/LAION-AI/aesthetic-predictor}{\nolinkurl{laion_aesthetic}} & Linear aesthetic predictor on normalized CLIP image/frame features & $\uparrow$ \\
\addlinespace[5pt]
\multicolumn{3}{@{}l}{\textbf{Image and feature distribution (26)}}\\*
\addlinespace[2pt]
\href{https://arxiv.org/abs/1606.03498}{\nolinkurl{inception_score}} & Classifier confidence and label diversity of generated images & $\uparrow$ \\
\href{https://arxiv.org/abs/1706.08500}{\nolinkurl{fid}} & Fr\'echet feature distance; Inception, CLIP, SwAV, frame, and object-crop/SceneFID variants & $\downarrow$ \\
\href{https://arxiv.org/abs/1801.01401}{\nolinkurl{kid}} & Kernel Inception Distance between image feature sets & $\downarrow$ \\
\href{https://arxiv.org/abs/1904.06991}{\nolinkurl{precision_recall}} & Feature-manifold fidelity and coverage of generated/reference sets & $\uparrow$ \\
\href{https://arxiv.org/abs/1904.06991}{\nolinkurl{improved_precision_recall}} & Improved precision/recall and per-image realism score & $\uparrow$ \\
\href{https://arxiv.org/abs/2312.15289}{\nolinkurl{fwd}} & Fr\'echet Wavelet Distance between image collections & $\downarrow$ \\
\href{https://arxiv.org/abs/2401.09603}{\nolinkurl{cmmd}} & CLIP embedding maximum mean discrepancy & $\downarrow$ \\
\href{https://arxiv.org/abs/1812.04948}{\nolinkurl{ppl}} & Perceptual path length of interpolated generator latents & $\downarrow$ \\
\href{https://arxiv.org/abs/2605.06797}{\nolinkurl{mind}} & Monge Inception Distance between image feature sets & $\downarrow$ \\
\href{https://arxiv.org/abs/2104.11222}{\nolinkurl{clean_fid}} & FID with standardized image resizing and preprocessing & $\downarrow$ \\
\href{https://arxiv.org/abs/2210.02410}{\nolinkurl{vendi_score}} & Effective diversity from a feature similarity matrix & $\uparrow$ \\
\href{https://arxiv.org/abs/2206.08549}{\nolinkurl{rarity_score}} & Uncommonness of generated samples relative to real features & $\uparrow$ \\
\href{https://arxiv.org/abs/2104.14767}{\nolinkurl{trend}} & Divergence of fitted truncated generalized-normal feature densities & $\downarrow$ \\
\href{https://arxiv.org/abs/2302.04440}{\nolinkurl{fld}} & Feature Likelihood Divergence using train, test, and generated features & $\downarrow$ \\
\href{https://arxiv.org/abs/2205.13445}{\nolinkurl{multimodal_mid}} & Mutual Information Divergence for paired image--text features & $\uparrow$ \\
\href{https://arxiv.org/abs/1907.08175}{\nolinkurl{fjd}} & Fr\'echet Joint Distance over conditional joint embeddings & $\downarrow$ \\
\href{https://arxiv.org/abs/1905.00643}{\nolinkurl{crosslid}} & Cross Local Intrinsic Dimensionality of feature collections & $\downarrow$ \\
\href{https://arxiv.org/abs/2103.11521}{\nolinkurl{cfid}} & Conditional Fr\'echet Inception Distance from paired embeddings & $\downarrow$ \\
\href{https://arxiv.org/abs/2210.15235}{\nolinkurl{ssd}} & Semantic Similarity Distance for text--image consistency & $\downarrow$ \\
\href{https://arxiv.org/abs/1812.04948}{\nolinkurl{linear_separability}} & Latent linear separability from a classifier confusion matrix & $\uparrow$ \\
\href{https://openreview.net/forum?id=PdZhf6PiAb}{\nolinkurl{rke}} & R\'enyi kernel entropy mode count and relative entropy on features & $\uparrow$ \\
\href{https://arxiv.org/abs/2403.05352}{\nolinkurl{fdd}} & Fr\'echet Denoised Distance using a pretrained denoising autoencoder & $\downarrow$ \\
\href{https://arxiv.org/abs/2004.12361}{\nolinkurl{cis}} & Conditional Inception Score from condition-bucketed class probabilities & $\uparrow$ \\
\href{https://arxiv.org/abs/2010.06715}{\nolinkurl{rnd}} & Random network distillation diversity on features or images & $\uparrow$ \\
\href{https://arxiv.org/abs/2310.17261}{\nolinkurl{attribute_sad}} & Semantic Attribute Distance from precomputed HCS features & $\downarrow$ \\
\href{https://arxiv.org/abs/2310.17261}{\nolinkurl{attribute_pad}} & Pairwise Attribute Distance from precomputed HCS features & $\downarrow$ \\
\addlinespace[5pt]
\multicolumn{3}{@{}l}{\textbf{Video distribution (3)}}\\*
\addlinespace[2pt]
\href{https://arxiv.org/abs/1812.01717}{\nolinkurl{fvd}} & Fr\'echet distance between I3D features of reference/generated clips & $\downarrow$ \\
\href{https://arxiv.org/abs/2407.16124}{\nolinkurl{fvmd}} & Fr\'echet Video Motion Distance from tracked motion features & $\downarrow$ \\
\href{https://arxiv.org/abs/2410.05203}{\nolinkurl{jedi}} & Video JEPA feature distance with a polynomial-kernel MMD & $\downarrow$ \\
\addlinespace[5pt]
\multicolumn{3}{@{}l}{\textbf{Perceptual, reconstruction, and layout (11)}}\\*
\addlinespace[2pt]
\href{https://arxiv.org/abs/1801.03924}{\nolinkurl{lpips}} & Learned perceptual distance between reference/generated images & $\downarrow$ \\
\href{https://doi.org/10.1109/tip.2003.819861}{\nolinkurl{ssim}} & Structural Similarity Index for paired images & $\uparrow$ \\
\href{https://doi.org/10.1109/acssc.2003.1292216}{\nolinkurl{ms_ssim}} & Multi-scale structural similarity for paired images & $\uparrow$ \\
\href{https://doi.org/10.1109/MSP.2008.930649}{\nolinkurl{psnr}} & Peak signal-to-noise ratio from paired-image reconstruction error & $\uparrow$ \\
\href{https://arxiv.org/abs/2104.14294}{\nolinkurl{dino_similarity}} & Cosine similarity of DINO/\href{https://arxiv.org/abs/2304.07193}{DINOv2} reference/generated features & $\uparrow$ \\
\href{https://arxiv.org/abs/2306.09344}{\nolinkurl{dreamsim}} & DreamSim perceptual distance between paired images & $\downarrow$ \\
\href{https://doi.org/10.1109/tip.2011.2131660}{\nolinkurl{cpbd}} & No-reference sharpness from cumulative blur-detection probability & $\uparrow$ \\
\href{https://doi.org/10.1109/tip.2011.2109730}{\nolinkurl{fsim}} & Feature Similarity Index for paired images & $\uparrow$ \\
\href{https://github.com/OpenEnvision/WorldFoundry/blob/main/worldfoundry/evaluation/tasks/metrics/mask_accuracy/mask_accuracy_core.py}{\nolinkurl{mask_accuracy}} & Pixel-wise mask accuracy and intersection-over-union (\href{https://openaccess.thecvf.com/content_cvpr_2015/html/Long_Fully_Convolutional_Networks_2015_CVPR_paper.html}{FCN}) & $\uparrow$ \\
\href{https://arxiv.org/abs/2504.00983}{\nolinkurl{object_detection}} & Phrase-matching object-detection success rate & $\uparrow$ \\
\href{https://arxiv.org/abs/2208.06162}{\nolinkurl{lqs}} & Layout Quality Score: recall, precision, location, and area consistency & $\uparrow$ \\
\addlinespace[5pt]
\multicolumn{3}{@{}l}{\textbf{Editing and semantic consistency (7)}}\\*
\addlinespace[2pt]
\href{https://arxiv.org/abs/2402.07562}{\nolinkurl{semsr}} & Semantic Shift Rate for trigger/origin/target image triplets & $\uparrow$ \\
\href{https://arxiv.org/abs/2309.14756}{\nolinkurl{irs}} & Image Realism Score calibrated using real-image statistics & $\uparrow$ \\
\href{https://arxiv.org/abs/1905.10887}{\nolinkurl{cas}} & Classification Accuracy Score: synthetic-data training, real-data testing & $\uparrow$ \\
\href{https://doi.org/10.3390/s23229287}{\nolinkurl{manipulation_direction}} & Alignment of image and text change vectors for an edit & $\uparrow$ \\
\href{https://arxiv.org/abs/1802.09178}{\nolinkurl{vs_similarity}} & HDGAN visual--semantic similarity of paired embeddings & $\uparrow$ \\
\href{https://doi.org/10.1109/access.2023.3348778}{\nolinkurl{quality_loss}} & CLIP alignment multiplied by unintended text-presence probability & $\downarrow$ \\
\href{https://arxiv.org/abs/2304.13427}{\nolinkurl{object_wise_consistency}} & Object-wise IoU and detection success against guidance boxes & $\uparrow$ \\
\addlinespace[5pt]
\multicolumn{3}{@{}l}{\textbf{Existing-result utilities (5)}}\\*
\addlinespace[2pt]
\href{https://github.com/OpenEnvision/WorldFoundry/blob/main/worldfoundry/evaluation/tasks/metrics/registry.py}{\nolinkurl{artifact_count}} & Counts artifacts recorded on a generation result & -- \\
\href{https://github.com/OpenEnvision/WorldFoundry/blob/main/worldfoundry/evaluation/tasks/metrics/registry.py}{\nolinkurl{required_artifacts_present}} & Returns one when all required artifacts are present & $\uparrow$ \\
\href{https://github.com/OpenEnvision/WorldFoundry/blob/main/worldfoundry/evaluation/tasks/metrics/registry.py}{\nolinkurl{has_artifact}} & Parameterized presence check: \nolinkurl{has_artifact:<name>} & $\uparrow$ \\
\href{https://github.com/OpenEnvision/WorldFoundry/blob/main/worldfoundry/evaluation/tasks/metrics/registry.py}{\nolinkurl{numeric}} & Emits numeric scores and timings stored in result metadata & -- \\
\href{https://github.com/OpenEnvision/WorldFoundry/blob/main/worldfoundry/evaluation/tasks/metrics/registry.py}{\nolinkurl{numeric_value}} & Parameterized numeric field: \nolinkurl{numeric:<name>} & -- \\
\end{longtable}
\endgroup

\end{CJK*}
\end{document}